\documentclass{article} 
\usepackage{iclr2025_conference,times}
\usepackage{times}  
\usepackage{helvet}  
\usepackage{courier}  
\usepackage[hyphens]{url}  
\usepackage{graphicx} 
\usepackage{natbib}  
\usepackage{caption} 
\usepackage[ruled]{algorithm2e} 
\usepackage{algorithmic}
\usepackage{threeparttable}
\usepackage{amsmath}
\usepackage{booktabs}
\usepackage{newfloat}
\usepackage{listings}
\usepackage{subfig}
\usepackage{enumitem}
\usepackage{multirow}
\usepackage{xcolor}
\usepackage{wrapfig}

\usepackage{amsmath,amsfonts,bm}

\def\eqref#1{equation~\ref{#1}}

\def\1{\bm{1}}

\DeclareMathAlphabet{\mathsfit}{\encodingdefault}{\sfdefault}{m}{sl}
\SetMathAlphabet{\mathsfit}{bold}{\encodingdefault}{\sfdefault}{bx}{n}

\usepackage{hyperref}
\usepackage{url}
\usepackage{listings}

\usepackage{wrapfig}

\newcommand{\rev}[1]{\textcolor{red}{#1}}

\newcommand{\topparaheading}[1]{\vspace{-0.2em}\paragraph{#1.}}

\newcommand{\blue}[1]{{\color{blue}{#1}}}

\title{Task-Oriented Diffusion Inversion for High-Fidelity Text-based Editing}

\author{Yangyang Xu$^{1}$~,~Wenqi Shao$^{2}$~,~Yong Du$^{3}$,~Haiming Zhu$^{4}$~,~Yang Zhou$^{4,5}$~,~Ping Luo$^{1,2}$~,~Shengfeng He$^{4}$ \\
$^1$The University of Hong Kong
$^2$Shanghai AI Lab
$^3$Ocean University of China\\
$^4$Singapore Management University
$^5$South China University of Technology \\
\texttt{\small{cnnlstm@gmail.com;}}
\texttt{\small{shengfenghe@smu.edu.sg}}
}

\iclrfinalcopy 
\begin{document}

{\maketitle
\vspace{-1cm}
\begin{figure}[h]
\centering
    \captionsetup[subfloat]{labelformat=empty,justification=centering}
    \subfloat[Source]{
     \begin{minipage}{0.12\linewidth}
     \includegraphics[height=\linewidth,width=\linewidth]{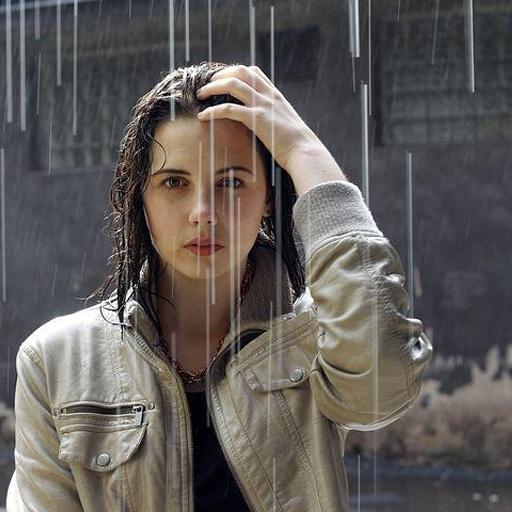}\vspace{3mm}
     \includegraphics[height=\linewidth,width=\linewidth]{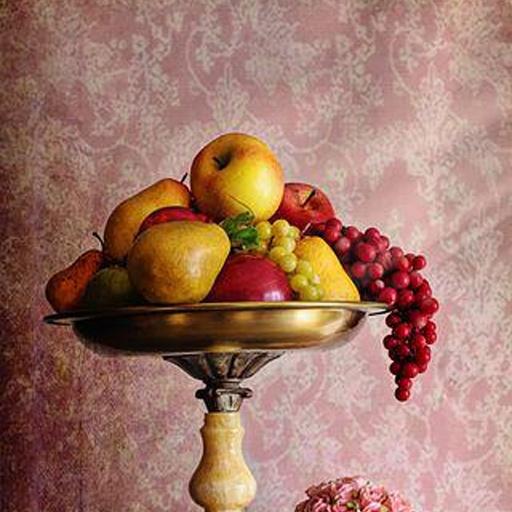}\vspace{3mm}
     \includegraphics[height=\linewidth,width=\linewidth]{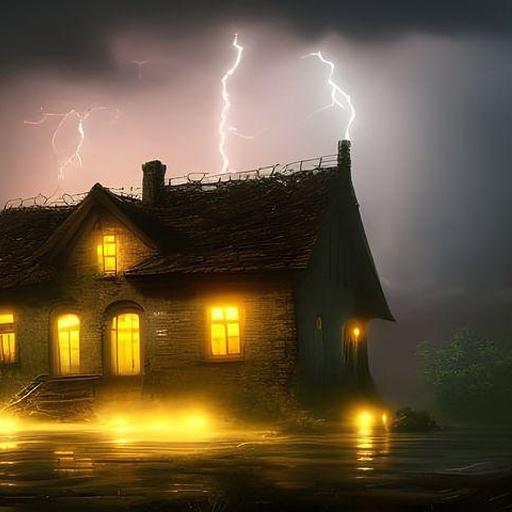}\vspace{3mm}
     \includegraphics[height=\linewidth,width=\linewidth]{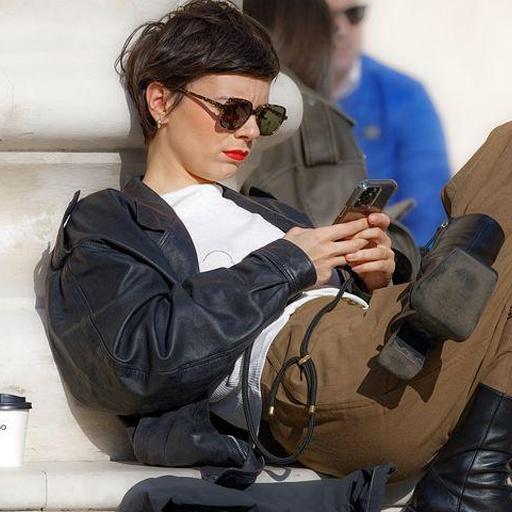}
    \end{minipage}
     }
    \put(-10,115){{\scriptsize{\textsc{A woman in a \rev{jacket} standing in the rain $\rightarrow$ A woman in a \rev{blouse} standing in the rain}}}}
    \hspace{-2.8mm}
    \subfloat[DDIM]{
     \begin{minipage}{0.12\linewidth}
     \includegraphics[height=\linewidth,width=\linewidth]{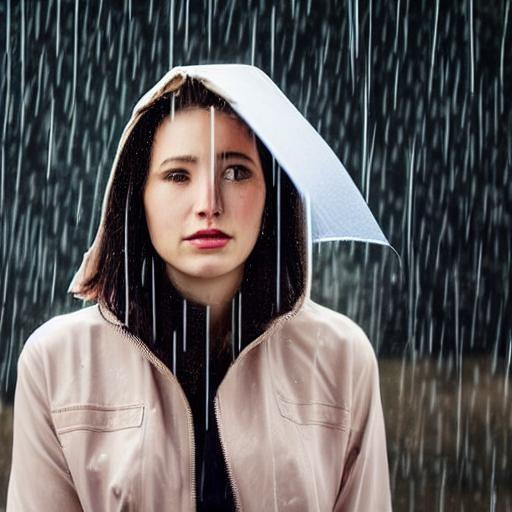}\vspace{3mm}
     \includegraphics[height=\linewidth,width=\linewidth]{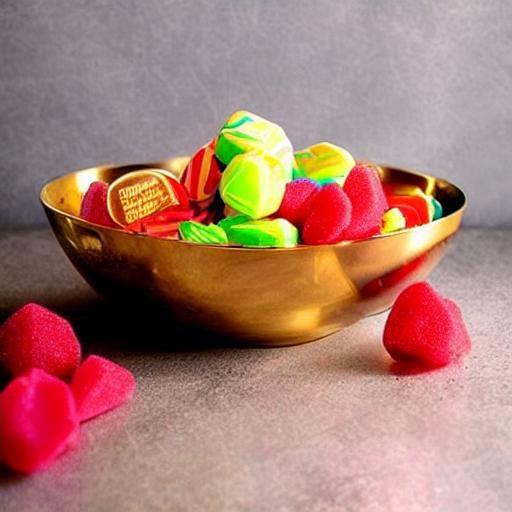}\vspace{3mm}
     \includegraphics[height=\linewidth,width=\linewidth]{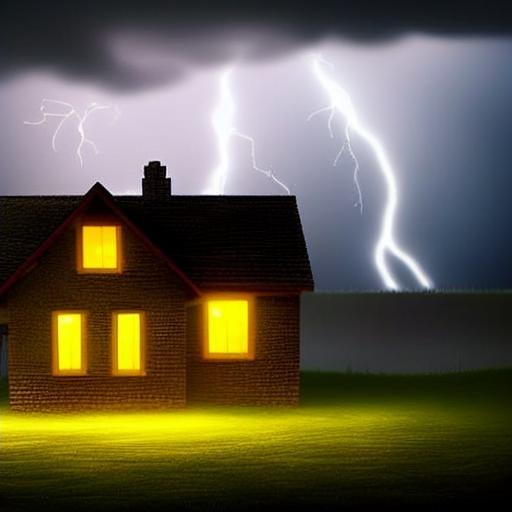}\vspace{3mm}
     \includegraphics[height=\linewidth,width=\linewidth]{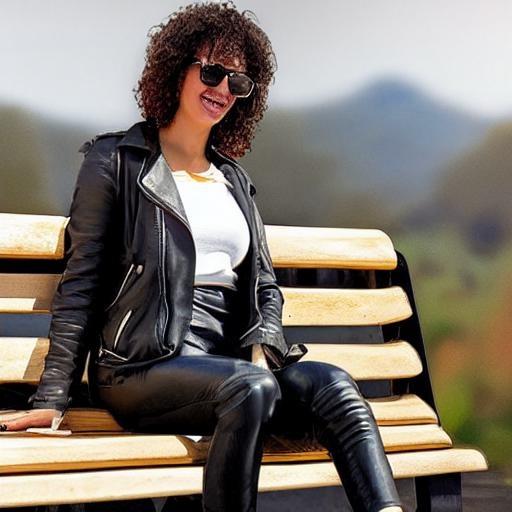}
     \end{minipage}
     }
    \put(-40,57.5){{\scriptsize{\textsc{A gold plated bowl filled with \rev{fruit} $\rightarrow$ A gold plated bowl filled with \rev{candy}}}}}
    \hspace{-2.8mm}
    \subfloat[NTI]{
     \begin{minipage}{0.12\linewidth}
     \includegraphics[height=\linewidth,width=\linewidth]{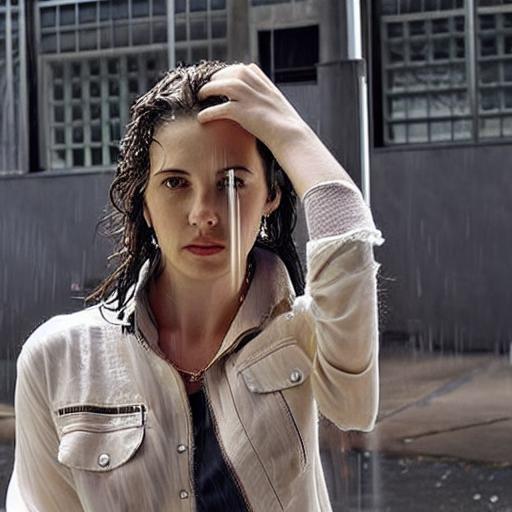}\vspace{3mm}
     \includegraphics[height=\linewidth,width=\linewidth]{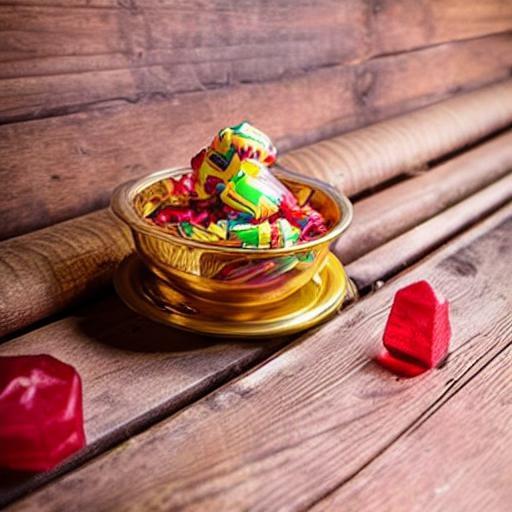}\vspace{3mm}
     \includegraphics[height=\linewidth,width=\linewidth]{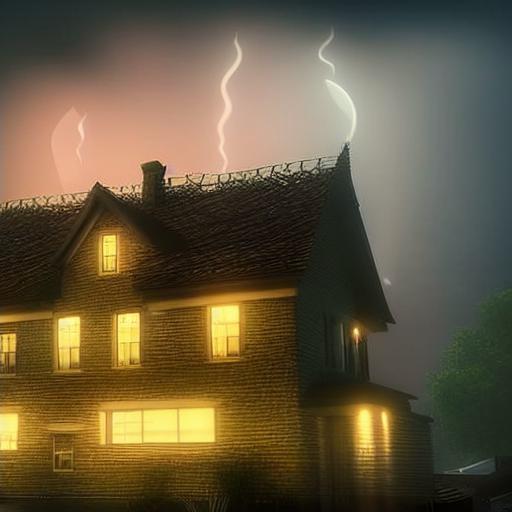}\vspace{3mm}
     \includegraphics[height=\linewidth,width=\linewidth]{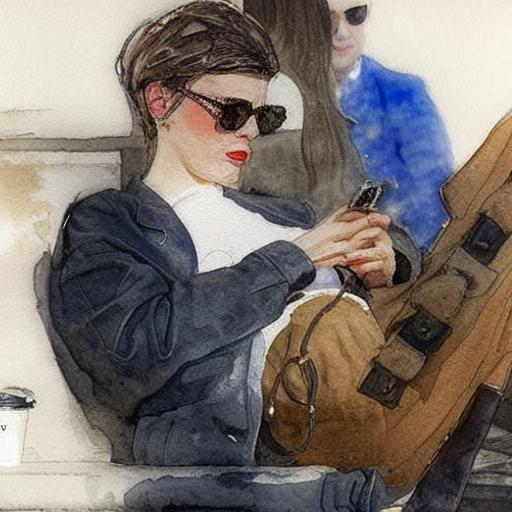}
     \end{minipage}
     }
    \put(-90,0){{\scriptsize{\textsc{A house with \rev{lightning} and rain on it $\rightarrow$A house with rain on it}}}}
    \hspace{-2.8mm}
    \subfloat[NPI]{
     \begin{minipage}{0.12\linewidth}
     \includegraphics[height=\linewidth,width=\linewidth]{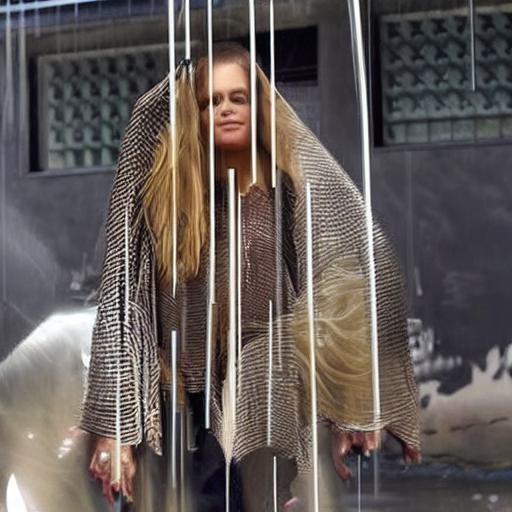}\vspace{3mm}
     \includegraphics[height=\linewidth,width=\linewidth]{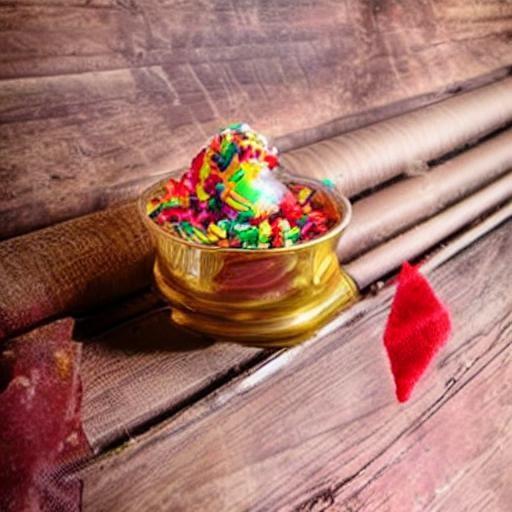}\vspace{3mm}
     \includegraphics[height=\linewidth,width=\linewidth]{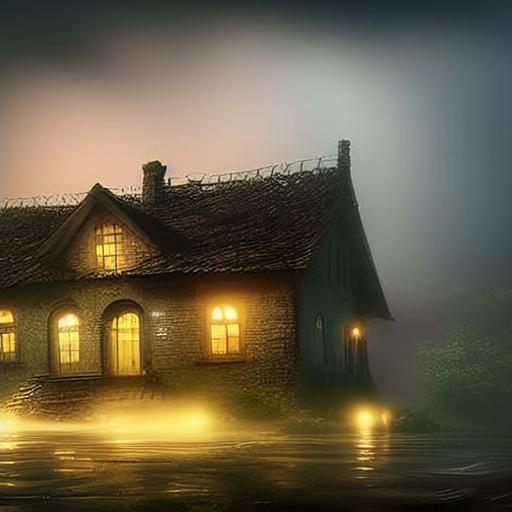}\vspace{3mm}
     \includegraphics[height=\linewidth,width=\linewidth]{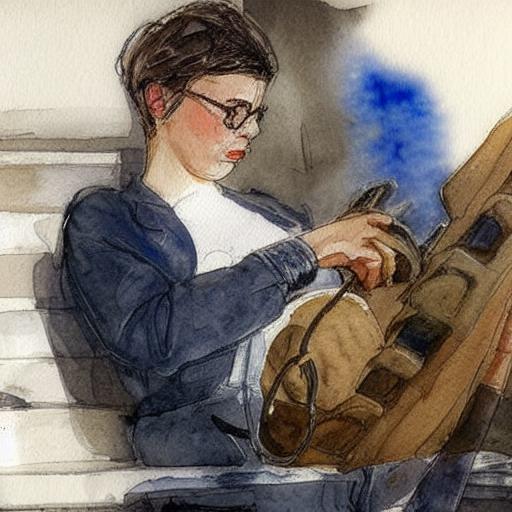}
     \end{minipage}
     }
    \put(-97.5,-57.5){{\scriptsize{\textsc{A woman in sunglasses ... $\rightarrow$ \rev{Watercolor style of} a woman in sunglasses ...}}}}
    \hspace{-2.8mm}
    \subfloat[NMG]{
     \begin{minipage}{0.12\linewidth}
     \includegraphics[height=\linewidth,width=\linewidth]{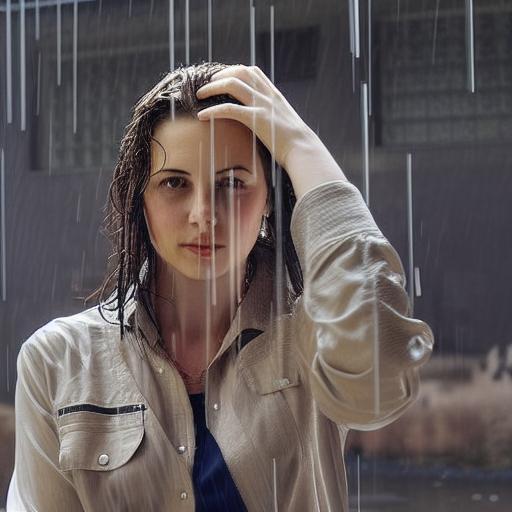}\vspace{3mm}
     \includegraphics[height=\linewidth,width=\linewidth]{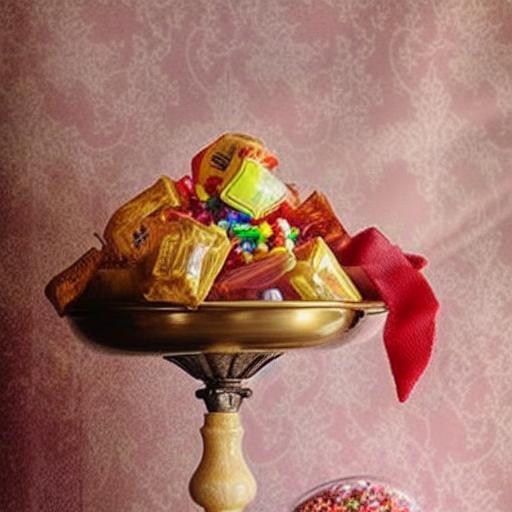}\vspace{3mm}
     \includegraphics[height=\linewidth,width=\linewidth]{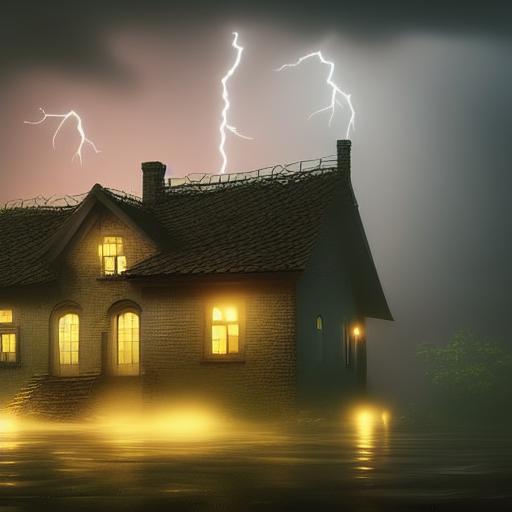}\vspace{3mm}
     \includegraphics[height=\linewidth,width=\linewidth]{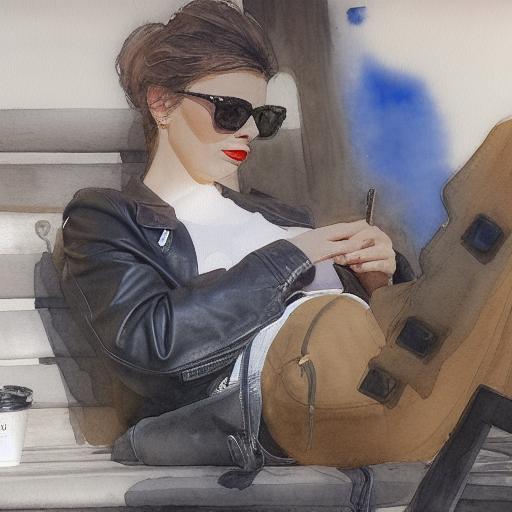}
     \end{minipage}
     }
    \hspace{-2.8mm}
    \subfloat[PNPInv]{
     \begin{minipage}{0.12\linewidth}
     \includegraphics[height=\linewidth,width=\linewidth]{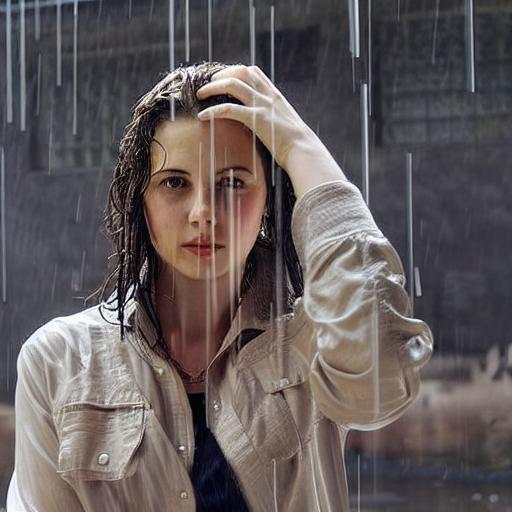}\vspace{3mm}
     \includegraphics[height=\linewidth,width=\linewidth]{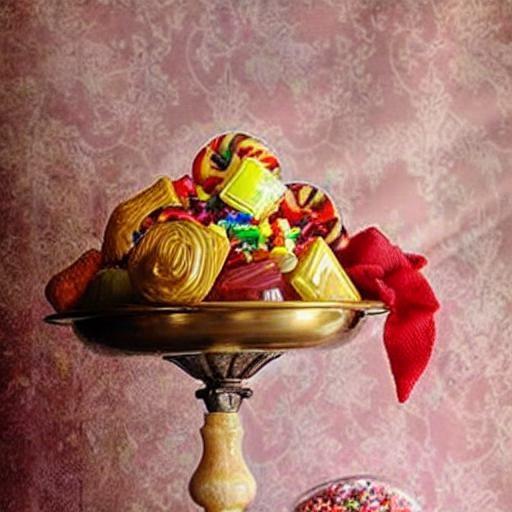}\vspace{3mm}
     \includegraphics[height=\linewidth,width=\linewidth]{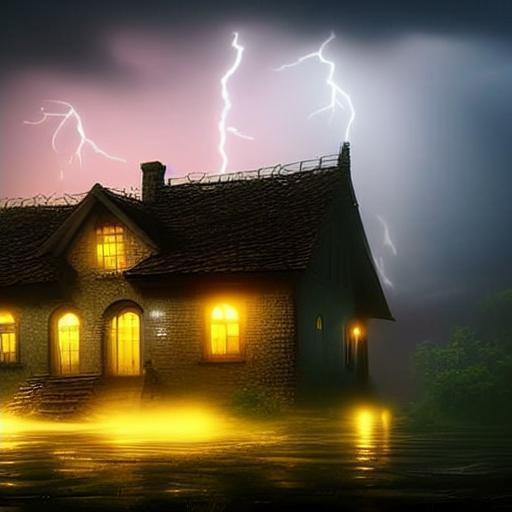}\vspace{3mm}
     \includegraphics[height=\linewidth,width=\linewidth]{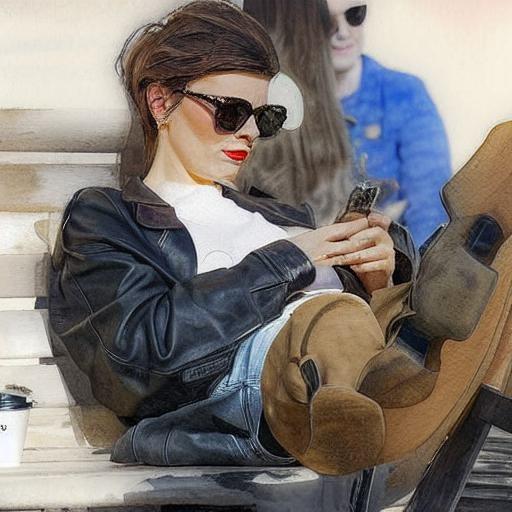}
     \end{minipage}
     }
    \hspace{-2.8mm}
    \subfloat[SPDInv]{
     \begin{minipage}{0.12\linewidth}
     \includegraphics[height=\linewidth,width=\linewidth]{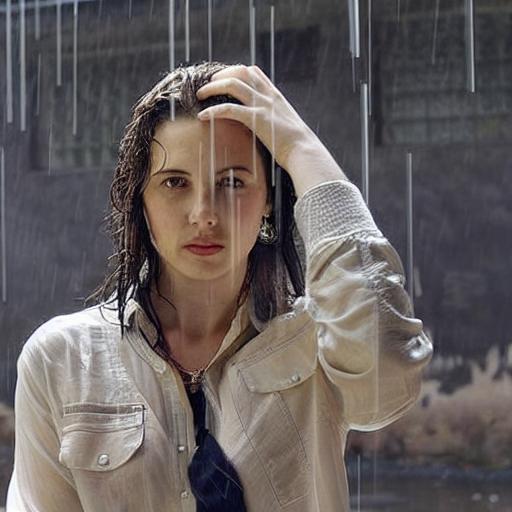}\vspace{3mm}
     \includegraphics[height=\linewidth,width=\linewidth]{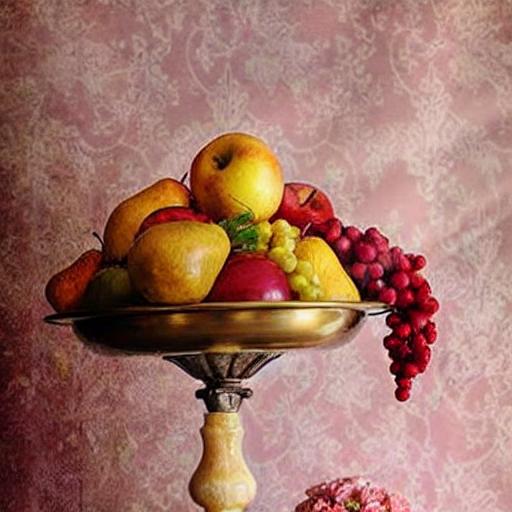}\vspace{3mm}
     \includegraphics[height=\linewidth,width=\linewidth]{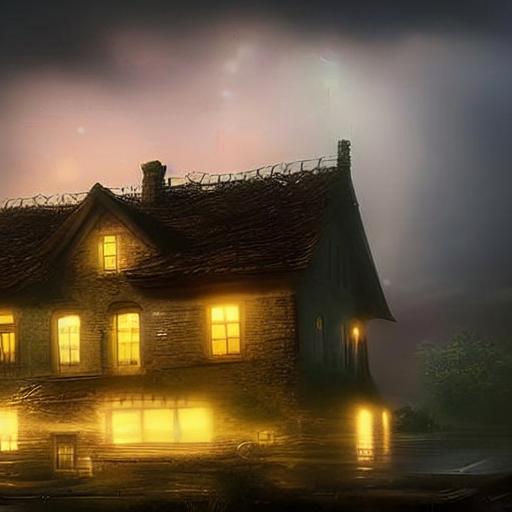}\vspace{3mm}
     \includegraphics[height=\linewidth,width=\linewidth]{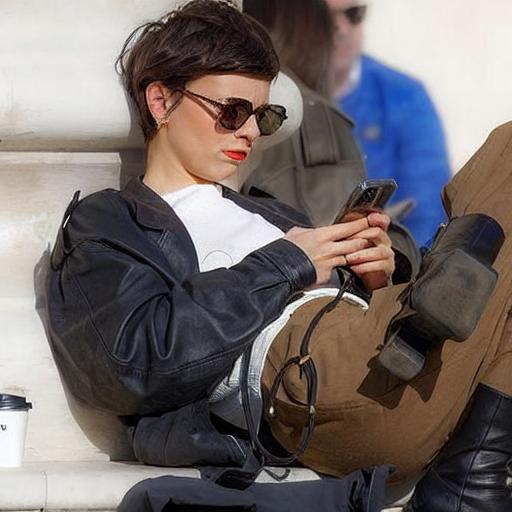}
     \end{minipage}
     }
    \hspace{-2.8mm}
    \subfloat[TODInv]{
     \begin{minipage}{0.12\linewidth}
     \includegraphics[height=\linewidth,width=\linewidth]{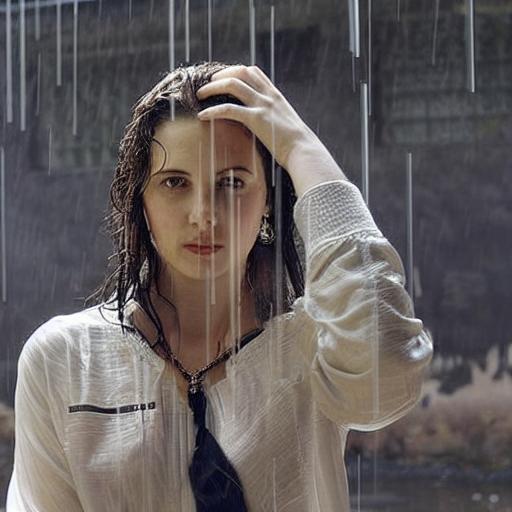}\vspace{3mm}
     \includegraphics[height=\linewidth,width=\linewidth]{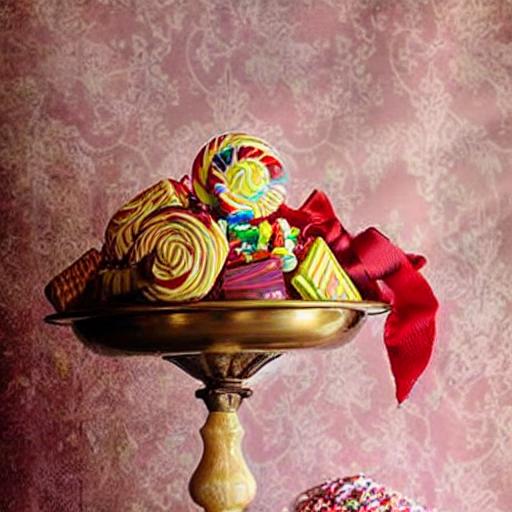}\vspace{3mm}
     \includegraphics[height=\linewidth,width=\linewidth]{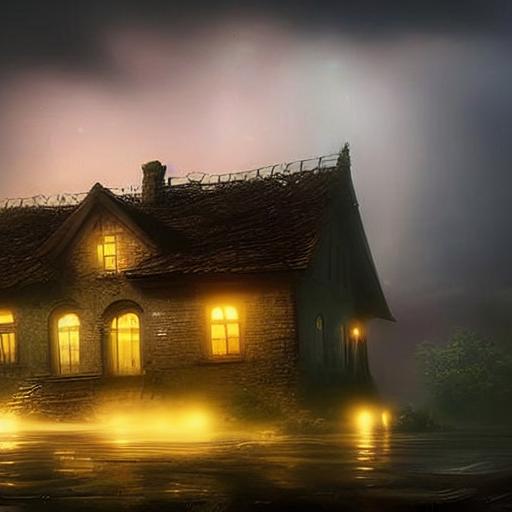}\vspace{3mm}
     \includegraphics[height=\linewidth,width=\linewidth]{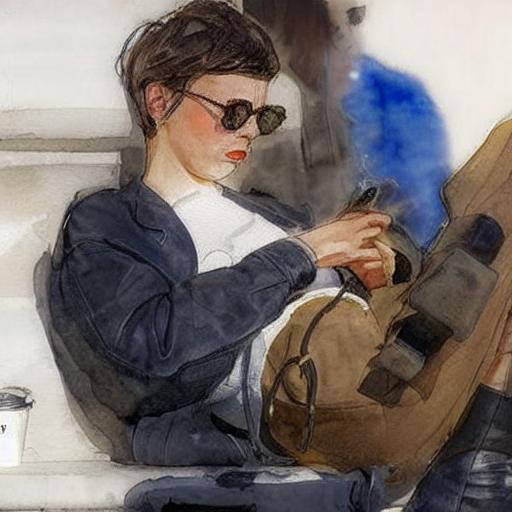}
     \end{minipage}
     }
\vspace{-2mm}
\caption{Our TODInv framework seamlessly integrates the inversion process with editing tasks, enabling diverse high-fidelity text-guided edits such as object replacement, object removal, and stylization. The edited images not only retain the original background but also perfectly align with the target prompts.}
\label{fig:teaser}
\end{figure}
}

\begin{abstract}
Recent advancements in text-guided diffusion models have unlocked powerful image manipulation capabilities, yet balancing reconstruction fidelity and editability for real images remains a significant challenge. In this work, we introduce \textbf{T}ask-\textbf{O}riented \textbf{D}iffusion \textbf{I}nversion (\textbf{TODInv}), a novel framework that inverts and edits real images tailored to specific editing tasks by optimizing prompt embeddings within the extended \(\mathcal{P}^*\) space. By leveraging distinct embeddings across different U-Net layers and time steps, TODInv seamlessly integrates inversion and editing through reciprocal optimization, ensuring both high fidelity and precise editability. This hierarchical editing mechanism categorizes tasks into structure, appearance, and global edits, optimizing only those embeddings unaffected by the current editing task. Extensive experiments on benchmark dataset reveal TODInv's superior performance over existing methods, delivering both quantitative and qualitative enhancements while showcasing its versatility with few-step diffusion model.
\end{abstract}

\section{Introduction}
\label{secintro}

Text-guided diffusion models~\cite{rombach2022high,xue2024raphael,saharia2022photorealistic} have achieved significant success in synthesizing realistic images due to their controllability and diversity. Leveraging these effective text-guided diffusion models, numerous works have explored the generative priors of pre-trained diffusion models and successfully applied these capabilities to various downstream tasks~\cite{zhao2023unleashing,qi2023fatezero,wu2023tune,chen2022diffusiondet,ji2023ddp,baranchuk2021label}, particularly in text-driven image and video editing~\cite{wu2023tune,chai2023stablevideo,qi2023fatezero,tumanyan2023plug,hertz2022prompt,text2video-zero,saharia2022photorealistic,cao2023masactrl}. These technologies enable users to edit images according to their desires via text modification.

When editing a real image \( x_0 \), many text driven image editing methods~\cite{hertz2022prompt,cao2023masactrl,tumanyan2023plug,parmar2023zero} require to invert \( x_0 \) into the latent space of a pre-trained diffusion model to obtain the corresponding latent codes \( \{z_t\}_{t=T}^1 \), which is the inverse process of the diffusion model's sampling procedure. There are two key aspects to this task: the fidelity of the reconstruction and the editability of the latent codes~\cite{garibi2024renoise,pan2023effective}. A naive approach to this task is Denoising Diffusion Implicit Models (DDIM) inversion~\cite{dhariwal2021diffusion,song2021denoising}, which reverses the source image according to the DDIM sampling schedule. However, applying DDIM inversion to text-guided diffusion models often fails due to Classifier Free Guidance (CFG)~\cite{ho2022classifier}, which uses conditional text as input and magnifies the approximation error.

To eliminate the approximation error in DDIM inversion, many works~\cite{sohl2015deep,mokady2023null,han2024proxedit,miyake2023negative} align the differences between conditional and unconditional trajectories to ensure that the source image is faithfully reconstructed. In addition to aligning the two trajectories directly, several works reduce the approximation error at each timestep by optimizing the latent codes. Specifically, AIDI~\cite{pan2023effective}, FPI~\cite{meiri2023fixed}, and ReNoise~\cite{garibi2024renoise} introduce a fixed-point iteration process in each inversion step to obtain accurate latent codes. Furthermore, SPDInv~\cite{li2024source} optimizes latent codes directly based on the difference between two adjacent latent codes. Despite the progress made in fidelity reconstruction, the optimized latent codes often exhibit reduced editability~\cite{garibi2024renoise,parmar2023zero}.

\begin{wrapfigure}{r}{6cm}
\centering
\vspace{-5mm}
\subfloat[$\mathcal{P}$ Space]{
 \begin{minipage}{0.495\linewidth}
 \includegraphics[width=\linewidth]{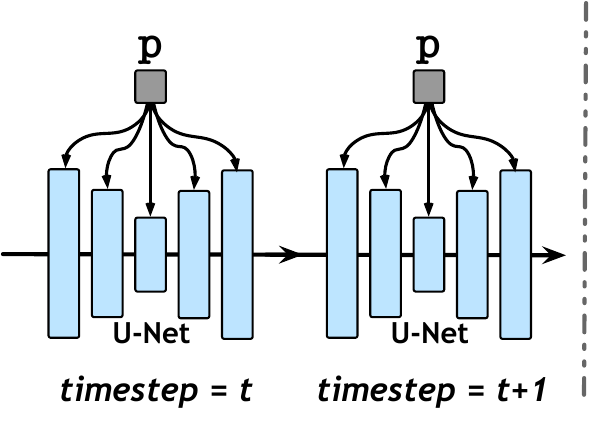}
 \label{space1}
 \vspace{-5mm}
 \end{minipage}
 }
\subfloat[$\mathcal{P}^*$ Space]{
 \begin{minipage}{0.495\linewidth}
 \includegraphics[width=\linewidth]{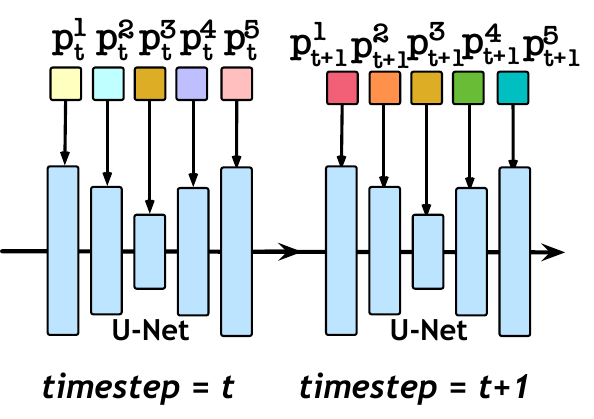}
 \label{space4}
 \vspace{-5mm}
 \end{minipage}
 }
\vspace{-3mm}
\caption{Illustration of original and extended prompt spaces.}
\vspace{-5mm}
\label{fig:spaces}
\end{wrapfigure}

To achieve an ideal balance between reconstruction fidelity and editability, we argue that these two tasks must be intrinsically linked and not treated separately. The inversion process should be highly tailored to the specific editing task at hand. This necessity arises because different edited outputs are modified at varying sampling steps or layers of a diffusion model~\cite{patashnik2023localizing,liew2022magicmix}. As a result, for a given real image, it is crucial to obtain distinct optimal latent codes corresponding to each editing output.





Furthermore, we discern that various text-driven image editing tasks can be broadly categorized into three distinct classes: structure editing, appearance editing, and structure-appearance (i.e., global) editing. The modulation of appearance and structure is controlled by different layers within the U-Net architecture during the diffusion process. This leads us to assert that varying levels of editing should correspondingly activate different tiers of text embeddings. These insights motivate the creation of an inversion framework that dynamically integrates edit instructions in a hierarchical manner, thereby ensuring both high fidelity and precise editability.
In this paper, we propose a novel \textbf{T}ask-\textbf{O}riented \textbf{D}iffusion \textbf{I}nversion (\textbf{TODInv}) framework designed to invert and edit real images tailored to specific editing tasks. Our approach focuses on inverting to prompt embeddings in individual layers. This method represents the input real image through a sequence of prompt embeddings, which can be effectively edited in downstream applications. In particular, we optimize the prompt embeddings within the extended prompt embedding space \(\mathcal{P}^*\)~\cite{alaluf2023neural}. As illustrated in Fig.~\ref{fig:spaces}, unlike the original prompt space \(\mathcal{P}\), which shares the same embedding across different time steps and U-Net layers, the \(\mathcal{P}^*\) space employs distinct embeddings at different layers and time steps. This extended space integrates the disentanglement and expressiveness of time and space, benefiting our inversion in two key aspects:
\\
\romannumeral1) The expressiveness of this latent space facilitates the minimization of inversion errors, significantly enhancing reconstruction accuracy.\\
\romannumeral2) {Compared to the original \(\mathcal{P}\) space, \(\mathcal{P}^*\) space is more disentangled}, which allows for more precise optimization tailored to the specific editing type.

To obtain a faithful reconstruction tailored to the target editing task, we optimize only those prompt embeddings that are agnostic to the current editing, thereby minimizing approximation errors without compromising editability. We conduct extensive experiments on benchmark datasets utilizing various text-driven image editing technologies~\cite{hertz2022prompt,cao2023masactrl,tumanyan2023plug}. As shown in Fig.~\ref{fig:teaser}, the experimental results indicate that our method outperforms existing diffusion inversion techniques in both quantitative and qualitative evaluations. Additionally, our method demonstrates strong performance with few-step diffusion models, further showcasing its versatility and effectiveness.


In summary, our contributions are as follows:
\begin{itemize}[leftmargin=*,nosep]
    \item We present TODInv, a novel diffusion inversion framework that seamlessly links and jointly optimizes inversion and editing processes, achieving both faithful reconstruction and high editability.
    \item We introduce a task-oriented prompt optimization strategy, categorizing various editing tasks into three types. For each class of editing, we minimize the approximation error by optimizing specific prompt embeddings that are irrelevant to the current editing.
    \item Extensive experiments on benchmark dataset demonstrate the effectiveness of our method over state-of-the-art techniques. Our inversion model also supports few-step diffusion models.
\end{itemize}

\section{Related Works}
\label{sec:related}

\topparaheading{Image Editing via Diffusion Models}

Diffusion models~\cite{rombach2022high,saharia2022photorealistic,ramesh2022hierarchical} have made significant advancements in generating diverse and high-fidelity images guided by text prompts. Leveraging these powerful models, numerous works have harnessed their generative capabilities for text-driven image editing. For instance, Prompt-to-Prompt (P2P)~\cite{hertz2022prompt} manipulates attention modules in Stable Diffusion~\cite{rombach2022high} for localized and global edits. Plug-and-Play (PNP)~\cite{tumanyan2023plug} adjusts spatial features and self-attention modules for fine-grained edits, while Pix2pix-Zero~\cite{parmar2023zero} retains cross-attention maps for image-to-image translation. Recently, MasaCtrl~\cite{cao2023masactrl} has enabled complex non-rigid editing by converting the self-attention module into mutual self-attention. Additionally, several works~\cite{wu2023tune,liu2023videop2p,tokenflow2023,zhang2023controlvideo} have extended these methods to video editing. To apply these techniques to real images, inverting the images to the latent space of the diffusion model is a crucial first step.

\topparaheading{Inversion in Diffusion Models}

Early inversion methods for real image editing focused on Generative Adversarial Networks (GANs)~\cite{xu2023rigid,Xu2021ICCV,creswell2018inverting,abdal2019image2stylegan,abdal2020image2styleganpp,xia2021gan}. The advent of diffusion models has shifted attention to diffusion-based inversion methods, which can be categorized into Denoising Diffusion Probabilistic Models (DDPM)-based~\cite{huberman2023edit,wu2023latent} and Denoising Diffusion Implicit Models (DDIM)-based approaches~\cite{garibi2024renoise,dhariwal2021diffusion,song2021denoising,pan2023effective,li2024source,meiri2023fixed}. DDPM-based methods leverage the denoising process but require a large number of inversion steps~\cite{wu2023latent,huberman2023edit}. DDIM-based methods introduce a deterministic DDIM sampler for inversion. However, when CFG is used, DDIM inversion often fails to achieve high-fidelity reconstruction~\cite{mokady2023null}. To address these issues, several works~\cite{mokady2023null,han2024proxedit,miyake2023negative} align the conditional and unconditional trajectories by optimizing the null text token or the prompt embedding. Concurrently, methods like EDICT~\cite{wallace2023edict} and BDIA~\cite{zhang2023exact} introduce invertible networks for inversion. PNPInv~\cite{ju2023direct} merges differences between reconstruction and editing branches, while NMG~\cite{cho2024noise} utilizes spatial context from DDIM inversion for faithful editing. Despite these advancements, existing methods still suffer from approximation errors in DDIM inversion, as the process approximates latent \( x_t \) using \( x_{t-1} \). To eliminate these errors, techniques like AIDI~\cite{pan2023effective}, FPI~\cite{meiri2023fixed}, and ReNoise~\cite{garibi2024renoise} introduce fixed-point iteration processes to optimize latent codes. SPDInv~\cite{li2024source} reformulates this iteration as a loss function. However, directly optimizing latent codes often results in reduced editability~\cite{garibi2024renoise,parmar2023zero}.

In contrast to existing solutions, our task-oriented inversion approach optimizes specific prompt embeddings in an extended prompt space for both inversion and editing, thereby avoiding the trade-off between faithful reconstruction and editability. While our method shares similarities with related works~\cite{mokady2023null,dong2023prompt,han2024proxedit} in prompt optimization, it distinguishes itself in two key aspects:
1) We optimize prompt embeddings to minimize approximation errors in the text-conditioned trajectory of DDIM inversion, rather than merely aligning null-text and text-conditioned trajectories.
2) Our approach specifically connects the inversion process to the editing tasks by optimizing prompt embeddings in the extended \(\mathcal{P}^*\) space, focusing on embeddings irrelevant to the current editing task. This ensures high-fidelity reconstruction tailored to specific edits without compromising the ability to perform diverse and precise modifications.

\topparaheading{Extended Spaces of Diffusion Models}

To better leverage the generative capabilities of diffusion models, several works have analyzed the latent space of these models. Voynov \emph{et al.}~\cite{voynov2023p} extended the original prompt space to \(\mathcal{P}+\) by using different embeddings for different U-Net layers, disentangling structure and appearance. Prospect~\cite{zhang2023prospect} categorized denoising timesteps into style, content, and layout embeddings. NeTI~\cite{alaluf2023neural} introduced a space-time space \(\mathcal{P}*\) for personalized generation. Our work integrates temporal and layer-wise prompt spaces into a unified space, leveraging its expressiveness and disentanglement to achieve high-fidelity reconstruction and editability in diffusion inversion.

\section{Methodology}

\subsection{Preliminaries}

\begin{figure*}
\centering
\hspace{-1.5mm}\includegraphics[width=0.95\textwidth]{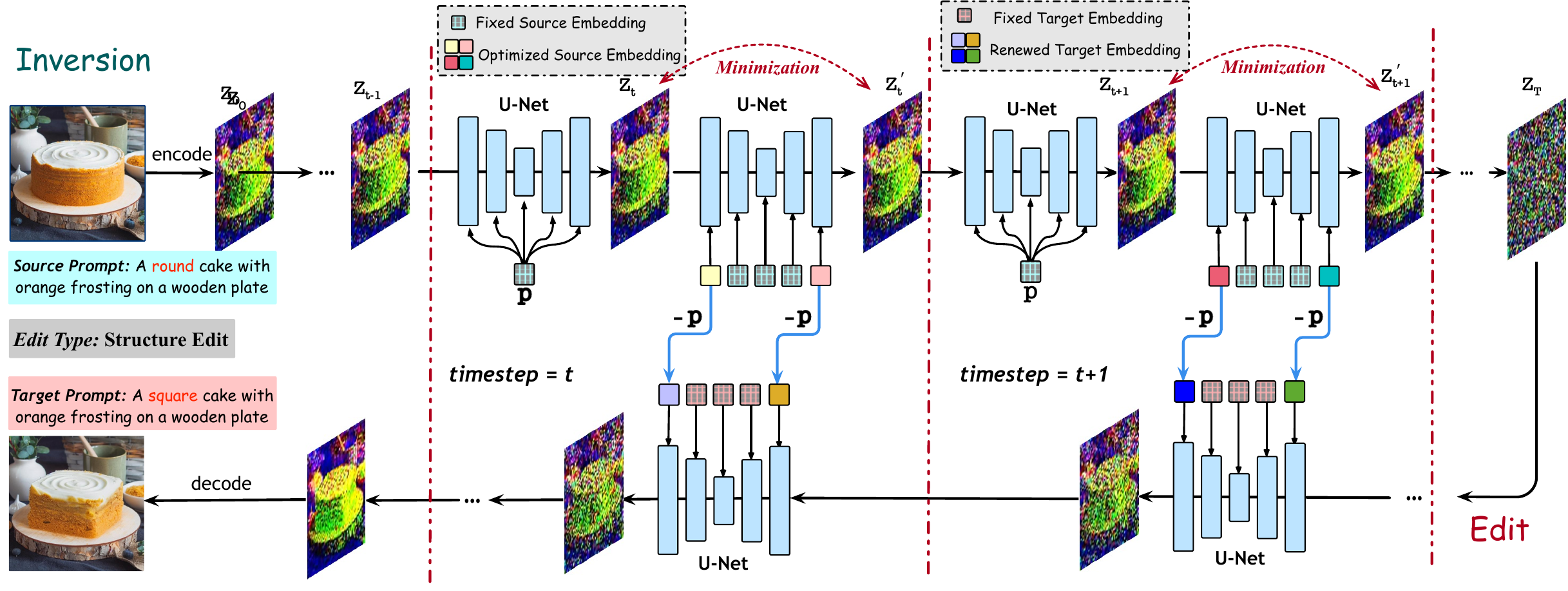}
\vspace{-2.5mm}
\caption{Overview of our TODInv. Given a real image, we first encode the image to the initial latent code $z_0$ using the encoder of Stable Diffusion. In timestep $t$, we get the latent code $z_{t}$ based on latent code $z_{t-1}$ and fixed source prompt embedding $p$ using Eq.~\ref{eq.ddim_inversion2}, but bring the approximation error. Then we use $z_{t}$ to predict latent code $z^{\prime}_{t}$ and minimize their distance by optimizing specific prompt embeddings according to the edit class. The final latent code $z_T$ can be cooperated with various editing methods, with the renewed the target prompts using Eq.~\ref{eq.ddim_inversion9} (the blue arrows)). \emph{Note that only the structure of ``\textsc{cake}'' is edited in this example, which belongs to \textbf{structure edit}, We only optimize the appearance-related prompt embeddings (denoted by the colorful boxes without grids). For more detailed illustration on how to select the optimization layers, please see in Fig.~\ref{fig:select}.}}
\label{fig:overview}
\vspace{-3mm}
\end{figure*}

In this section, we present the background of diffusion models and then analyze the approximation error in DDIM Inversion.

\subsubsection{Diffusion Models}

Diffusion models aim at mapping the random noise $z_T$ to a series latent code $\{z_t\}_{t=T}^1$, where $T$ is the number of timestep, and finally generate a clean image or latent code $z_0$, A diffusion model consists of a training process and a reverse inference process. To train a diffusion model, we add the noise $\epsilon \in \mathcal{N}(0,1)$ to the real image $z_0$ to get the latent variable $z_t$ using follow equation:

\begin{equation}
{z}_t = \sqrt \alpha_t z_0 + \sqrt{1-\alpha_t} \epsilon,
\end{equation}
where $\alpha$ is the hyper-parameter. In a text-guided diffusion model, the text prompt embedding $p$ is conditioned on the network ${\epsilon}_\theta$ to predict the noise, and it is trained using the following equation:
\begin{equation}
\mathcal{L}_{\mathrm{DM}}=\| {\epsilon}-{\epsilon}_\theta ({z}_t, p, t) \|^2_2.
\end{equation}

During the inference, the clean image $z_0$ can be generated from random noise $z_T$ using deterministic DDIM sampler~\cite{song2021denoising} step by step:
\begin{equation}
z_{t-1} = \phi_t z_t + \psi_t \epsilon_\theta(z_t, p, t),
\label{eq.ddim_sampler}
\end{equation}
where $\phi_t$ and $\psi_t$ are sampler parameters, and $\phi_t=\frac{\sqrt{\alpha _{t-1}}}{\sqrt{\alpha _t}}$, $\psi_t=\sqrt{\alpha _{t-1}}\left( \sqrt{\frac{1}{\alpha _{t-1}}-1}-\sqrt{\frac{1}{\alpha _t}-1} \right)$.

\subsubsection{DDIM Inversion}

Diffusion inversion is a reverse process of sampling, which aims to invert a clean image $z_0$ to the noise latent code $z_T$. According to Eq.~\ref{eq.ddim_sampler}, $z_T$ can be inverted from $z_0$ by following equation iteratively:

\begin{equation}
z_{t} = \frac{z_{t-1} - \psi_t \epsilon_\theta(\blue{z_t}, p, t)}{\phi_t}.
\label{eq.ddim_inversion}
\end{equation}

However, directly computing $z_{t}$ using Eq.~\ref{eq.ddim_inversion} is infeasible since the network $\epsilon_\theta(\cdot,\cdot)$ needs the $z_{t}$ as input. DDIM inversion assumes that the Ordinary Differential Equation (ODE) process can be reversed in the limit of infinitesimally small steps, and replace $z_t$ with $z_{t-1}$ for the noise prediction:

\begin{equation}
z_{t} \approx \frac{z_{t-1} - \psi_t \epsilon_\theta(\blue{z_{t-1}}, p, t)}{\phi_t}.
\label{eq.ddim_inversion2}
\end{equation}

This approximation error is introduced into every timestep of DDIM inversion, the accumulated errors decrease the reconstruction quality and editing ability~\cite{pan2023effective,meiri2023fixed,li2024source,garibi2024renoise}. Moreover, in the recent few-step diffusion models~\cite{luo2023latent,luo2023lcm,sauer2023adversarial,song2023consistency}, the approximation error between $z_{t-1}$ and $z_t$ is significantly large, DDIM inversion suffers worse performance on reconstruction~\cite{garibi2024renoise}.

\subsection{Approximation Error Minimization}
\label{sec:aem}
For minimizing the approximation error in the DDIM inversion, existing works~\cite{pan2023effective,meiri2023fixed,garibi2024renoise,li2024source} optimize the latent code $z_{t}$ directly in each timestep. In those works, the fidelity reconstruction can be guaranteed, but compromises the editability.

Instead, we optimize the prompt embeddings, rather than original latent codes. A naive solution is optimizing the prompt embedding in the original prompt space $\mathcal{P}$. In timestep $t$, we first get the latent code $z_{t}$ based on $z_{t-1}$ with DDIM inversion (using Eq.~\ref{eq.ddim_inversion2}), then we take the obtained $z_{t}$ and prompt embedding $p$ to predict another latent code $z_t^\prime$, and we minimizing the distance between the input and output codes by optimizing prompt embedding $p$. The above description can be represented as:
\begin{equation}
z_{t}^\prime = \frac{z_{t-1} - \psi_t \epsilon_\theta({z_{t}}, p, t)}{\phi_t},
\label{eq.ddim_inversion3}
\end{equation}
\begin{equation}
{p}^*= \arg \min _{p} \| z_{t}^\prime - z_{t}\|^2_2.
\label{eq.ddim_inversion5}
\end{equation}
However, optimizing prompt embedding directly has two drawbacks. Firstly, for the original space $\mathcal{P}$, a single text embedding is injected to networks regardless of timesteps and layers of U-Net, the optimization of this shared text embedding limits the minimization of Eq.~\ref{eq.ddim_inversion5} across different timesteps. Secondly, as indicated by the customized diffusion works~\cite{ruiz2023dreambooth,xu2024dreamanime}, the optimized $p^*$ also encodes the image context after optimization, leading to the decreased editability.

\begin{wrapfigure}{r}{6cm}
\centering
\includegraphics[width=0.4\textwidth]{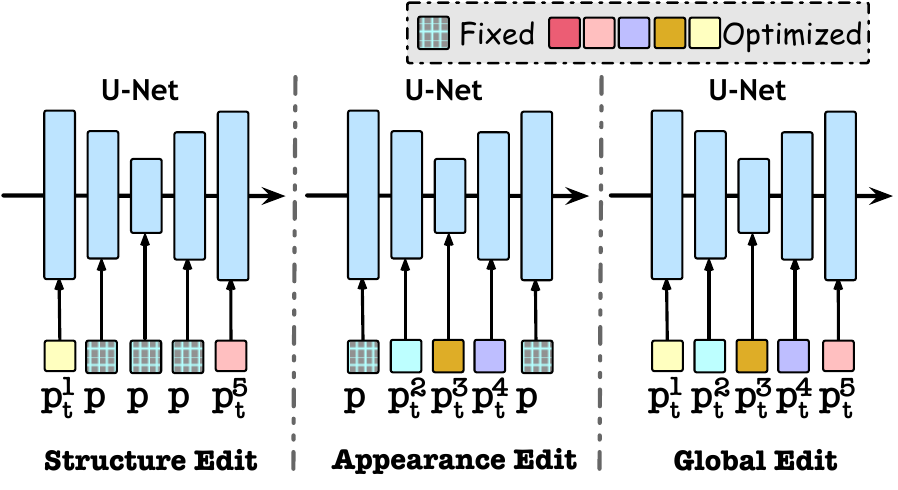}
\vspace{-3mm}
\caption{We categorize all kinds of editing tasks into three classes and divide different layers of U-Net into structure and appearance layers according to their resolutions. For each kind of editing, we only optimize the prompt embeddings that are irrelevant to this editing.}
\label{fig:select}
\vspace{-5mm}
\end{wrapfigure}

\subsection{Task-Oriented Prompt Optimization}
For achieving the high fidelity reconstruction meanwhile preserving the editability, we argue that the inversion process should be oriented to the edit task, as a universally optimal latent code adept at both faithful reconstruction and diverse editing tasks is unattainable. We observe various image editing tasks can be broadly categorized into three classes: structure editing (``\textsc{edit a \blue{round} yellow cake to \blue{square} yellow cake}''), appearance editing (``\textsc{edit a round \blue{yellow} cake to round \blue{red} cake}''), and global editing (``\textsc{edit a \blue{round yellow} cake to \blue{square red} cake}''). On the other hand, It's evidenced that the structure and appearance are modulated by different layers' prompts~\cite{alaluf2023neural,voynov2023p}. This leads us to assert that varying levels of editing should correspondingly different layers of text embeddings.

In our task-oriented inversion, to avoid embedding the content of specific prompts which decreases the editability after minimizing the approximation error, we only optimize the prompt embeddings that are irrelevant to current editing (see in Fig~\ref{fig:select}). For example, for the appearance editing, we only update those embeddings related to the structures. As the appearance-related prompt embeddings are kept fixed, the editability will not be decreased. We chose the extended prompt space $\mathcal{P^*}$ proposed by~\cite{alaluf2023neural} for optimization, as it is evidenced to be more expressive and disentangled.

Let $p_t^i \in \mathcal{P^*}$ denotes the prompt embedding injected to the $i$ resolution layer of U-Net at $t$ timestep, we follow~\cite{alaluf2023neural,voynov2023p} that class different layer prompt embeddings into two groups according to the resolution: the structure prompt set in the low-resolution layers: $P_t^{str}=[p_t^i,i\in low~res~layers]$, and the appearance prompt set controls the high-resolution layers: $P_t^{app}=[p_t^j,j\in high~res~layers]$, we first get the latent code $z_{t}^\prime$ by replacing $p$ with $[P_t^{str},P_t^{app}]$ in Eq.~\ref{eq.ddim_inversion3}:
\begin{equation}
z_{t}^\prime = \frac{z_{t-1} - \psi_t \epsilon_\theta({z_{t}}, [P_t^{str},P_t^{app}], t)}{\phi_t}.
\label{eq.ddim_inversion7}
\end{equation}
Then, for the appearance-related editing, we optimize the irrelevant structure embeddings set $P_t^{str}$, and vice versa. For the global editing, we optimize all the prompt embeddings, which can be represented as:
\begin{equation}
\begin{split}
&{P}^*_{t} =
\begin{cases}
\arg \displaystyle \min_{{P}{_t^{app}}} \| z_{t} - z_{t}^\prime\|^2_2  & {if}~{\texttt{structure~editing}};\\
\arg \displaystyle \min_{{P}{_t^{str}}} \| z_{t} - z_{t}^\prime\|^2_2  & {elif}~{\texttt{appearance~editing}};\\
\arg \displaystyle \min_{{{P}{_t^{str}},{P}{_t^{app}}}} \| z_{t} - z_{t}^\prime\|^2_2  & {else}~{\texttt{global~editing}}.\\
\end{cases}
\end{split}
\label{eq.ddim_inversion8}
\end{equation}
We follow~\cite{li2024source,dong2023prompt} that set the maximum optimization steps as $K$ in each timestep, meanwhile, we also set a threshold $\delta$ to control the termination of the optimization process. By feeding the latent code $z_t$ with the optimized prompt embeddings ${P}^*_{t}$ to the U-Net, with the DDIM sampler, the original image can be reconstructed faithfully. More importantly, with task-oriented optimization, the editability will not be decreased. 
If the same image undergoes multiple types of edits during iterative editing, we choose global editing for optimization. This is because applying different edit categories requires optimizing prompt embeddings across all layers, similar to the global editing category.

During the editing, we leverage the difference between the original and optimized embeddings on the target prompt ${P}^{target}_t$, that is:
\begin{equation}
\tilde{P}{^{target}_t} = {P}^*_t - {P}{_t} + {P}^{target}_t,
\label{eq.ddim_inversion9}
\end{equation}
where $\tilde{P}^{target}_t$ is the renewed target prompt embedding. Incorporated with various text-driven image editing methods~\cite{cao2023masactrl,hertz2022prompt,tumanyan2023plug}, we can edit the real image with target prompt.

\section{Experiments}

\subsection{Experimental Settings}
\label{expset}

\begin{table*}[htbp]
\caption{{Qualitative comparisons with related works using various text-guided editing methods.}}
\vspace{-0.3cm}
\small
\centering
  \renewcommand\arraystretch{0.01}
\setlength{\tabcolsep}{0.05mm}{
\begin{threeparttable}
\begin{tabular}{c|c|c|cccc|cc|c}
\toprule
\toprule
\multicolumn{2}{c|}{\textbf{Method}}           & \textbf{Structure}          & \multicolumn{4}{c|}{\textbf{Background Preservation}} & \multicolumn{2}{c|}{\textbf{CLIP Similarity}} & \multirow{2}{*}{\textbf{Times(s)} $\downarrow$} \\
\cmidrule{1-9}
\textbf{Inverse}          & \textbf{Editing}            & \textbf{Distance}$_{^{\times 10^3}}$ $\downarrow$ & \textbf{PSNR} $\uparrow$     & \textbf{LPIPS}$_{^{\times 10^3}}$ $\downarrow$  & \textbf{MSE}$_{^{\times 10^4}}$ $\downarrow$     & \textbf{SSIM}$_{^{\times 10^2}}$ $\uparrow$    & \textbf{Whole}  $\uparrow$          & \textbf{Edited}  $\uparrow$       \\

\midrule
\textbf{DDIM} & \textbf{P2P}            &69.43           &17.87      &208.80   &219.88   &71.14     &{25.01}      &\textbf{22.44}   &\textbf{11.55}    \\
\textbf{NTI}  & \textbf{P2P}                 &13.44           &{27.03}    &{60.67}  &{35.86}  &{84.11}   &24.75        &21.86     &137.54    \\
\textbf{NPI}  & \textbf{P2P}                 &16.17           &26.21      &69.01    &39.73    &83.40     &24.61        &21.87     &11.75  \\
\textbf{StyleD}  & \textbf{P2P}      &{11.65}         &26.05      &66.10    &38.63    &83.42     &24.78        &21.72     &382.98 \\
\textbf{AIDI}    & \textbf{P2P}              &12.16           &27.01      &56.39    &36.90    &84.27     &24.92        &20.86         &87.21 \\

\textbf{FPI}  & \textbf{P2P}                &{14.71}         &{26.61}    &{61.97}  &{37.64}  &{83.52}   &{23.93}      &{21.35}   &11.75   \\
\textbf{NMG}  & \textbf{P2P}                   &26.64           &25.38     &88.31   &112.77    &81.73   &24.90        &22.16       &16.71  \\
\textbf{ProxEdit} & \textbf{P2P}           &8.80           &28.31      &{44.13}  &{25.72}  &{85.74}   &24.15        &21.36         &{11.75}     \\

\textbf{PNPInv} & \textbf{P2P}                &{11.65}         &{27.22}    &{54.55}  &{32.86}  &{84.76}   &{25.02}      &{22.10}   &19.94   \\
\textbf{SPDInv} & \textbf{P2P}      &{8.81}          &\textbf{28.60}    &{36.01}  &\textbf{24.54}  &{86.23}   &{25.26}      &-      &27.04  \\

\midrule

{\textbf{TODInv}}            & \textbf{P2P}      &\textbf{8.37}       &\textbf{28.39}  &\textbf{39.86}  &{25.71}         &\textbf{86.04} &\textbf{25.47}      & {21.91}      &21.02    \\

\midrule
\midrule
\textbf{DDIM} & \textbf{MasaCtrl}                & 28.38           & 22.17  & 106.62  & 86.97  & 79.67 & 23.96        & 21.16    &\textbf{11.55}      \\
\textbf{AIDI} &\textbf{MasaCtrl}                 & 55.93 & 19.25 & 177.57 & 178.13 & 75.58 & 24.01 &21.07 & 87.21 \\
\textbf{NMG}  & \textbf{MasaCtrl}                     & 40.54 & 20.35 & 127.85 & 135.17 & 77.52 &{24.56} &21.33 &{16.71} \\
\textbf{ProxEdit} & \textbf{MasaCtrl}           &21.28           &23.81      &{85.52}  &{66.47}  &{81.62}   &23.60        &20.94         &{11.75}     \\
\textbf{PNPInv}  & \textbf{MasaCtrl}                 & {24.70}           & {22.64}  & {87.94}  & {81.09}  & {81.33} & {24.38}        & \textbf{21.35}   & {19.94}      \\
\textbf{SPDInv}  & \textbf{MasaCtrl}                 &\textbf{20.48} &{24.12} &{71.74} &{64.77} &{82.54} &{24.61} &- & 27.04\\

\midrule
{\textbf{TODInv}}       & \textbf{MasaCtrl}      &\textbf{19.39}              &\textbf{24.36}         &\textbf{70.17}  &\textbf{62.27}  &\textbf{82.95}        &\textbf{24.74}            &21.20           &21.02  \\

\midrule
\midrule
\textbf{DDIM} &\textbf{PNP} & 28.22 & 22.28 & 113.33 & 83.51 & 79.00 & 25.41 &22.55 &\textbf{11.55}  \\
\textbf{AIDI}  &\textbf{PNP} &25.36 &{23.11} &{98.10} &{78.19} &{80.57} &{25.03}&\textbf{22.70} & 87.21 \\
\textbf{PNPInv} & \textbf{PNP}    &{24.29} & 22.46 & 106.06 & 80.45 & 79.68 &25.41 &22.62  &{19.94}  \\
\textbf{SPDInv} & \textbf{PNP}    &\textbf{15.58} &\textbf{26.72} &{91.55} &\textbf{34.69} &{82.04} &{25.14}&-  & 27.04\\
\midrule
{\textbf{TODInv}}       & \textbf{PNP}                &{21.06}       & {25.13}  &\textbf{78.49}  & {50.16}  &\textbf{82.83} &\textbf{26.08}         &{22.50}      &21.02    \\

\midrule
\midrule
\textbf{DDIM} & \textbf{P2P-Zero} & 61.68 & 20.44  & 172.22 & 144.12 & 74.67 & 22.80 & 20.54 &\textbf{11.55} \\
\textbf{PNPInv} & \textbf{P2P-Zero} & {49.22} & {21.53} & {138.98} & {127.32} & {77.05} & {23.31} & {21.05} &{19.94} \\
\midrule
\textbf{TODInv}             & \textbf{P2P-Zero}         &\textbf{49.86}       & \textbf{21.34}  & \textbf{139.47}  & \textbf{134.66}  & \textbf{76.91} & \textbf{24.19}         & \textbf{21.15}      &21.02    \\

\midrule
\midrule
\textbf{DDIM\tnote{\dag}} & \textbf{ReNoise}  &216.17 &14.52  &319.53 &464.16 &54.30 &21.17 & 18.38 &\textbf{0.56}\\
\textbf{ReNoise\tnote{\dag}} & \textbf{ReNoise} & {107.56} & {15.60} & {271.39} & {704.96} & {62.48} & {25.64} & {23.64} &{2.56} \\
\midrule
\textbf{TODInv\tnote{\dag}}             & \textbf{ReNoise}                &\textbf{86.91}       &\textbf{17.81}  &\textbf{194.00}  &\textbf{224.86}  &\textbf{65.15} &\textbf{26.36}         &\textbf{23.83}      &{4.02}    \\
\bottomrule
\bottomrule
\end{tabular}

\begin{tablenotes}
   \footnotesize
   \item[\dag] use SDXL-Turbo as base model
\end{tablenotes}
\end{threeparttable}}


\label{tab:inversion_based_editing}
\vspace{-0.3cm}
\end{table*} 

\textbf{Dataset.} To evaluate the effectiveness of our hierarchical inversion, we conduct experiments on the PIE-Bench dataset proposed by PNPInv~\cite{ju2023direct}, which consists of 700 images with 9 editing types. Each image is annotated with the source and target prompts. Meanwhile, this dataset also provides the editing region masks for evaluation. For more detailed information about this dataset, please refer to~\cite{ju2023direct}.

\textbf{Evaluation Metrics.} We follow PNPInv~\cite{ju2023direct} which uses several metrics to evaluate our method. We first use the \textbf{Structure Distance} assessed by DINO score~\cite{caron2021emerging} to evaluate the structure distance between original and edited images. Note that this metric cannot be used to evaluate structural edits, as neither higher nor lower values effectively reflect the desired changes. However, we follow the official evaluation proposed by~\cite{ju2023direct}, which adopts a ``lower is better'' approach for the entire dataset. We also introduce several metrics to evaluating the background preservation, which includes \textbf{PSNR}, \textbf{LPIPS}~\cite{zhang2018perceptual}, \textbf{MSE}, and \textbf{SSIM}~\cite{wang2004image}. Those metrics are calculated on the unedited regions, which are defined by the PIE-Bench dataset. Additionally, we introduce CLIP Similarity~\cite{wu2021godiva} to evaluate the text-image consistency between edited images and corresponding target editing text prompts. We follow PNPInv~\cite{ju2023direct} that evaluates CLIP similarity both on the whole image and edited regions, which is denoted by \textbf{Whole} and \textbf{Edited}. At last, we introduce the \textbf{Inference Times} to evaluate different methods' inversion time costs on a single image.


\textbf{Image Editing Methods.} We incooperate with various inversion methods with four text-guided image editing methods, including P2P~\cite{hertz2022prompt}, MasaCtrl~\cite{cao2023masactrl}, PNP~\cite{tumanyan2023plug}, and Pixel-Zero~\cite{parmar2023zero}. Note that not all inversion method provides the source code with MasaCtrl, PNP, and Pixel-Zero editing, we only compare all methods with P2P editing. Since there is no editing method available for the few-step diffusion models, we follow ReNoise which edits the images by replacing the target word directly.

\begin{figure*}[!t]
    \centering
    \captionsetup[subfloat]{labelformat=empty,justification=centering}
    \subfloat[Source]{
     \begin{minipage}{0.0825\linewidth}
     \includegraphics[width=\linewidth]{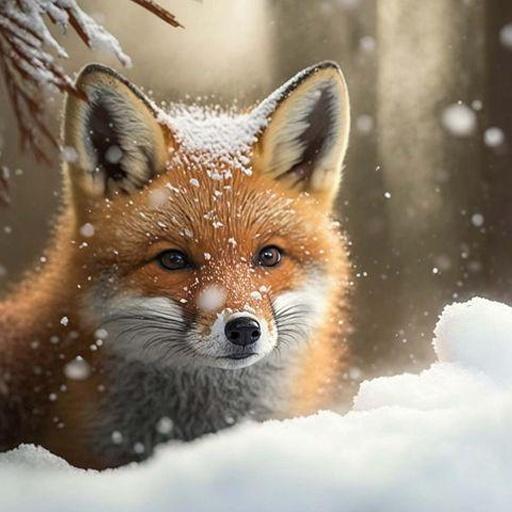}\vspace{3mm}
     \includegraphics[width=\linewidth]{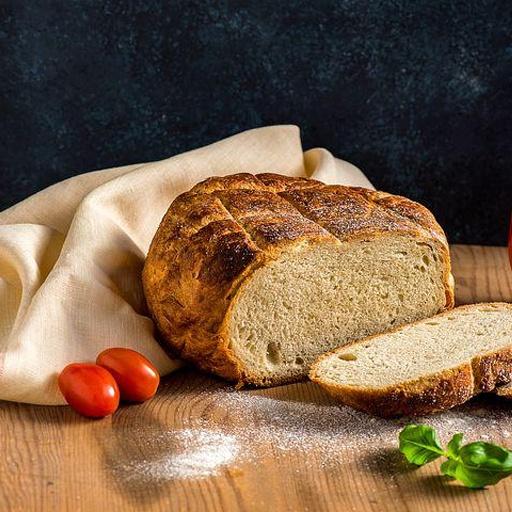}\vspace{3mm}
     \includegraphics[width=\linewidth]{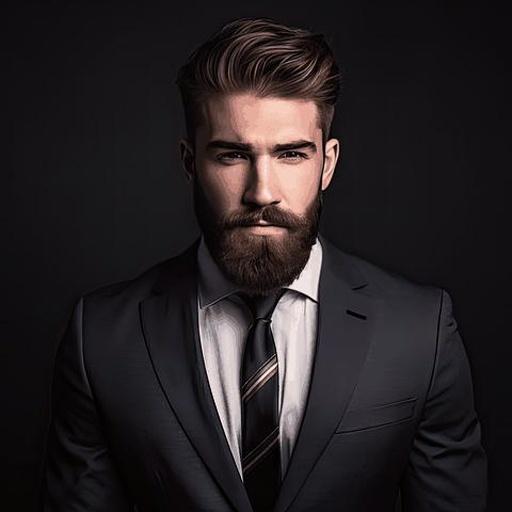}\vspace{3mm}
     \includegraphics[width=\linewidth]{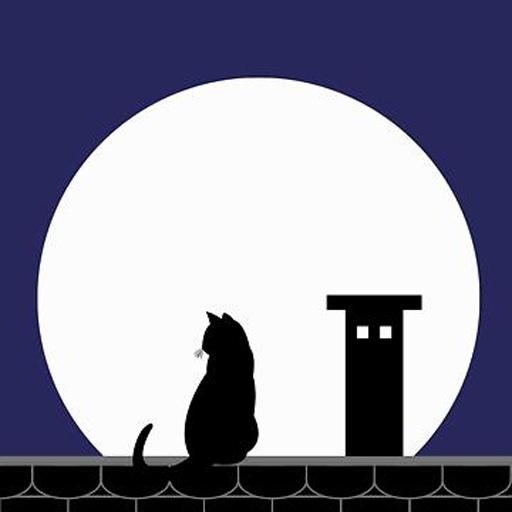}
     \end{minipage}
     }
     \put(60,85){{\scriptsize{\textsc{A fox is sitting in the \rev{snow} $\rightarrow$ A fox is sitting in the \rev{bed}}}}}
    \hspace{-2.8mm}
    \subfloat[DDIM]{
     \begin{minipage}{0.0825\linewidth}
     \includegraphics[width=\linewidth]{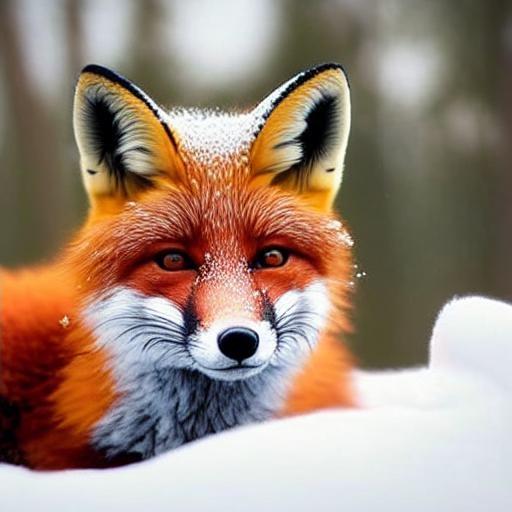}\vspace{3mm}
     \includegraphics[width=\linewidth]{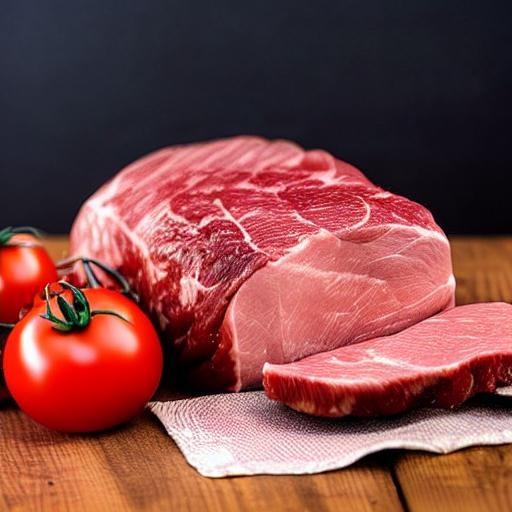}\vspace{3mm}
     \includegraphics[width=\linewidth]{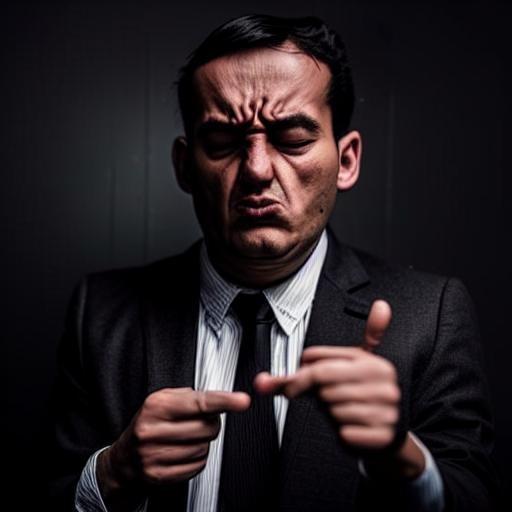}\vspace{3mm}
     \includegraphics[width=\linewidth]{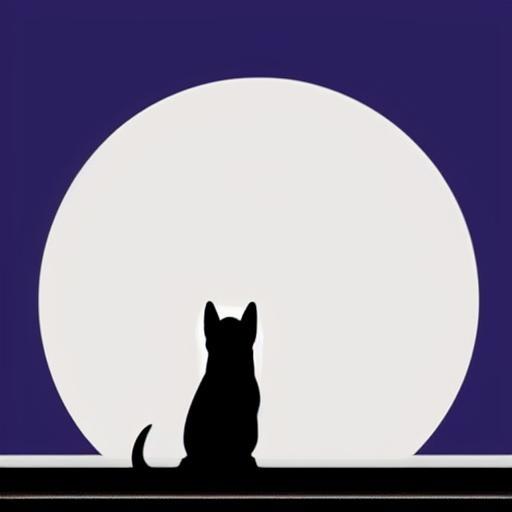}
     \end{minipage}
     }
     \put(5,42.5){{\scriptsize{\textsc{\rev{Bread} on a wooden table with ... $\rightarrow$ \rev{Meat} on a wooden table with ...}}}}
    \hspace{-2.8mm}
    \subfloat[NTI]{
     \begin{minipage}{0.0825\linewidth}
     \includegraphics[width=\linewidth]{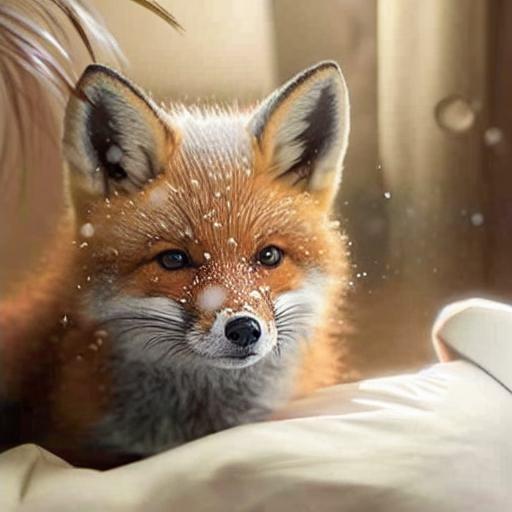}\vspace{3mm}
     \includegraphics[width=\linewidth]{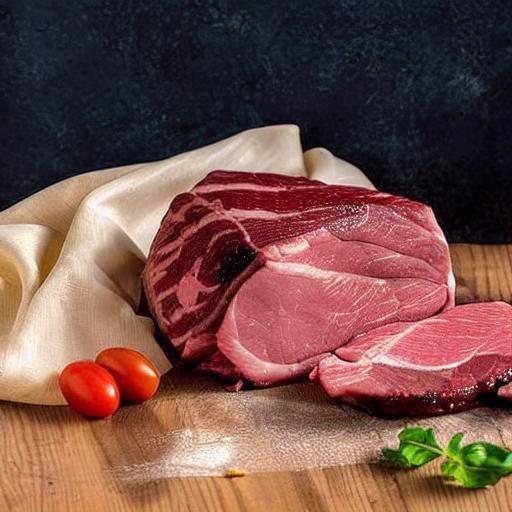}\vspace{3mm}
     \includegraphics[width=\linewidth]{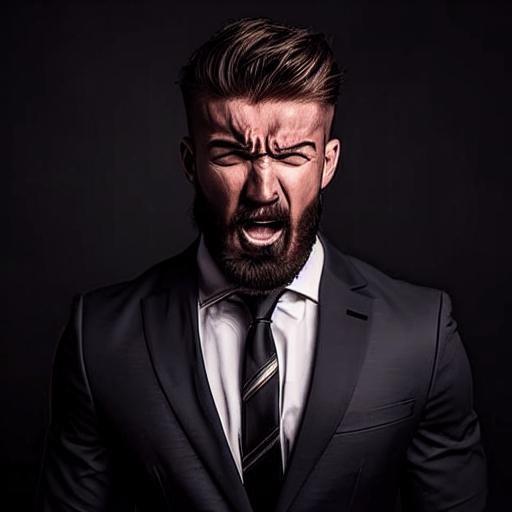}\vspace{3mm}
     \includegraphics[width=\linewidth]{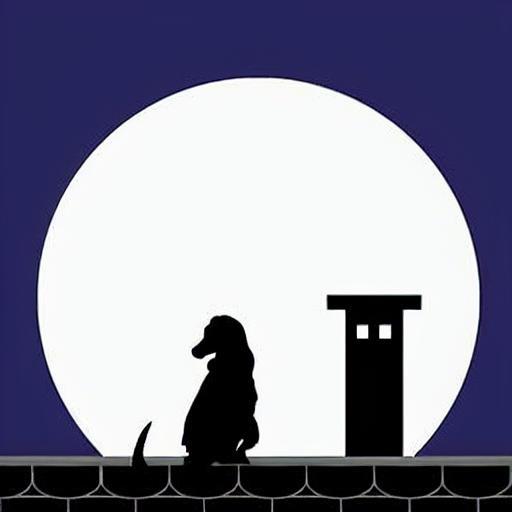}
     \end{minipage}
     }
    \put(40,0){{\scriptsize{\textsc{A \rev{serious} man $\rightarrow$ A \rev{angry} man}}}}
    \hspace{-2.8mm}
    \subfloat[NPI]{
     \begin{minipage}{0.0825\linewidth}
     \includegraphics[width=\linewidth]{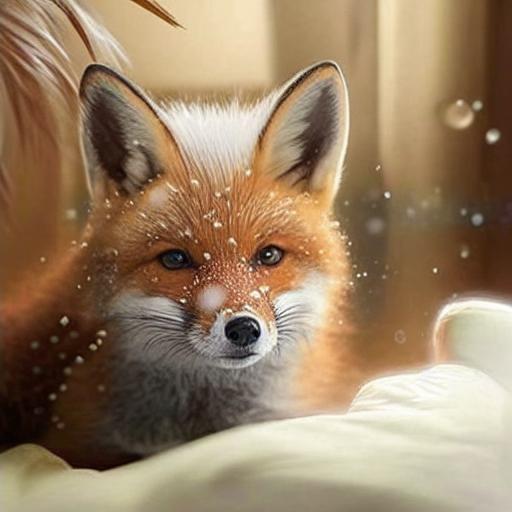}\vspace{3mm}
     \includegraphics[width=\linewidth]{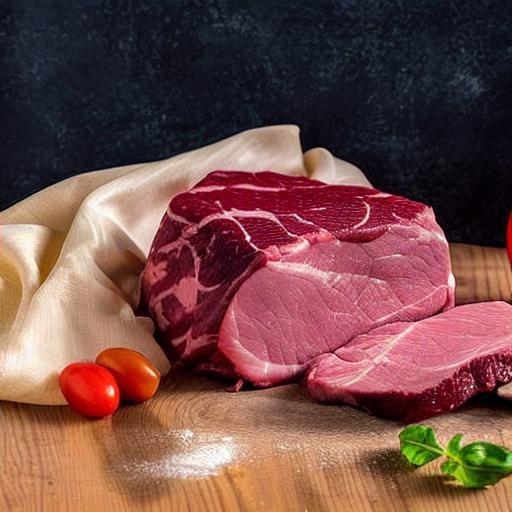}\vspace{3mm}
     \includegraphics[width=\linewidth]{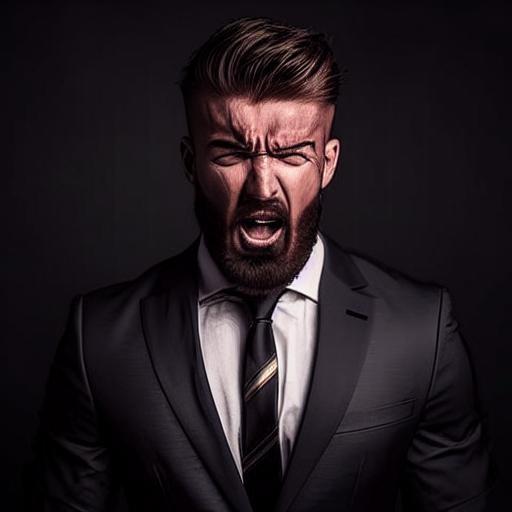}\vspace{3mm}
     \includegraphics[width=\linewidth]{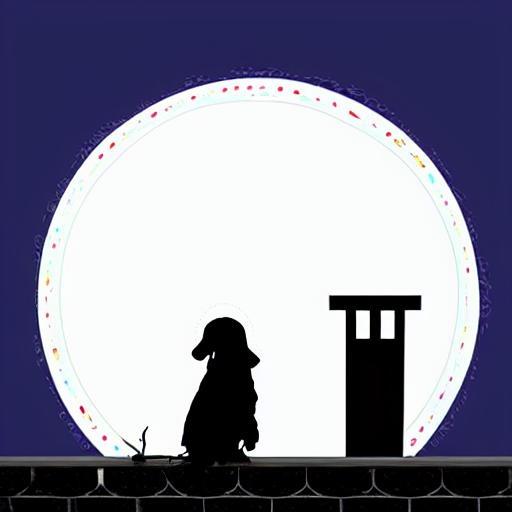}
     \end{minipage}
     }
    \put(-30,-42.5){{\scriptsize{\textsc{\rev{Cat} on the roof at night $\rightarrow$ \rev{Dog} on the roof at night}}}}
    \hspace{-2.8mm}
    \subfloat[AIDI]{
     \begin{minipage}{0.0825\linewidth}
     \includegraphics[width=\linewidth]{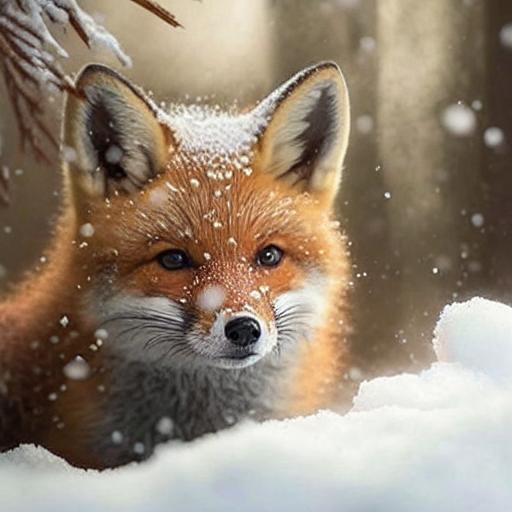}\vspace{3mm}
     \includegraphics[width=\linewidth]{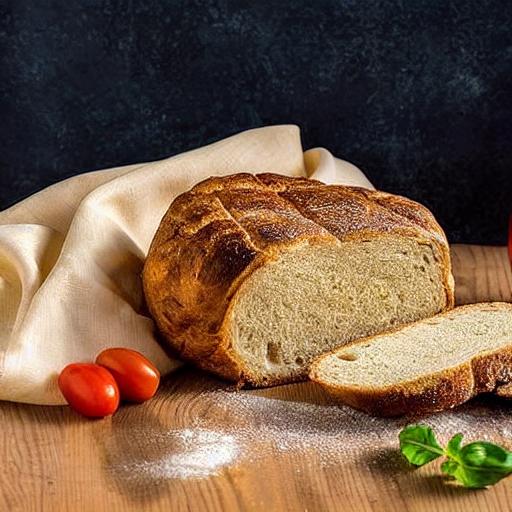}\vspace{3mm}
     \includegraphics[width=\linewidth]{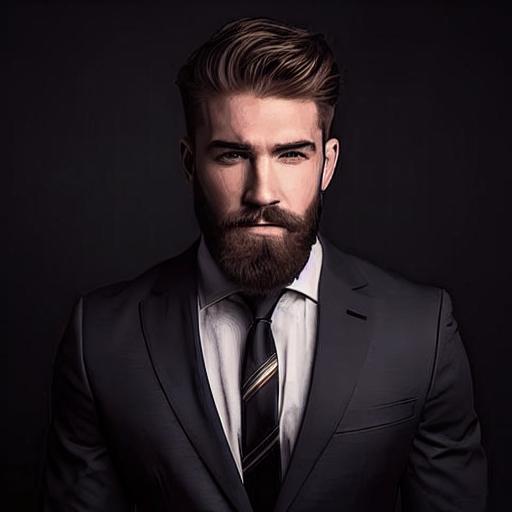}\vspace{3mm}
     \includegraphics[width=\linewidth]{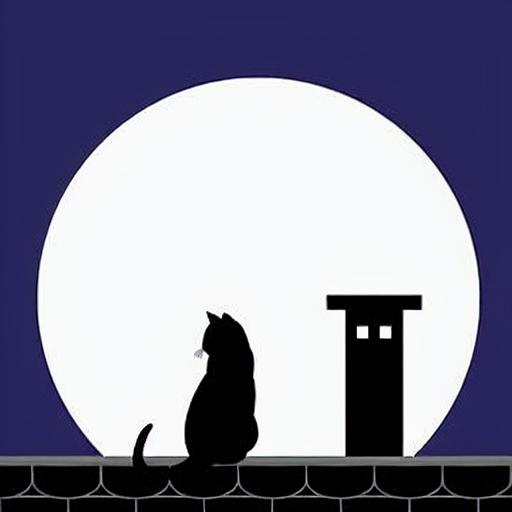}
     \end{minipage}
     }
    \hspace{-2.8mm}
    \subfloat[NMG]{
     \begin{minipage}{0.0825\linewidth}
     \includegraphics[width=\linewidth]{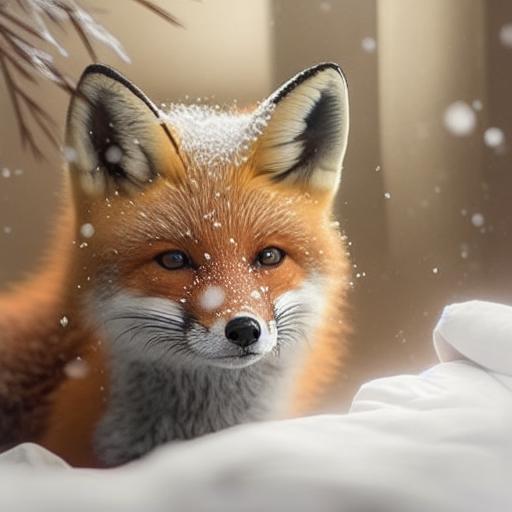}\vspace{3mm}
     \includegraphics[width=\linewidth]{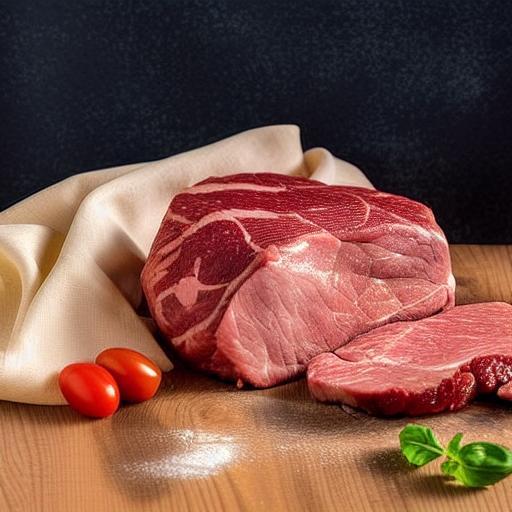}\vspace{3mm}
     \includegraphics[width=\linewidth]{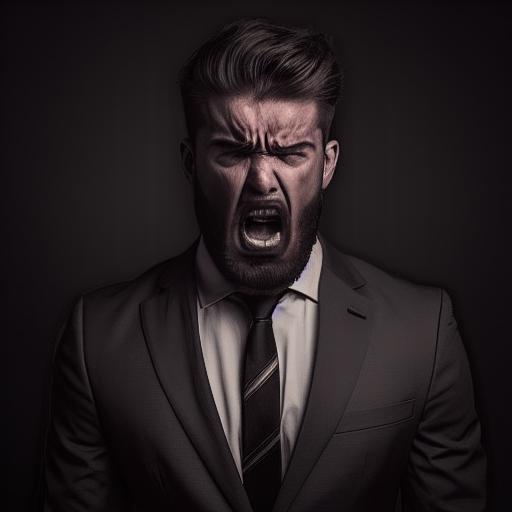}\vspace{3mm}
     \includegraphics[width=\linewidth]{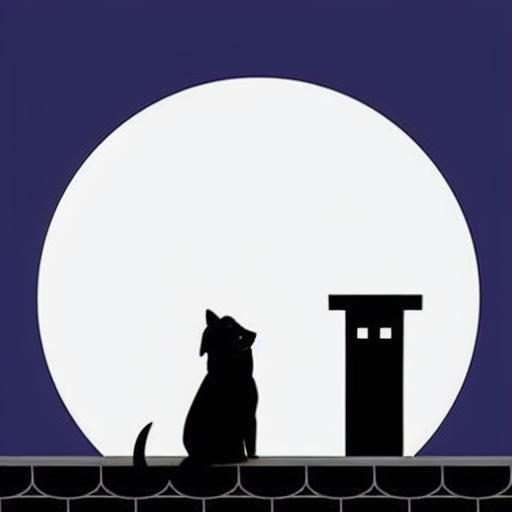}
     \end{minipage}
     }
    \hspace{-2.8mm}
    \subfloat[ProxEdit]{
     \begin{minipage}{0.0825\linewidth}
     \includegraphics[width=\linewidth]{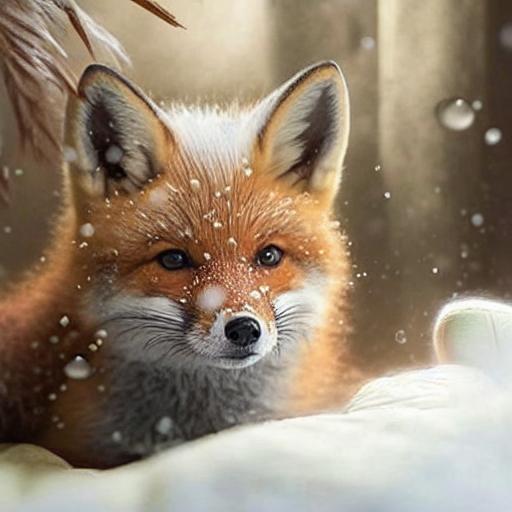}\vspace{3mm}
     \includegraphics[width=\linewidth]{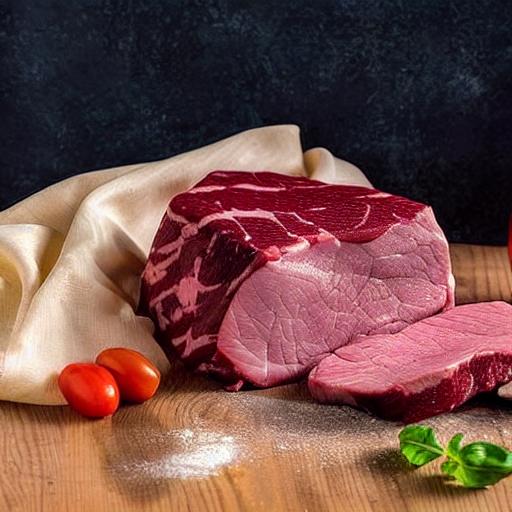}\vspace{3mm}
     \includegraphics[width=\linewidth]{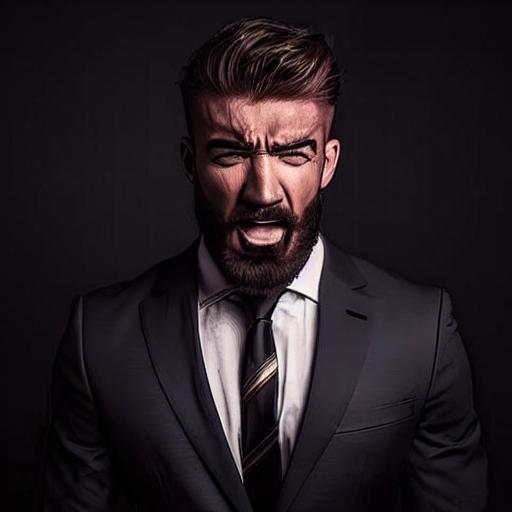}\vspace{3mm}
     \includegraphics[width=\linewidth]{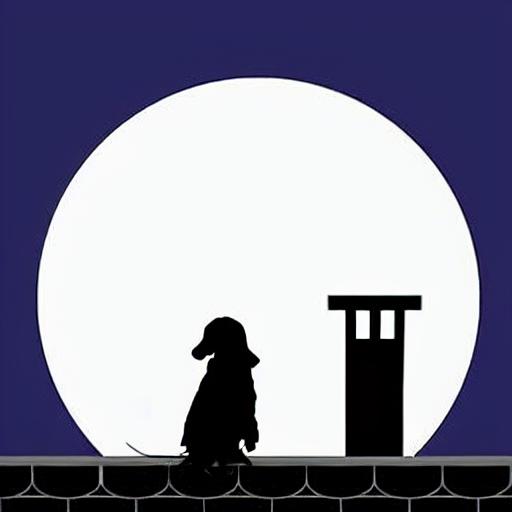}
     \end{minipage}
     }
    \hspace{-2.8mm}
    \subfloat[StyleD]{
     \begin{minipage}{0.0825\linewidth}
     \includegraphics[width=\linewidth]{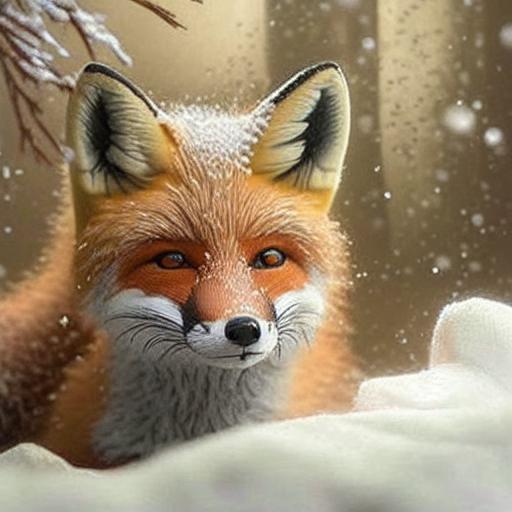}\vspace{3mm}
     \includegraphics[width=\linewidth]{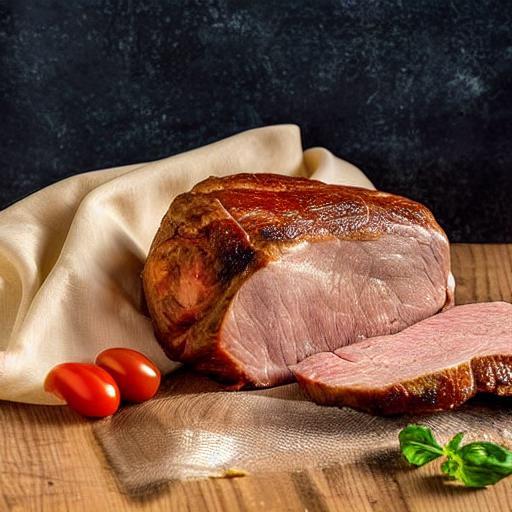}\vspace{3mm}
     \includegraphics[width=\linewidth]{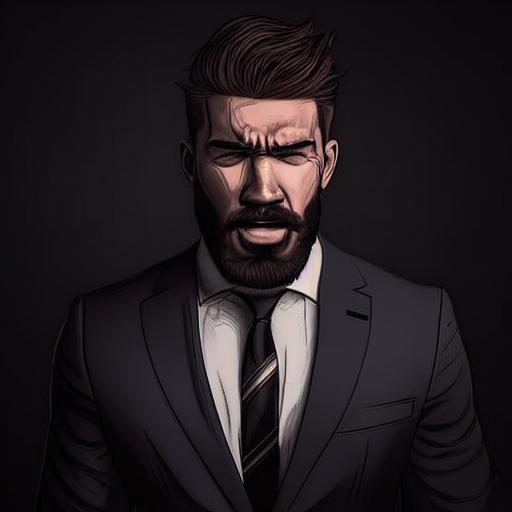}\vspace{3mm}
     \includegraphics[width=\linewidth]{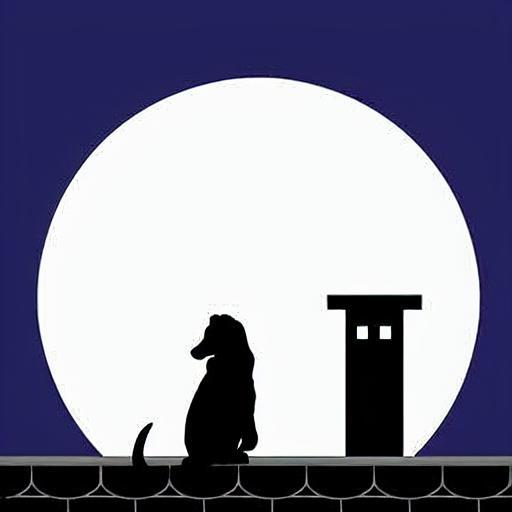}
     \end{minipage}
    }
    \hspace{-2.8mm}
    \subfloat[FPI]{
     \begin{minipage}{0.0825\linewidth}
     \includegraphics[width=\linewidth]{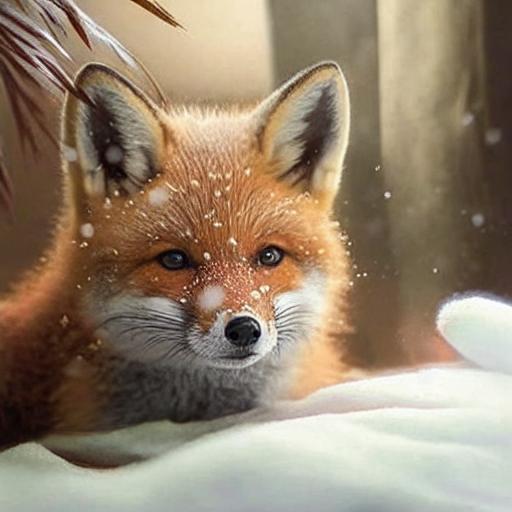}\vspace{3mm}
     \includegraphics[width=\linewidth]{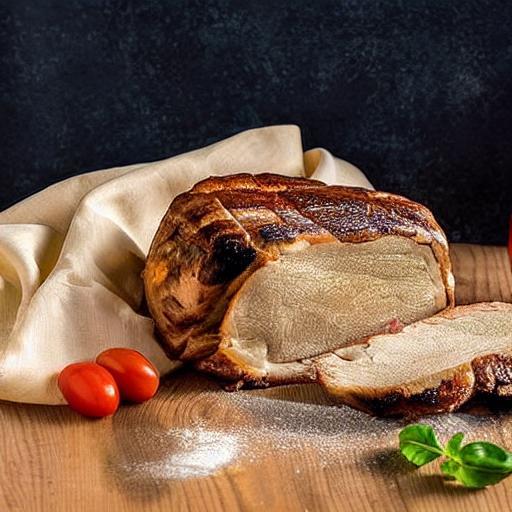}\vspace{3mm}
     \includegraphics[width=\linewidth]{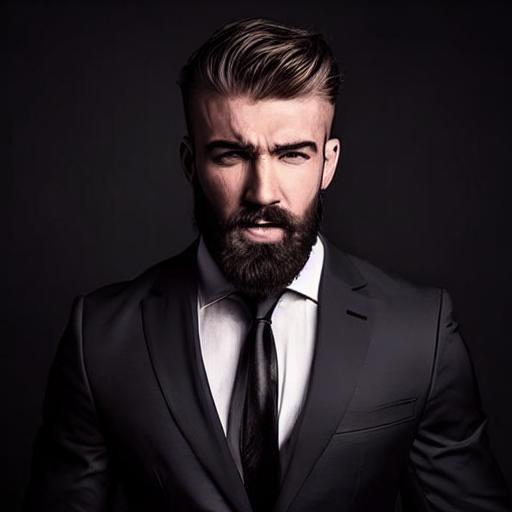}\vspace{3mm}
     \includegraphics[width=\linewidth]{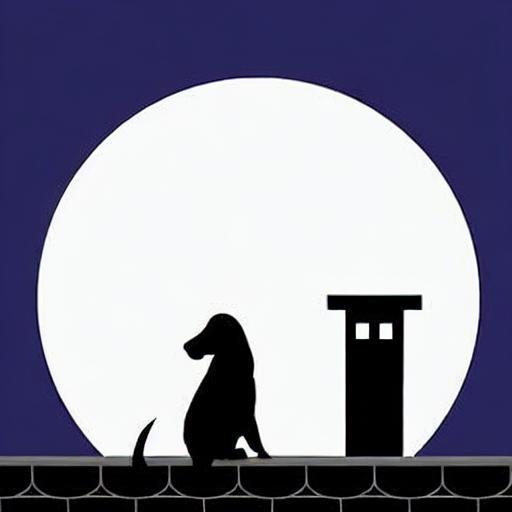}
     \end{minipage}
     }
    \hspace{-2.8mm}
    \subfloat[PNPInv]{
     \begin{minipage}{0.0825\linewidth}
     \includegraphics[width=\linewidth]{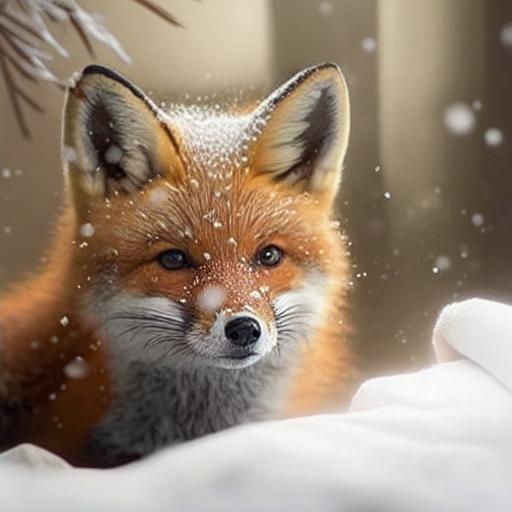}\vspace{3mm}
     \includegraphics[width=\linewidth]{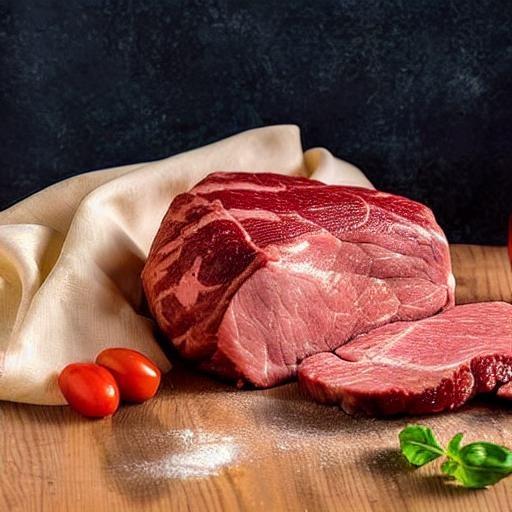}\vspace{3mm}
     \includegraphics[width=\linewidth]{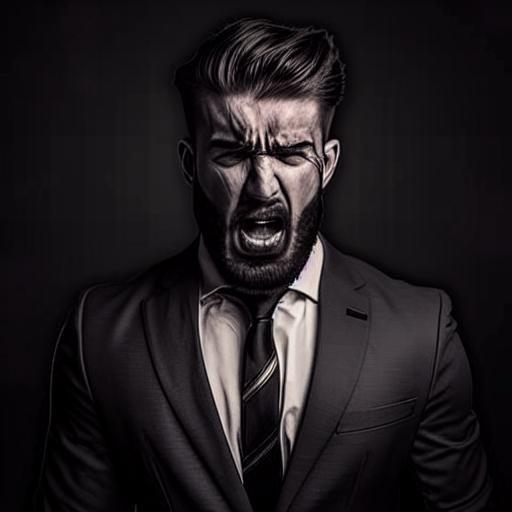}\vspace{3mm}
     \includegraphics[width=\linewidth]{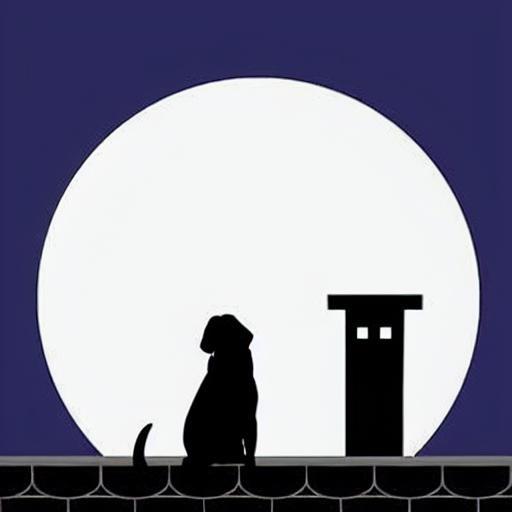}
     \end{minipage}
     }
    \hspace{-2.8mm}
    \subfloat[SPDInv]{
     \begin{minipage}{0.0825\linewidth}
     \includegraphics[width=\linewidth]{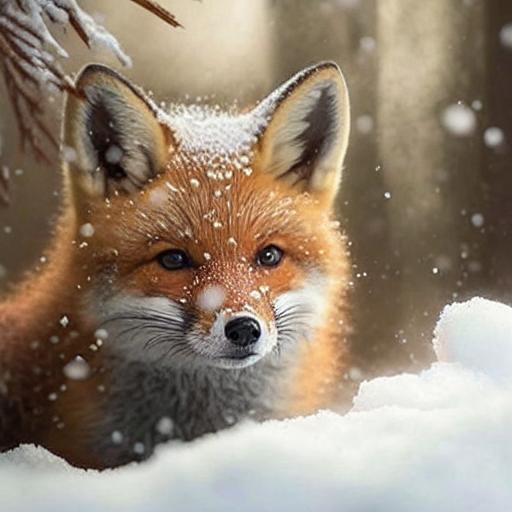}\vspace{3mm}
     \includegraphics[width=\linewidth]{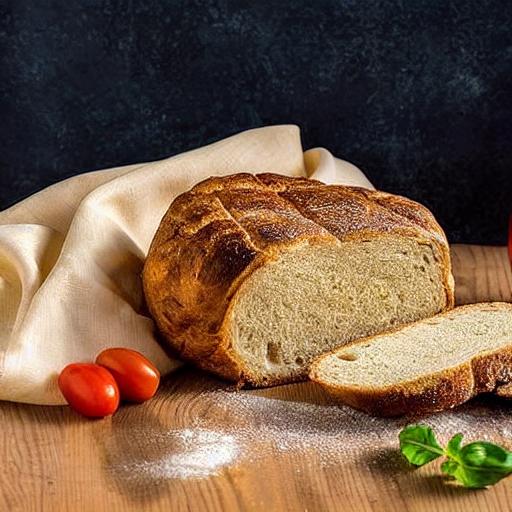}\vspace{3mm}
     \includegraphics[width=\linewidth]{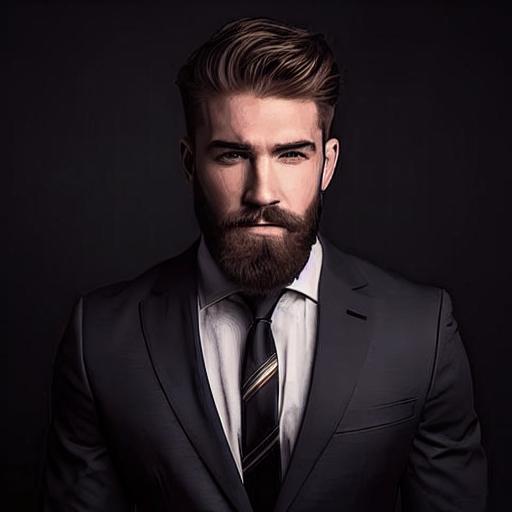}\vspace{3mm}
     \includegraphics[width=\linewidth]{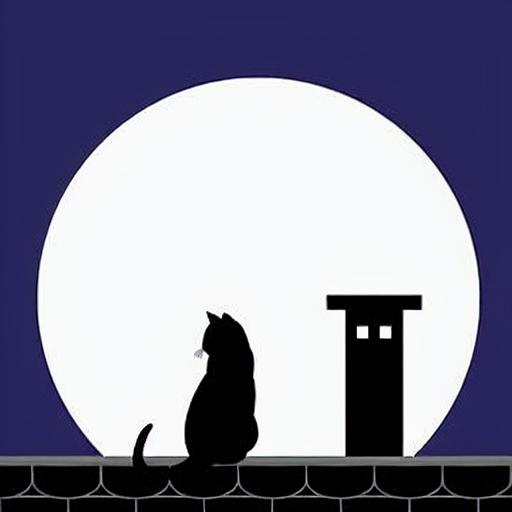}
     \end{minipage}
     }
    \hspace{-2.8mm}
    \subfloat[TODInv]{
     \begin{minipage}{0.0825\linewidth}
     \includegraphics[width=\linewidth]{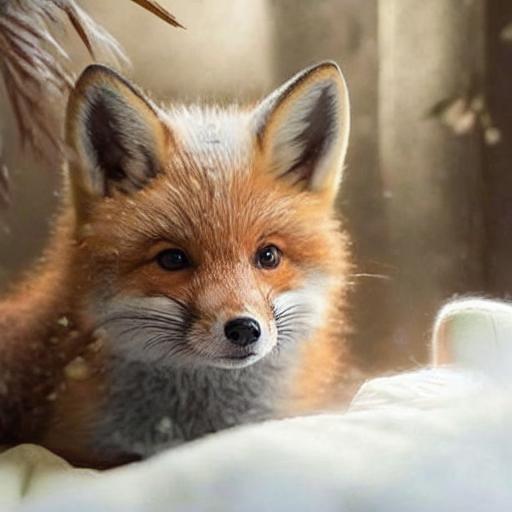}\vspace{3mm}
     \includegraphics[width=\linewidth]{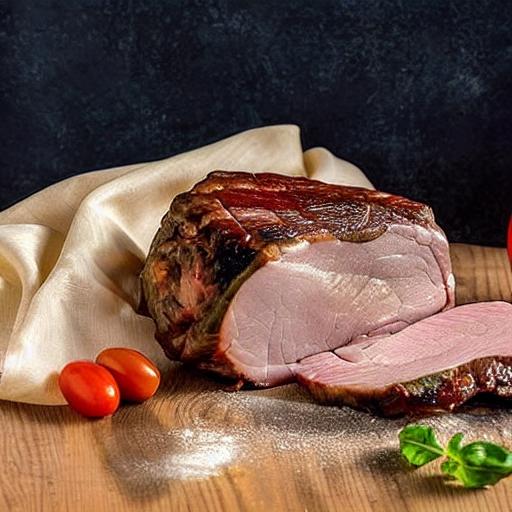}\vspace{3mm}
     \includegraphics[width=\linewidth]{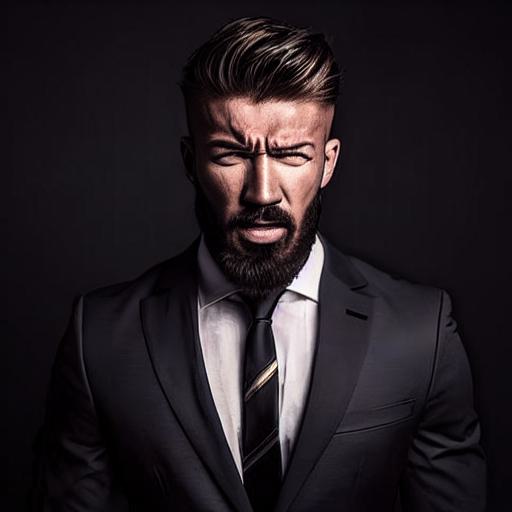}\vspace{3mm}
     \includegraphics[width=\linewidth]{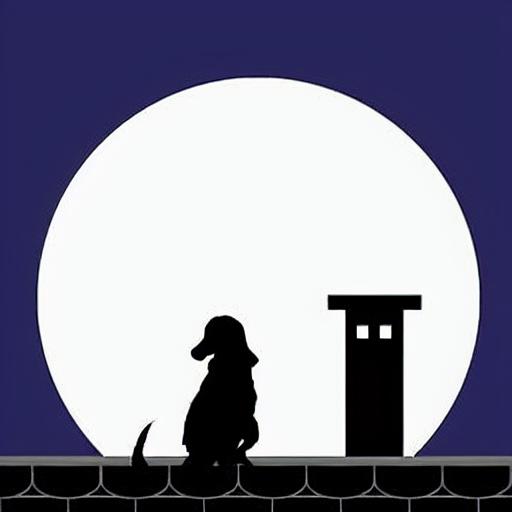}
     \end{minipage}
     }
\vspace{-2mm}
\caption{Qualitative comparison with various inversion methods using P2P editing method.}\vspace{-3mm}
\label{fig:figure_inv}
\end{figure*}

\subsection{Implementation Details}
\label{sec:imp}

We implement the proposed method in PyTorch on a PC with Nvidia GeForce RTX 3090. We use Stable Diffusion V1.4 as our main text-guided diffusion model and set the CFG scale as 7.5. We use the AdamW optimizer~\cite{loshchilov2017decoupled} with the learning rate is set to be 0.001. We categorize 9 editing types in PIE-Bench dataset into three classes. Particularly, the structure editing contains Add Object, Delete Object, Change Content, and Change Pose. The appearance editing contains Change Color, Change Material, and Change Style, and the global editing only contains Change Background. Additionally, the U-Net of diffusion model has 4 resolution layer scales: $64\times64$, $32\times32$, $16\times16$, and $8\times8$. Inspired by~\cite{voynov2023p}, we take the resolutions of $64\times64$ and $32\times32$ as appearance layers, and $16\times16$, $8\times8$ as structure layers. We set the maximization optimization steps \emph{K=10}, and follow~\cite{mokady2023null,li2024source} set threshold $\delta$ as $5e^{-6}$.

\subsection{Quantitative Comparison}

We present the quantitative comparisons with state-of-the-art methods based on various text-guided image editing methods in Tab.~\ref{tab:inversion_based_editing}, we can see that our TODInv outperforms competitors with various editing techniques on most of the evaluation metrics. SPDInv is beyond our method on some reconstruction metrics, but it has a worse editability. As discussed in Sec.\ref{sec:aem}, that is because it optimizes the latent code directly for the faithful reconstruction, but ignores the important editing task, the same conclusion also can be drawn from Fig.~\ref{fig:teaser} and Fig.~\ref{fig:figure_inv}, as it always failed on image editing. Thanks to our task-oriented prompt optimization, our method achieves faithful reconstruction and high editability performance. On the other hand, our method is more efficient than optimization works, because we optimize prompt embedding in the expressive $\mathcal{P^*}$ space, which is easier for optimization.

\subsection{Qualitative Comparison}
The qualitative comparison with various inversion methods based on P2P~\cite{hertz2022prompt} edit can be seen in Fig.~\ref{fig:teaser} and Fig.~\ref{fig:figure_inv}. We can see that the edited images obtained by DDIM always present an inconsistent background or structure with the source images, as pointed out by NTI~\cite{dong2023prompt}, that is aroused by the CFG used in the sampling process.

Besides, all methods fail to replace the ``\textsc{Jacket}'' with ``\textsc{Blouse}'' in $1_{st}$ sample of Fig.~\ref{fig:teaser} except ours, which indicates the effectiveness of our model in object replacement. The same conclusion also can be drawn from the $1_{st}$ sample of Fig.~\ref{fig:figure_inv}, as none of competitors can remove the ``\textsc{Snow}'' on the fox's face. By disentangling the structure and appearance editing in the $\mathcal{P^*}$ space, our method is also skilled at changing the style of images, such as stylizing real images into ``\textsc{Watercolor}''. We notice that SPDInv, AIDI, and FPI fail to replace the ``\textsc{Bread}'' with ``\textsc{Meat}'' in the $2_{nd}$ sample of Fig.~\ref{fig:figure_inv}, that is because all of them optimize the latent code for the faithful reconstruction, but reduces the editability. By minimizing the approximation error in each inversion timestep with a specific layer's prompt optimization, our method not only preserves the source background and structure but also supports various edits. As shown in Fig.~\ref{fig:figure_PNP}, our method presents excellent editability incooperated with PNP editing.
For more qualitative comparisons using other editing methods, please see the supplementary material.


\subsection{Ablation Studies}

\begin{table}[htbp]
\centering%
\caption{{Qualitative comparisons with various variants using P2P editing.}}
\vspace{-4mm}
\small
\centering
\renewcommand\arraystretch{0.01}
\setlength{\tabcolsep}{0.05mm}{
\begin{threeparttable}
\begin{tabular}{c|c|c|c|c|c|c|c|c}
\toprule
\toprule

\multirow{2}{*}{{\textbf{Variant}}}              & \textbf{Structure}          & \multicolumn{4}{c|}{\textbf{Background Preservation}} & \multicolumn{2}{c|}{\textbf{CLIP Similarity}} & \multirow{2}{*}{\textbf{Times(s)} $\downarrow$} \\
\cmidrule{2-8} &\textbf{Distance}$_{^{\times 10^3}}$ $\downarrow$ & \textbf{PSNR} $\uparrow$     & \textbf{LPIPS}$_{^{\times 10^3}}$ $\downarrow$  & \textbf{MSE}$_{^{\times 10^4}}$ $\downarrow$     & \textbf{SSIM}$_{^{\times 10^2}}$ $\uparrow$    & \textbf{Whole}  $\uparrow$          & \textbf{Edited}  $\uparrow$       \\

\midrule
\emph{Opti.} in $\mathcal{P}$         &36.16           & 21.62      &121.03   & 103.34   &78.10    &25.51        &{22.28}     &21.02    \\
\emph{Opti.} in $\mathcal{P}_t$       &35.93           &21.71     & 120.47   &102.25   & 78.15     &25.49      &22.38   &21.02    \\
\emph{Opti.} in $\mathcal{P}$+        &36.40           & 21.63      &120.92   &102.61   &78.12     &25.57      &22.33   &21.02    \\
\midrule
\emph{T=50, K=25}      &8.32           & 28.36     & 40.04  &25.68   &85.92    &{25.47}      &21.89   &29.04    \\
\emph{T=50, K=50}       &8.29           &28.37     & 39.93  &\textbf{25.66}  &85.93     &25.45     &21.89   &45.04   \\

\emph{T=10, K=10}       &25.01             & 23.89       &85.51   &65.81  & 81.28    &\textbf{25.64}     &22.02   &\textbf{6.79}    \\
\emph{T=100, K=10}      & 35.10           &21.70     &119.20   &102.36   &78.29    &25.61      &\textbf{22.26}  &41.23    \\
\midrule
{\emph{w/o} TOPO}        &8.55           &28.18      &41.24  &26.48   &85.80    &{24.46}      &{20.14}   &21.02    \\
\midrule

{\textbf{TODInv}}             &\textbf{8.37}       &\textbf{28.39}  &\textbf{39.86}  &{25.71}         &\textbf{86.04} &{25.47}      & {21.91}      &21.02    \\

\bottomrule
\bottomrule
\end{tabular}
\end{threeparttable}}
\label{tab:ablation}
\vspace{-3mm}
\end{table}

In this section, we conduct an ablation experiment to analyze different choices in our TODInv. We first analyze the effectiveness of optimization in extended prompt space $\mathcal{P}^*$. Particularly, we develop three variants: 1) \emph{Opti.} in $\mathcal{P}$, we optimize the prompt embedding in the original prompt embedding space $\mathcal{P}$, in which all timesteps and layers of U-Net share the same optimized prompt embedding. 2) \emph{Opti.} in $\mathcal{P}_t$, we optimize the prompt only in different timestep, in which all layers of U-Net share the same optimized prompt embedding. 3) \emph{Opti.} in $\mathcal{P}$+, we optimize the prompt only in different layers of U-Net, and all timesteps share the same optimized embeddings. We also conduct an ablation study to investigate the effect of different sampling steps $T$ and optimization steps $K$. We develop two variants with different sampling steps \emph{T}, \emph{T=10}, and \emph{T=100}, with the default optimization steps \emph{K=10}, and develop another two variants with different optimization steps \emph{K=25}, \emph{K=50} with \emph{T=50}. Additionally, for evaluating the effectiveness of our Task-Oriented Optimization by proposing variant \emph{w/o} Task-Oriented Prompt Optimization (TOPO) that optimizes all layers of U-Net regardless of the editing types. We conduct above ablation experiment using P2P editing on the PIE-Bench dataset.

\begin{wrapfigure}{r}{6cm}
    \vspace{-5mm}
    \centering
    \captionsetup[subfloat]{labelformat=empty,justification=centering}
    \subfloat[Source]{
     \begin{minipage}{0.225\linewidth}
     \includegraphics[width=\linewidth]{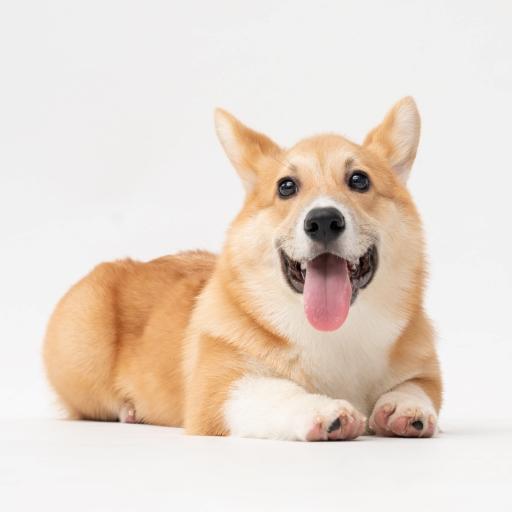}\vspace{3mm}
     \includegraphics[width=\linewidth]{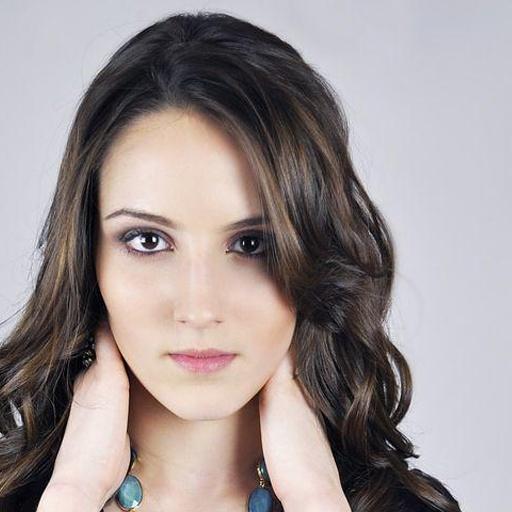}\vspace{3mm}
     \includegraphics[width=\linewidth]{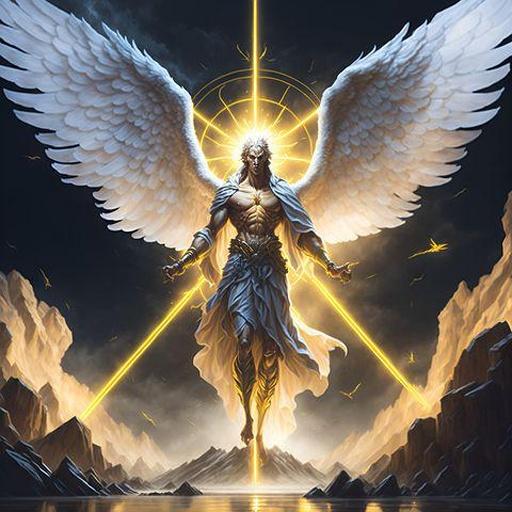}
     \end{minipage}
     }
    \put(-30,70){{\scriptsize{\textsc{A \rev{dog} is laying ... $\rightarrow$ A \rev{lion} is laying ...}}}}
    \hspace{-2.8mm}
    \subfloat[DDIM]{
     \begin{minipage}{0.225\linewidth}
     \includegraphics[width=\linewidth]{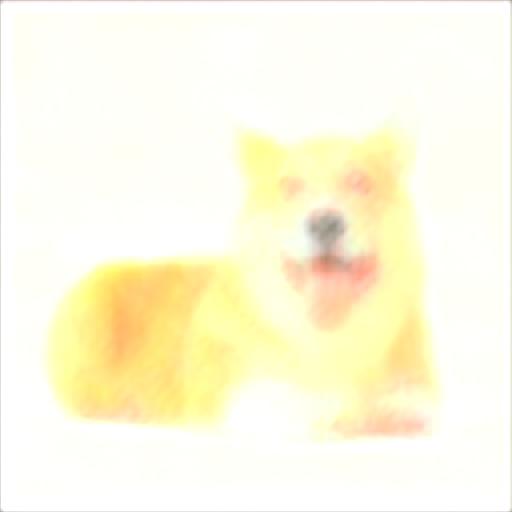}\vspace{3mm}
     \includegraphics[width=\linewidth]{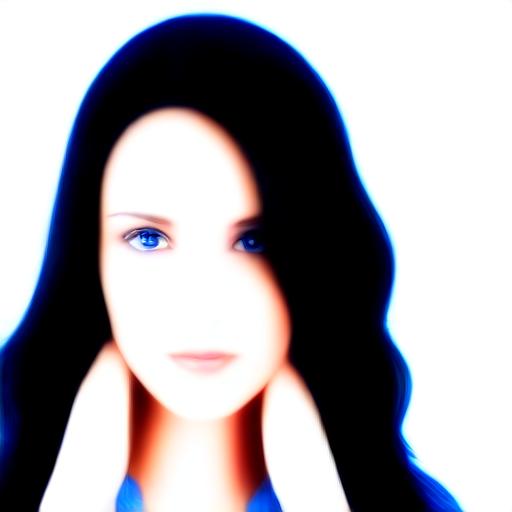}\vspace{3mm}
     \includegraphics[width=\linewidth]{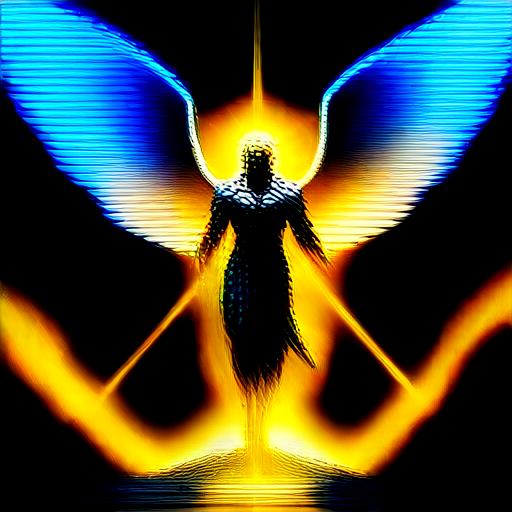}

     \end{minipage}
     }
   \put(-92.5,22.5){{\scriptsize{\textsc{Woman with \rev{brown} hair $\rightarrow$Woman with \rev{blue} hair}}}}
    \hspace{-2.8mm}
    \subfloat[ReNoise]{
     \begin{minipage}{0.225\linewidth}
     \includegraphics[width=\linewidth]{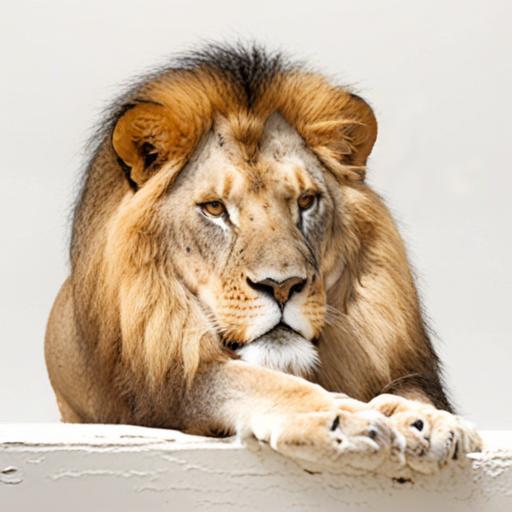}\vspace{3mm}
     \includegraphics[width=\linewidth]{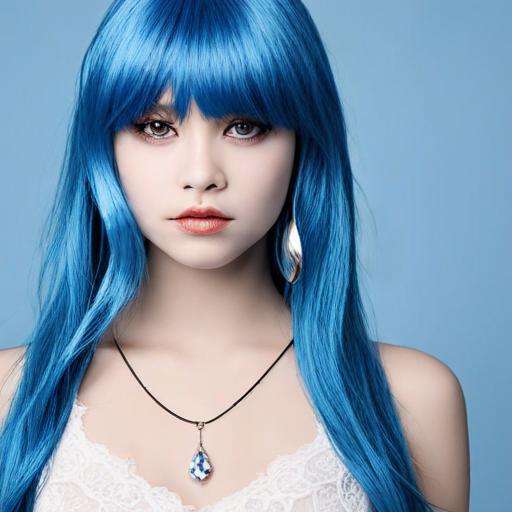}\vspace{3mm}
     \includegraphics[width=\linewidth]{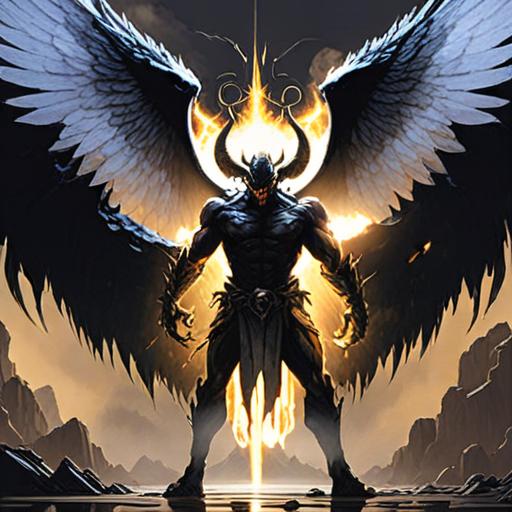}
     \end{minipage}
     }
   \put(-125,-25){\scriptsize{\textsc{An \rev{angel} with wings ... $\rightarrow$An \rev{demon} with wings ...}}}
    \hspace{-2.8mm}
    \subfloat[Ours]{
     \begin{minipage}{0.225\linewidth}
     \includegraphics[width=\linewidth]{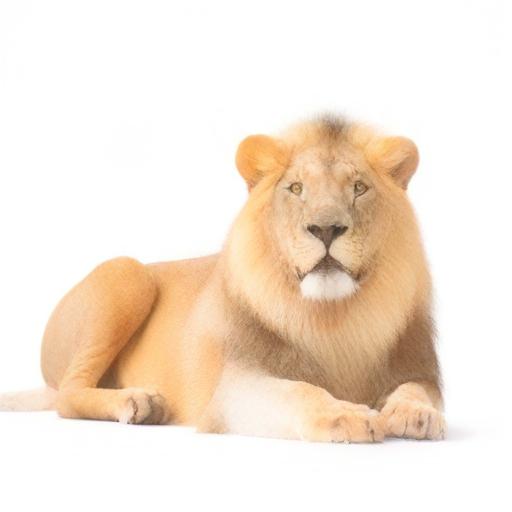}\vspace{3mm}
     \includegraphics[width=\linewidth]{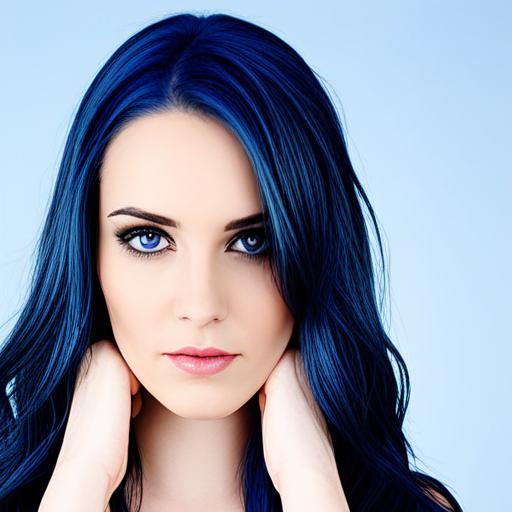}\vspace{3mm}
     \includegraphics[width=\linewidth]{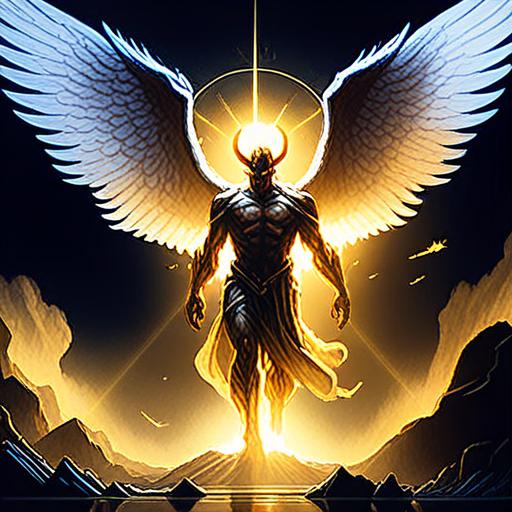}
     \end{minipage}
     }
\vspace{-2mm}
\caption{Qualitative comparison on SDXL-Turbo.}\vspace{-6mm}
\label{fig:figure_sdxl}
\end{wrapfigure}

The quantitative comparison of various variants is presented in Tab.~\ref{tab:ablation}. The variants \emph{Opti.} in $\mathcal{P}$, \emph{Opti.} in $\mathcal{P}{t}$, and \emph{Opti.} in $\mathcal{P}$+ demonstrate worse performance in both structure distance and reconstruction. This suggests that optimizing prompt embeddings in these three spaces does not guarantee faithful reconstruction. Additionally, these variants show higher editability (CLIP Similarity) compared to our TODInv, as the edited images, without the constraint of source images, have more freedom to generate content according to the target prompt. In comparison, our final model, TODInv, outperforms variants \emph{T=50, K=25} and \emph{T=50, K=50} across all metrics, although the latter variants require more processing time. The expressiveness of the $\mathcal{P}^*$ space facilitates more effective minimization of approximation error, and 10 steps are sufficient for this process. Furthermore, both variants \emph{T=10, K=10} and \emph{T=100, K=10} exhibit poorer reconstruction performance. Consequently, we adhere to existing work by setting \emph{T=50}.

Compared with variant \emph{w/o} TOPO, our final method gains the improvement in editability and reconstruction. Our task-oriented prompt optimization reduces the approximation error by optimizing prompt embeddings that are irrelevant to current editing, and achieves better editability without influencing the reconstruction, which evidences the effectiveness of our task-oriented strategy. For the qualitative comparison of various variants, and quantitative comparison on different editing types, please see in the Appendix.

\subsection{Extension on Few-step Diffusion Model}
Besides the Stable Diffusion, We also extend our method on a few-step diffusion model, SDXL-Turbo~\cite{sauer2023adversarial}. We set 4 inference steps for this model, and the optimization steps $K$ is 
set to be 10. We compare our method with DDIM inversion, and ReNoise~\cite{garibi2024renoise} in the bottom rows of Tab.~\ref{tab:inversion_based_editing}. 
Here we set ReNoise with the DDIM sampler for the fair comparison. We can see that our method outperforms DDIM and ReNoise both on the background preservation and CLIP similarity, with the similar inference time cost with ReNoise, which demonstrates our generalization ability on few-step diffusion model. The qualitative comparison are shown in Fig.~\ref{fig:figure_sdxl}, our method captures the source structure effectively.

\section{Conclusion and Limitation}
\label{sec:conclusion}

In this paper, we present TODInv, a framework that inverts and edits a real image using diffusion models tailored to specific editing tasks. We categorize various editing tasks into three types, for each kind of editing, we minimize the approximation error by optimizing specific prompt embeddings that are irrelevant to the current editing, achieving both faithful reconstruction and high editability. We conducted experiments on Stable Diffusion and SDXL-Turbo models, demonstrating the effectiveness of our TODInv over state-of-the-art methods. The primary limitation of TODInv is that it requires determining the editing types prior to inversion. However, this can be addressed by using a large language model to easily determine the types. Please refer to the Appendix for detailed instructions.

\bibliography{egbib}

\begin{thebibliography}{60}
\providecommand{\natexlab}[1]{#1}
\providecommand{\url}[1]{\texttt{#1}}
\expandafter\ifx\csname urlstyle\endcsname\relax
  \providecommand{\doi}[1]{doi: #1}\else
  \providecommand{\doi}{doi: \begingroup \urlstyle{rm}\Url}\fi

\bibitem[Abdal et~al.(2019)Abdal, Qin, and Wonka]{abdal2019image2stylegan}
Rameen Abdal, Yipeng Qin, and Peter Wonka.
\newblock Image2stylegan: How to embed images into the stylegan latent space?
\newblock In \emph{CVPR}, pp.\  4432--4441, 2019.

\bibitem[Abdal et~al.(2020)Abdal, Qin, and Wonka]{abdal2020image2styleganpp}
Rameen Abdal, Yipeng Qin, and Peter Wonka.
\newblock Image2stylegan++: How to edit the embedded images?
\newblock In \emph{CVPR}, pp.\  8296--8305, 2020.

\bibitem[Alaluf et~al.(2023)Alaluf, Richardson, Metzer, and
  Cohen-Or]{alaluf2023neural}
Yuval Alaluf, Elad Richardson, Gal Metzer, and Daniel Cohen-Or.
\newblock A neural space-time representation for text-to-image personalization.
\newblock \emph{ACM TOG}, 42\penalty0 (6):\penalty0 1--10, 2023.

\bibitem[Baranchuk et~al.(2022)Baranchuk, Rubachev, Voynov, Khrulkov, and
  Babenko]{baranchuk2021label}
Dmitry Baranchuk, Ivan Rubachev, Andrey Voynov, Valentin Khrulkov, and Artem
  Babenko.
\newblock Label-efficient semantic segmentation with diffusion models.
\newblock In \emph{ICLR}, 2022.

\bibitem[Cao et~al.(2023)Cao, Wang, Qi, Shan, Qie, and Zheng]{cao2023masactrl}
Mingdeng Cao, Xintao Wang, Zhongang Qi, Ying Shan, Xiaohu Qie, and Yinqiang
  Zheng.
\newblock Masactrl: Tuning-free mutual self-attention control for consistent
  image synthesis and editing.
\newblock In \emph{ICCV}, pp.\  22560--22570, 2023.

\bibitem[Caron et~al.(2021)Caron, Touvron, Misra, J{\'e}gou, Mairal,
  Bojanowski, and Joulin]{caron2021emerging}
Mathilde Caron, Hugo Touvron, Ishan Misra, Herv{\'e} J{\'e}gou, Julien Mairal,
  Piotr Bojanowski, and Armand Joulin.
\newblock Emerging properties in self-supervised vision transformers.
\newblock In \emph{CVPR}, pp.\  9650--9660, 2021.

\bibitem[Chai et~al.(2023)Chai, Guo, Wang, and Lu]{chai2023stablevideo}
Wenhao Chai, Xun Guo, Gaoang Wang, and Yan Lu.
\newblock Stablevideo: Text-driven consistency-aware diffusion video editing.
\newblock In \emph{ICCV}, pp.\  23040--23050, 2023.

\bibitem[Chen et~al.(2023)Chen, Sun, Song, and Luo]{chen2022diffusiondet}
Shoufa Chen, Peize Sun, Yibing Song, and Ping Luo.
\newblock Diffusiondet: Diffusion model for object detection.
\newblock In \emph{ICCV}, pp.\  19830--19843, 2023.

\bibitem[Cho et~al.(2024)Cho, Lee, Kim, Oh, and Jeong]{cho2024noise}
Hansam Cho, Jonghyun Lee, Seoung~Bum Kim, Tae-Hyun Oh, and Yonghyun Jeong.
\newblock Noise map guidance: Inversion with spatial context for real image
  editing.
\newblock In \emph{ICLR}, 2024.

\bibitem[Creswell \& Bharath(2018)Creswell and Bharath]{creswell2018inverting}
Antonia Creswell and Anil~Anthony Bharath.
\newblock Inverting the generator of a generative adversarial network.
\newblock \emph{IEEE TNNLS}, 30\penalty0 (7):\penalty0 1967--1974, 2018.

\bibitem[Dhariwal \& Nichol(2021)Dhariwal and Nichol]{dhariwal2021diffusion}
Prafulla Dhariwal and Alexander Nichol.
\newblock Diffusion models beat gans on image synthesis.
\newblock In \emph{NeurIPS}, pp.\  8780--8794, 2021.

\bibitem[Dong et~al.(2023)Dong, Xue, Duan, and Han]{dong2023prompt}
Wenkai Dong, Song Xue, Xiaoyue Duan, and Shumin Han.
\newblock Prompt tuning inversion for text-driven image editing using diffusion
  models.
\newblock In \emph{ICCV}, pp.\  7430--7440, 2023.

\bibitem[Garibi et~al.(2024)Garibi, Patashnik, Voynov, Averbuch-Elor, and
  Cohen-Or]{garibi2024renoise}
Daniel Garibi, Or~Patashnik, Andrey Voynov, Hadar Averbuch-Elor, and Daniel
  Cohen-Or.
\newblock Renoise: Real image inversion through iterative noising.
\newblock In \emph{ECCV}, 2024.

\bibitem[Geyer et~al.(2024)Geyer, Bar-Tal, Bagon, and Dekel]{tokenflow2023}
Michal Geyer, Omer Bar-Tal, Shai Bagon, and Tali Dekel.
\newblock Tokenflow: Consistent diffusion features for consistent video
  editing.
\newblock In \emph{ICLR}, 2024.

\bibitem[Han et~al.(2024)Han, Wen, Chen, Zhang, Song, Ren, Gao, Stathopoulos,
  He, Chen, et~al.]{han2024proxedit}
Ligong Han, Song Wen, Qi~Chen, Zhixing Zhang, Kunpeng Song, Mengwei Ren,
  Ruijiang Gao, Anastasis Stathopoulos, Xiaoxiao He, Yuxiao Chen, et~al.
\newblock Proxedit: Improving tuning-free real image editing with proximal
  guidance.
\newblock In \emph{WACV}, pp.\  4291--4301, 2024.

\bibitem[Hertz et~al.(2023)Hertz, Mokady, Tenenbaum, Aberman, Pritch, and
  Cohen-Or]{hertz2022prompt}
Amir Hertz, Ron Mokady, Jay Tenenbaum, Kfir Aberman, Yael Pritch, and Daniel
  Cohen-Or.
\newblock Prompt-to-prompt image editing with cross attention control.
\newblock In \emph{ICLR}, 2023.

\bibitem[Ho \& Salimans(2022)Ho and Salimans]{ho2022classifier}
Jonathan Ho and Tim Salimans.
\newblock Classifier-free diffusion guidance.
\newblock In \emph{NeurIPS Workshop}, 2022.

\bibitem[Huberman-Spiegelglas et~al.(2024)Huberman-Spiegelglas, Kulikov, and
  Michaeli]{huberman2023edit}
Inbar Huberman-Spiegelglas, Vladimir Kulikov, and Tomer Michaeli.
\newblock An edit friendly ddpm noise space: Inversion and manipulations.
\newblock In \emph{CVPR}, 2024.

\bibitem[Ji et~al.(2023)Ji, Chen, Xie, Hong, Liu, Liu, Lu, Li, and
  Luo]{ji2023ddp}
Yuanfeng Ji, Zhe Chen, Enze Xie, Lanqing Hong, Xihui Liu, Zhaoqiang Liu, Tong
  Lu, Zhenguo Li, and Ping Luo.
\newblock Ddp: Diffusion model for dense visual prediction.
\newblock In \emph{ICCV}, pp.\  21741--21752, 2023.

\bibitem[Ju(2023)]{PnPInversion}
Xuan Ju.
\newblock Pnpinversion, 2023.
\newblock URL \url{https://github.com/cure-lab/PnPInversion}.

\bibitem[Ju et~al.(2024)Ju, Zeng, Bian, Liu, and Xu]{ju2023direct}
Xuan Ju, Ailing Zeng, Yuxuan Bian, Shaoteng Liu, and Qiang Xu.
\newblock Pnp inversion: Boosting diffusion-based editing with 3 lines of code.
\newblock In \emph{ICLR}, 2024.

\bibitem[Khachatryan et~al.(2023)Khachatryan, Movsisyan, Tadevosyan, Henschel,
  Wang, Navasardyan, and Shi]{text2video-zero}
Levon Khachatryan, Andranik Movsisyan, Vahram Tadevosyan, Roberto Henschel,
  Zhangyang Wang, Shant Navasardyan, and Humphrey Shi.
\newblock Text2video-zero: Text-to-image diffusion models are zero-shot video
  generators.
\newblock In \emph{ICCV}, pp.\  15954--15964, 2023.

\bibitem[Li et~al.(2024)Li, Li, Guo, and Zhang]{li2024source}
Ruibin Li, Ruihuang Li, Song Guo, and Lei Zhang.
\newblock Source prompt disentangled inversion for boosting image editability
  with diffusion models.
\newblock In \emph{ECCV}, 2024.

\bibitem[Liew et~al.(2022)Liew, Yan, Zhou, and Feng]{liew2022magicmix}
Jun~Hao Liew, Hanshu Yan, Daquan Zhou, and Jiashi Feng.
\newblock Magicmix: Semantic mixing with diffusion models.
\newblock \emph{arXiv preprint arXiv:2210.16056}, 2022.

\bibitem[Liu et~al.(2024)Liu, Zhang, Li, Lin, and Jia]{liu2023videop2p}
Shaoteng Liu, Yuechen Zhang, Wenbo Li, Zhe Lin, and Jiaya Jia.
\newblock Video-p2p: Video editing with cross-attention control.
\newblock In \emph{CVPR}, 2024.

\bibitem[Loshchilov \& Hutter(2019)Loshchilov and
  Hutter]{loshchilov2017decoupled}
Ilya Loshchilov and Frank Hutter.
\newblock Decoupled weight decay regularization.
\newblock In \emph{ICLR}, 2019.

\bibitem[Luo et~al.(2023{\natexlab{a}})Luo, Tan, Huang, Li, and
  Zhao]{luo2023latent}
Simian Luo, Yiqin Tan, Longbo Huang, Jian Li, and Hang Zhao.
\newblock Latent consistency models: Synthesizing high-resolution images with
  few-step inference.
\newblock \emph{arXiv preprint arXiv:2310.04378}, 2023{\natexlab{a}}.

\bibitem[Luo et~al.(2023{\natexlab{b}})Luo, Tan, Patil, Gu, von Platen, Passos,
  Huang, Li, and Zhao]{luo2023lcm}
Simian Luo, Yiqin Tan, Suraj Patil, Daniel Gu, Patrick von Platen,
  Apolin{\'a}rio Passos, Longbo Huang, Jian Li, and Hang Zhao.
\newblock Lcm-lora: A universal stable-diffusion acceleration module.
\newblock \emph{arXiv preprint arXiv:2311.05556}, 2023{\natexlab{b}}.

\bibitem[Meiri et~al.(2023)Meiri, Samuel, Darshan, Chechik, Avidan, and
  Ben-Ari]{meiri2023fixed}
Barak Meiri, Dvir Samuel, Nir Darshan, Gal Chechik, Shai Avidan, and Rami
  Ben-Ari.
\newblock Fixed-point inversion for text-to-image diffusion models.
\newblock \emph{arXiv preprint arXiv:2312.12540}, 2023.

\bibitem[Miyake et~al.(2023)Miyake, Iohara, Saito, and
  Tanaka]{miyake2023negative}
Daiki Miyake, Akihiro Iohara, Yu~Saito, and Toshiyuki Tanaka.
\newblock Negative-prompt inversion: Fast image inversion for editing with
  text-guided diffusion models.
\newblock \emph{arXiv preprint arXiv:2305.16807}, 2023.

\bibitem[Mokady et~al.(2023)Mokady, Hertz, Aberman, Pritch, and
  Cohen-Or]{mokady2023null}
Ron Mokady, Amir Hertz, Kfir Aberman, Yael Pritch, and Daniel Cohen-Or.
\newblock Null-text inversion for editing real images using guided diffusion
  models.
\newblock In \emph{CVPR}, pp.\  6038--6047, 2023.

\bibitem[Pan et~al.(2023)Pan, Gherardi, Xie, and Huang]{pan2023effective}
Zhihong Pan, Riccardo Gherardi, Xiufeng Xie, and Stephen Huang.
\newblock Effective real image editing with accelerated iterative diffusion
  inversion.
\newblock In \emph{ICCV}, pp.\  15912--15921, 2023.

\bibitem[Parmar et~al.(2023)Parmar, Kumar~Singh, Zhang, Li, Lu, and
  Zhu]{parmar2023zero}
Gaurav Parmar, Krishna Kumar~Singh, Richard Zhang, Yijun Li, Jingwan Lu, and
  Jun-Yan Zhu.
\newblock Zero-shot image-to-image translation.
\newblock In \emph{SIGGRAPH}, pp.\  1--11, 2023.

\bibitem[Patashnik et~al.(2023)Patashnik, Garibi, Azuri, Averbuch-Elor, and
  Cohen-Or]{patashnik2023localizing}
Or~Patashnik, Daniel Garibi, Idan Azuri, Hadar Averbuch-Elor, and Daniel
  Cohen-Or.
\newblock Localizing object-level shape variations with text-to-image diffusion
  models.
\newblock In \emph{ICCV}, pp.\  23051--23061, 2023.

\bibitem[Qi et~al.(2023)Qi, Cun, Zhang, Lei, Wang, Shan, and
  Chen]{qi2023fatezero}
Chenyang Qi, Xiaodong Cun, Yong Zhang, Chenyang Lei, Xintao Wang, Ying Shan,
  and Qifeng Chen.
\newblock Fatezero: Fusing attentions for zero-shot text-based video editing.
\newblock In \emph{ICCV}, pp.\  15932--15942, 2023.

\bibitem[Ramesh et~al.(2022)Ramesh, Dhariwal, Nichol, Chu, and
  Chen]{ramesh2022hierarchical}
Aditya Ramesh, Prafulla Dhariwal, Alex Nichol, Casey Chu, and Mark Chen.
\newblock Hierarchical text-conditional image generation with clip latents.
\newblock \emph{arXiv preprint arXiv:2204.06125}, 1\penalty0 (2):\penalty0 3,
  2022.

\bibitem[Rombach et~al.(2022)Rombach, Blattmann, Lorenz, Esser, and
  Ommer]{rombach2022high}
Robin Rombach, Andreas Blattmann, Dominik Lorenz, Patrick Esser, and Bj{\"o}rn
  Ommer.
\newblock High-resolution image synthesis with latent diffusion models.
\newblock In \emph{CVPR}, pp.\  10684--10695, 2022.

\bibitem[Ruiz et~al.(2023)Ruiz, Li, Jampani, Pritch, Rubinstein, and
  Aberman]{ruiz2023dreambooth}
Nataniel Ruiz, Yuanzhen Li, Varun Jampani, Yael Pritch, Michael Rubinstein, and
  Kfir Aberman.
\newblock Dreambooth: Fine tuning text-to-image diffusion models for
  subject-driven generation.
\newblock In \emph{CVPR}, pp.\  22500--22510, 2023.

\bibitem[Saharia et~al.(2022)Saharia, Chan, Saxena, Li, Whang, Denton,
  Ghasemipour, Gontijo~Lopes, Karagol~Ayan, Salimans,
  et~al.]{saharia2022photorealistic}
Chitwan Saharia, William Chan, Saurabh Saxena, Lala Li, Jay Whang, Emily~L
  Denton, Kamyar Ghasemipour, Raphael Gontijo~Lopes, Burcu Karagol~Ayan, Tim
  Salimans, et~al.
\newblock Photorealistic text-to-image diffusion models with deep language
  understanding.
\newblock In \emph{NeurIPS}, volume~35, pp.\  36479--36494, 2022.

\bibitem[Sauer et~al.(2023)Sauer, Lorenz, Blattmann, and
  Rombach]{sauer2023adversarial}
Axel Sauer, Dominik Lorenz, Andreas Blattmann, and Robin Rombach.
\newblock Adversarial diffusion distillation.
\newblock \emph{arXiv preprint arXiv:2311.17042}, 2023.

\bibitem[Sohl-Dickstein et~al.(2015)Sohl-Dickstein, Weiss, Maheswaranathan, and
  Ganguli]{sohl2015deep}
Jascha Sohl-Dickstein, Eric Weiss, Niru Maheswaranathan, and Surya Ganguli.
\newblock Deep unsupervised learning using nonequilibrium thermodynamics.
\newblock In \emph{ICML}, pp.\  2256--2265, 2015.

\bibitem[Song et~al.(2021)Song, Meng, and Ermon]{song2021denoising}
Jiaming Song, Chenlin Meng, and Stefano Ermon.
\newblock Denoising diffusion implicit models.
\newblock In \emph{ICLR}, 2021.

\bibitem[Song et~al.(2023)Song, Dhariwal, Chen, and
  Sutskever]{song2023consistency}
Yang Song, Prafulla Dhariwal, Mark Chen, and Ilya Sutskever.
\newblock Consistency models.
\newblock In \emph{ICML}, 2023.

\bibitem[Tumanyan et~al.(2023)Tumanyan, Geyer, Bagon, and
  Dekel]{tumanyan2023plug}
Narek Tumanyan, Michal Geyer, Shai Bagon, and Tali Dekel.
\newblock Plug-and-play diffusion features for text-driven image-to-image
  translation.
\newblock In \emph{CVPR}, pp.\  1921--1930, 2023.

\bibitem[Voynov et~al.(2023)Voynov, Chu, Cohen-Or, and Aberman]{voynov2023p}
Andrey Voynov, Qinghao Chu, Daniel Cohen-Or, and Kfir Aberman.
\newblock $ p+ $: Extended textual conditioning in text-to-image generation.
\newblock \emph{arXiv preprint arXiv:2303.09522}, 2023.

\bibitem[Wallace et~al.(2023)Wallace, Gokul, and Naik]{wallace2023edict}
Bram Wallace, Akash Gokul, and Nikhil Naik.
\newblock Edict: Exact diffusion inversion via coupled transformations.
\newblock In \emph{CVPR}, pp.\  22532--22541, 2023.

\bibitem[Wang et~al.(2004)Wang, Bovik, Sheikh, and Simoncelli]{wang2004image}
Zhou Wang, Alan~C Bovik, Hamid~R Sheikh, and Eero~P Simoncelli.
\newblock Image quality assessment: from error visibility to structural
  similarity.
\newblock \emph{IEEE TIP}, 13\penalty0 (4):\penalty0 600--612, 2004.

\bibitem[Wu \& De~la Torre(2023)Wu and De~la Torre]{wu2023latent}
Chen~Henry Wu and Fernando De~la Torre.
\newblock A latent space of stochastic diffusion models for zero-shot image
  editing and guidance.
\newblock In \emph{ICCV}, pp.\  7378--7387, 2023.

\bibitem[Wu et~al.(2021)Wu, Huang, Zhang, Li, Ji, Yang, Sapiro, and
  Duan]{wu2021godiva}
Chenfei Wu, Lun Huang, Qianxi Zhang, Binyang Li, Lei Ji, Fan Yang, Guillermo
  Sapiro, and Nan Duan.
\newblock Godiva: Generating open-domain videos from natural descriptions.
\newblock \emph{arXiv preprint arXiv:2104.14806}, 2021.

\bibitem[Wu et~al.(2023)Wu, Ge, Wang, Lei, Gu, Shi, Hsu, Shan, Qie, and
  Shou]{wu2023tune}
Jay~Zhangjie Wu, Yixiao Ge, Xintao Wang, Stan~Weixian Lei, Yuchao Gu, Yufei
  Shi, Wynne Hsu, Ying Shan, Xiaohu Qie, and Mike~Zheng Shou.
\newblock Tune-a-video: One-shot tuning of image diffusion models for
  text-to-video generation.
\newblock In \emph{ICCV}, pp.\  7623--7633, 2023.

\bibitem[Xia et~al.(2023)Xia, Zhang, Yang, Xue, Zhou, and Yang]{xia2021gan}
Weihao Xia, Yulun Zhang, Yujiu Yang, Jing-Hao Xue, Bolei Zhou, and Ming-Hsuan
  Yang.
\newblock Gan inversion: A survey.
\newblock \emph{IEEE TPAMI}, 45\penalty0 (3):\penalty0 3121--3138, 2023.

\bibitem[Xu et~al.(2024)Xu, Xu, Zhang, Xu, and He]{xu2024dreamanime}
Chenshu Xu, Yangyang Xu, Huaidong Zhang, Xuemiao Xu, and Shengfeng He.
\newblock Dreamanime: Learning style-identity textual disentanglement for anime
  and beyond.
\newblock 2024.

\bibitem[Xu et~al.(2021)Xu, Du, Xiao, Xu, and He]{Xu2021ICCV}
Yangyang Xu, Yong Du, Wenpeng Xiao, Xuemiao Xu, and Shengfeng He.
\newblock From continuity to editability: Inverting gans with consecutive
  images.
\newblock In \emph{ICCV}, pp.\  13910--13918, 2021.

\bibitem[Xu et~al.(2023)Xu, He, Wong, and Luo]{xu2023rigid}
Yangyang Xu, Shengfeng He, Kwan-Yee~K Wong, and Ping Luo.
\newblock Rigid: Recurrent gan inversion and editing of real face videos.
\newblock In \emph{ICCV}, pp.\  13691--13701, 2023.

\bibitem[Xue et~al.(2024)Xue, Song, Guo, Liu, Zong, Liu, and
  Luo]{xue2024raphael}
Zeyue Xue, Guanglu Song, Qiushan Guo, Boxiao Liu, Zhuofan Zong, Yu~Liu, and
  Ping Luo.
\newblock Raphael: Text-to-image generation via large mixture of diffusion
  paths.
\newblock In \emph{NeurIPS}, volume~36, 2024.

\bibitem[Zhang et~al.(2023{\natexlab{a}})Zhang, Lewis, and
  Kleijn]{zhang2023exact}
Guoqiang Zhang, Jonathan~P Lewis, and W~Bastiaan Kleijn.
\newblock Exact diffusion inversion via bi-directional integration
  approximation.
\newblock \emph{arXiv preprint arXiv:2307.10829}, 2023{\natexlab{a}}.

\bibitem[Zhang et~al.(2018)Zhang, Isola, Efros, Shechtman, and
  Wang]{zhang2018perceptual}
Richard Zhang, Phillip Isola, Alexei~A Efros, Eli Shechtman, and Oliver Wang.
\newblock The unreasonable effectiveness of deep features as a perceptual
  metric.
\newblock In \emph{CVPR}, 2018.

\bibitem[Zhang et~al.(2024)Zhang, Wei, Jiang, Zhang, Zuo, and
  Tian]{zhang2023controlvideo}
Yabo Zhang, Yuxiang Wei, Dongsheng Jiang, Xiaopeng Zhang, Wangmeng Zuo, and
  Qi~Tian.
\newblock Controlvideo: Training-free controllable text-to-video generation.
\newblock 2024.

\bibitem[Zhang et~al.(2023{\natexlab{b}})Zhang, Dong, Tang, Huang, Huang, Ma,
  Lee, Deussen, and Xu]{zhang2023prospect}
Yuxin Zhang, Weiming Dong, Fan Tang, Nisha Huang, Haibin Huang, Chongyang Ma,
  Tong-Yee Lee, Oliver Deussen, and Changsheng Xu.
\newblock Prospect: Prompt spectrum for attribute-aware personalization of
  diffusion models.
\newblock \emph{ACM TOG}, 42\penalty0 (6):\penalty0 1--14, 2023{\natexlab{b}}.

\bibitem[Zhao et~al.(2023)Zhao, Rao, Liu, Liu, Zhou, and
  Lu]{zhao2023unleashing}
Wenliang Zhao, Yongming Rao, Zuyan Liu, Benlin Liu, Jie Zhou, and Jiwen Lu.
\newblock Unleashing text-to-image diffusion models for visual perception.
\newblock In \emph{ICCV}, pp.\  5729--5739, 2023.

\end{thebibliography}
\bibliographystyle{iclr2025_conference}

\newpage
\appendix
\section{Appendix}

\subsection{Qualitative Comparison of different variants}

\begin{figure*}[h]
    \centering
    \captionsetup[subfloat]{labelformat=empty,justification=centering}
    \subfloat[Source]{
     \begin{minipage}{0.1\linewidth}
     \includegraphics[width=\linewidth]{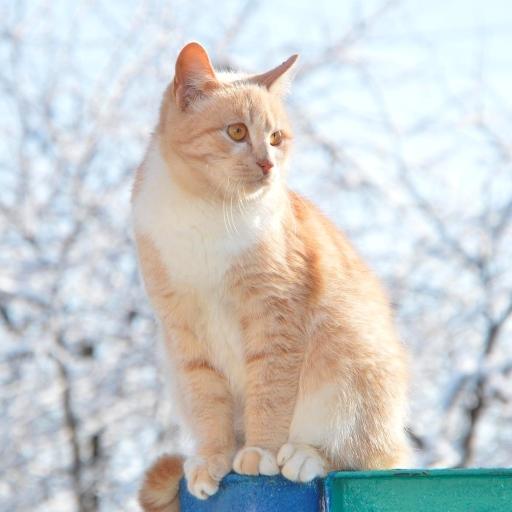}\vspace{3mm}
     \includegraphics[width=\linewidth]{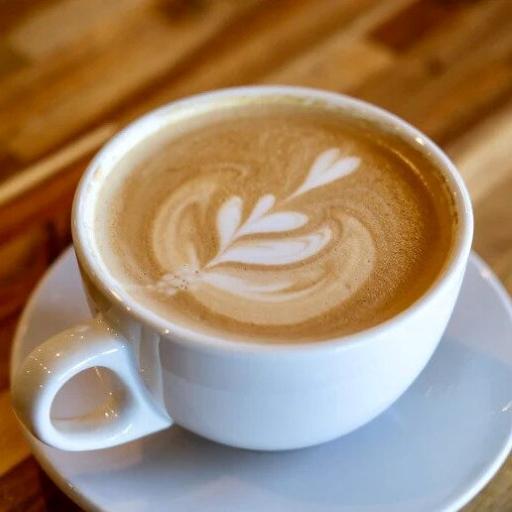}\vspace{3mm}
     \includegraphics[width=\linewidth]{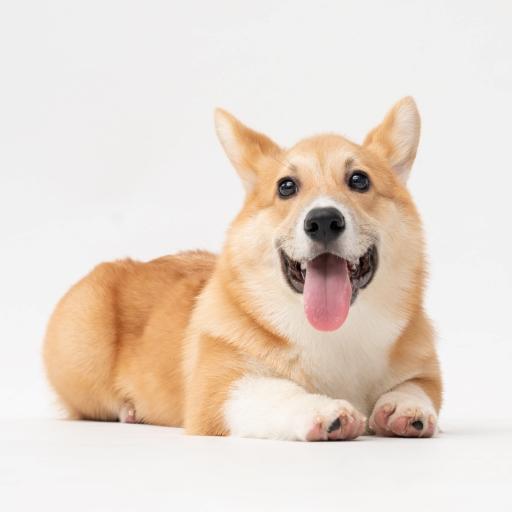}\vspace{3mm}
     \includegraphics[width=\linewidth]{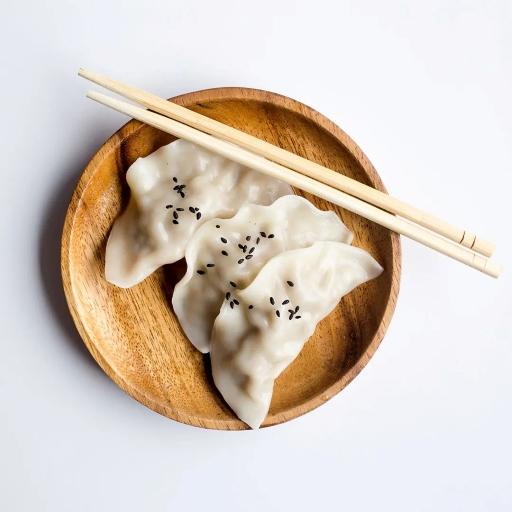}\vspace{3mm}
     \includegraphics[width=\linewidth]{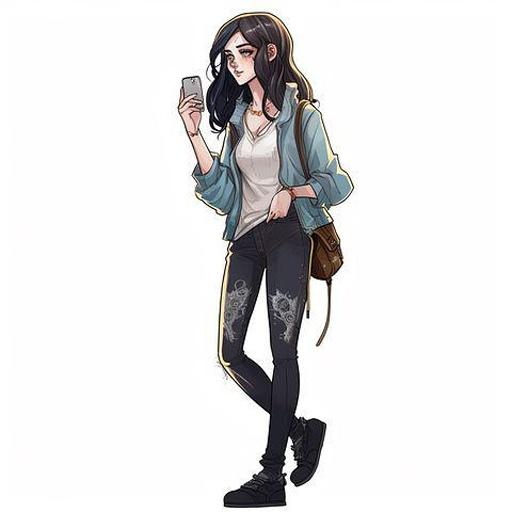}\vspace{3mm}
     \includegraphics[width=\linewidth]{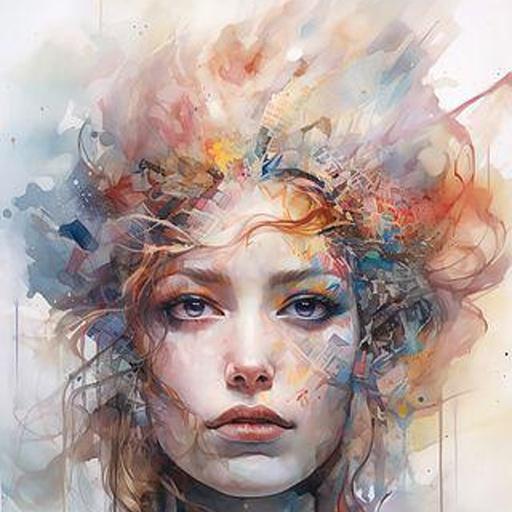}
     \end{minipage}
     }
     \put(12,147){{\scriptsize{\textsc{An \rev{orange} cat sitting on top of a fence $\rightarrow$ An \rev{Black} cat sitting on top of a fence}}}}
     \put(20,97.5){{\scriptsize{\textsc{a cup of coffee with drawing of \rev{tulip} $\rightarrow$ a cup of coffee with drawing of \rev{lion}}}}}
     \put(-5,48){{\scriptsize{\textsc{a \rev{dog} is laying down on a white background $\rightarrow$ a \rev{lion} is laying down on a white background}}}}
     \put(15,-0.5){{\scriptsize{\textsc{three white \rev{dumplings} on brown bowl $\rightarrow$ three white \rev{sushi} on brown bowl}}}}
     \put(-55,-49){{\scriptsize{\textsc{a girl with black hair and a white shirt is holding a \rev{phone} $\rightarrow$ a girl with black hair and a white shirt is holding a \rev{coffee}}}}}
     \put(-40,-99){{\scriptsize{\textsc{a painting of a woman with \rev{colorful paint} on her face $\rightarrow$ a painting of a woman with \rev{drab paint} on her face}}}}
    \hspace{-2.8mm}
    \subfloat[\emph{Opti.} \\ in $\mathcal{P}$]{
     \begin{minipage}{0.1\linewidth}
     \includegraphics[width=\linewidth]{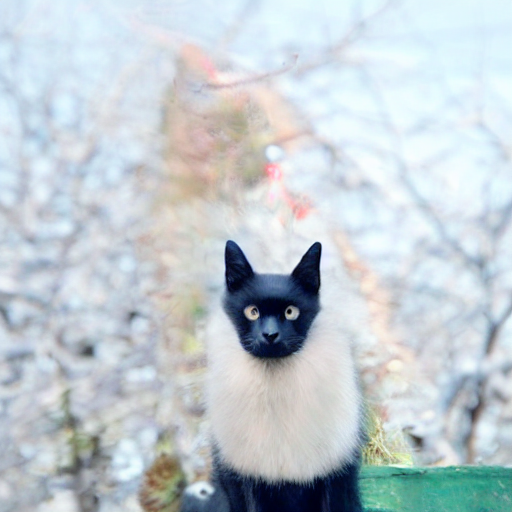}\vspace{3mm}
     \includegraphics[width=\linewidth]{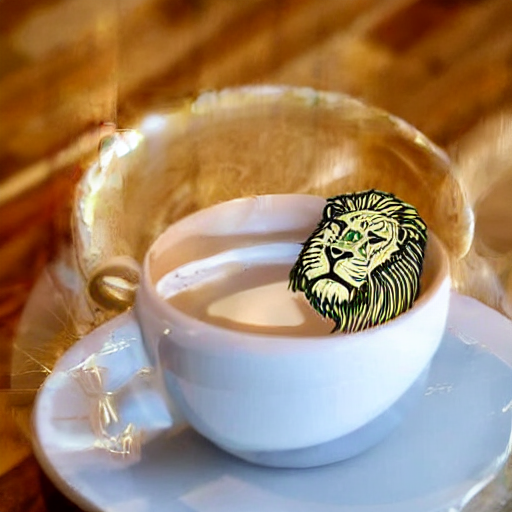}\vspace{3mm}
     \includegraphics[width=\linewidth]{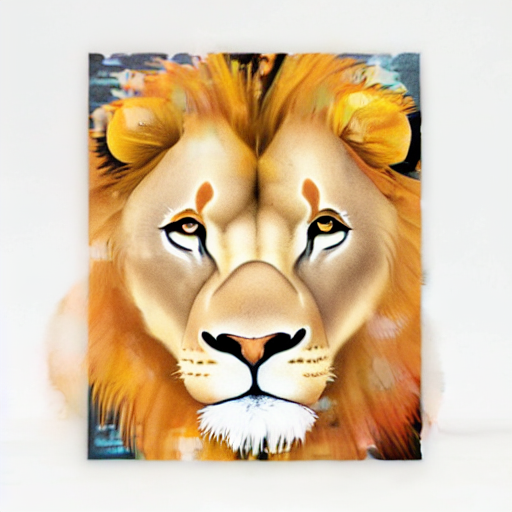}\vspace{3mm}
     \includegraphics[width=\linewidth]{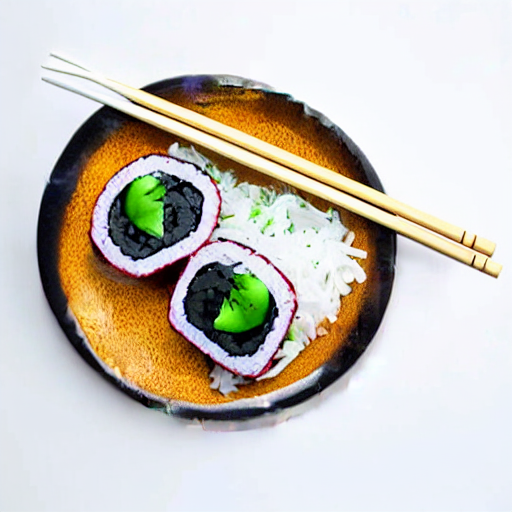}\vspace{3mm}
     \includegraphics[width=\linewidth]{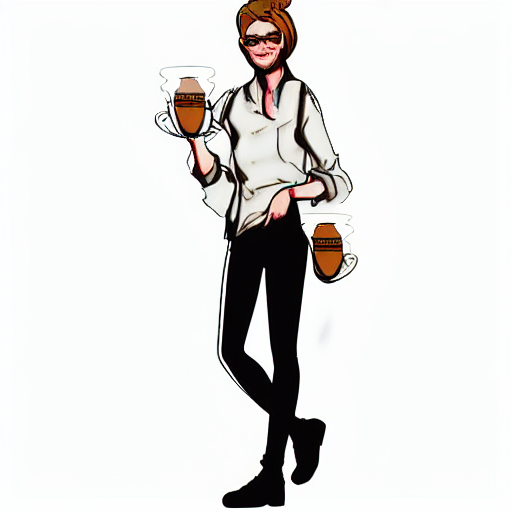}\vspace{3mm}
     \includegraphics[width=\linewidth]{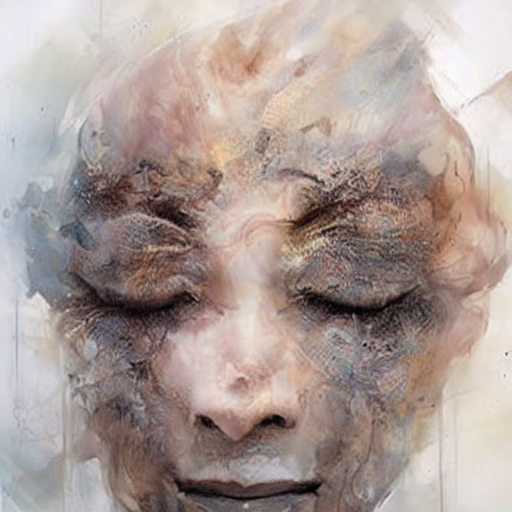}
     \end{minipage}
     }
    \hspace{-2.8mm}
    \subfloat[\emph{Opti.} \\ in $\mathcal{P}_t$]{
     \begin{minipage}{0.1\linewidth}
     \includegraphics[width=\linewidth]{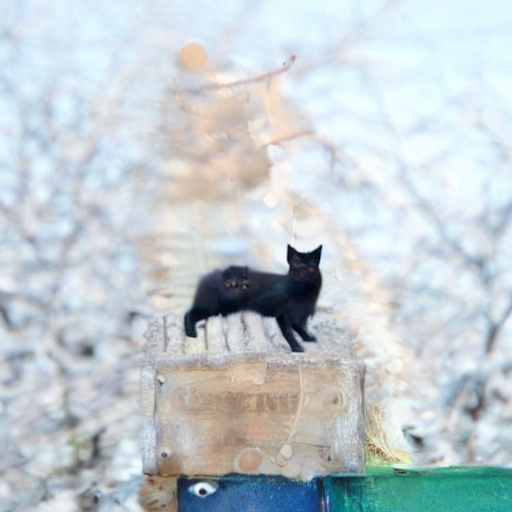}\vspace{3mm}
     \includegraphics[width=\linewidth]{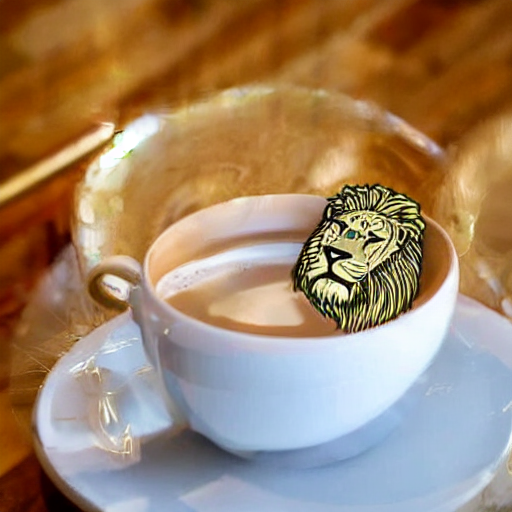}\vspace{3mm}
     \includegraphics[width=\linewidth]{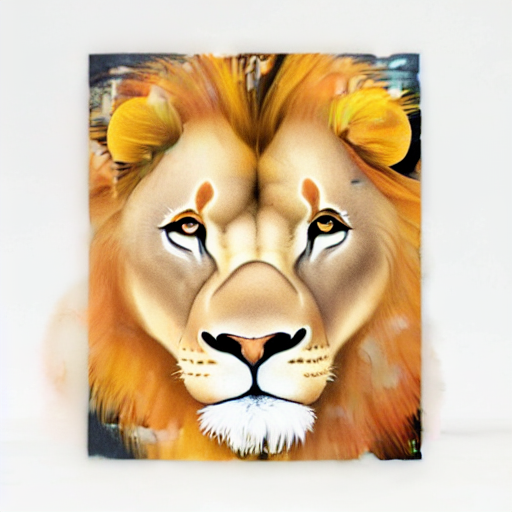}\vspace{3mm}
     \includegraphics[width=\linewidth]{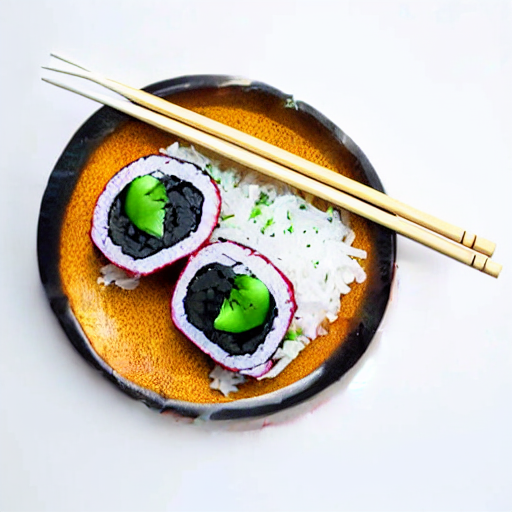}\vspace{3mm}
     \includegraphics[width=\linewidth]{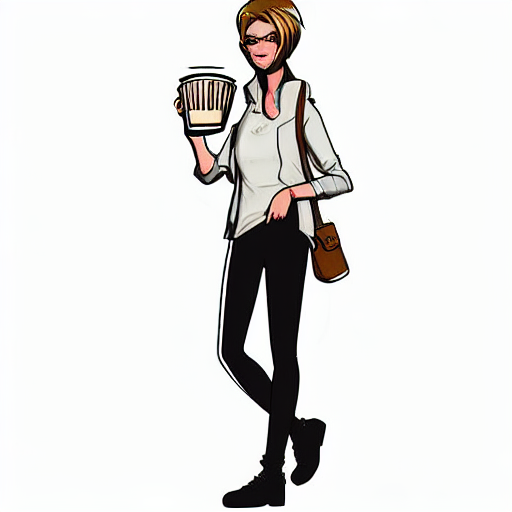}\vspace{3mm}
     \includegraphics[width=\linewidth]{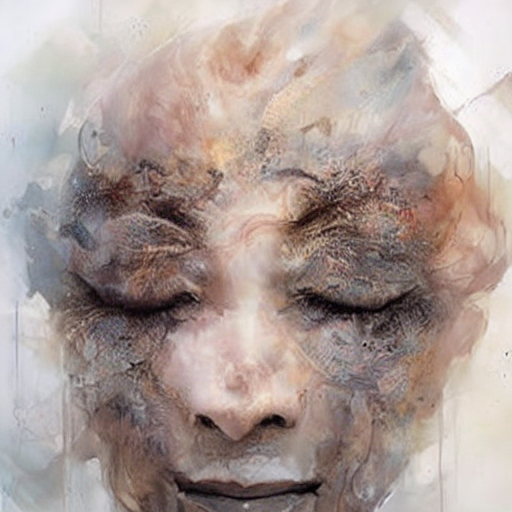}
     \end{minipage}
     }
    \hspace{-2.8mm}
    \subfloat[\emph{Opti.} \\ in $\mathcal{P}$+]{
     \begin{minipage}{0.1\linewidth}
     \includegraphics[width=\linewidth]{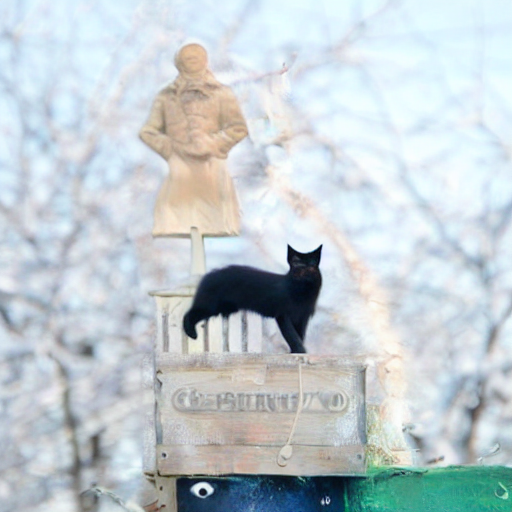}\vspace{3mm}
     \includegraphics[width=\linewidth]{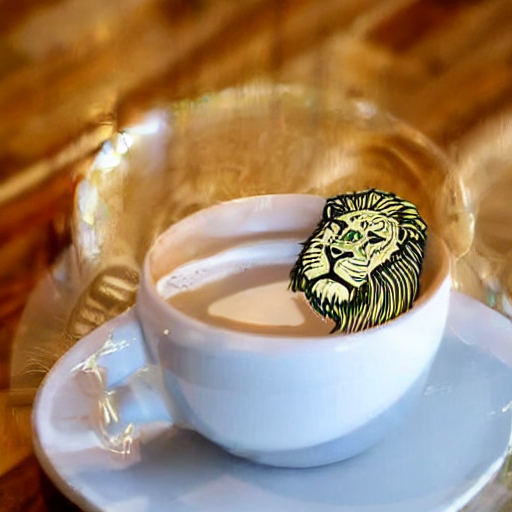}\vspace{3mm}
     \includegraphics[width=\linewidth]{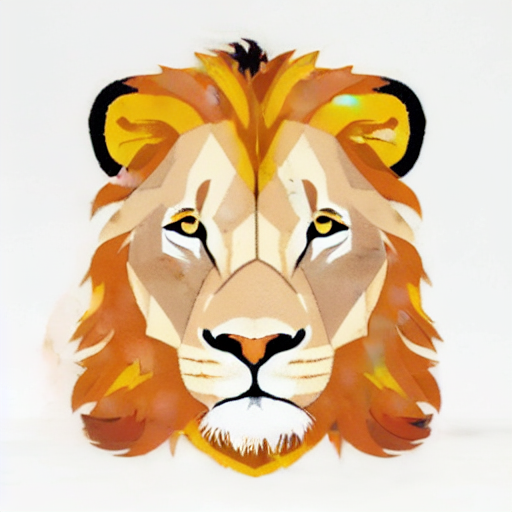}\vspace{3mm}
     \includegraphics[width=\linewidth]{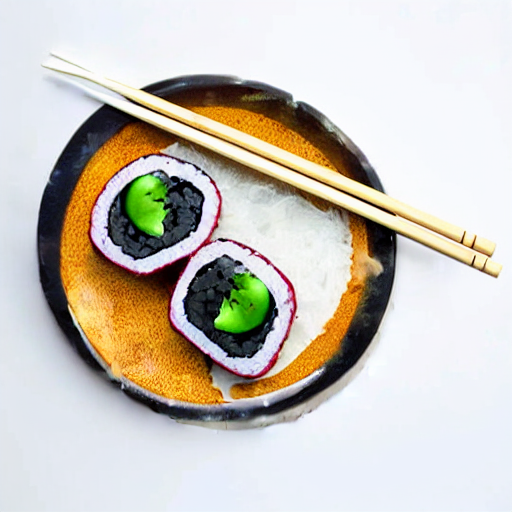}\vspace{3mm}
     \includegraphics[width=\linewidth]{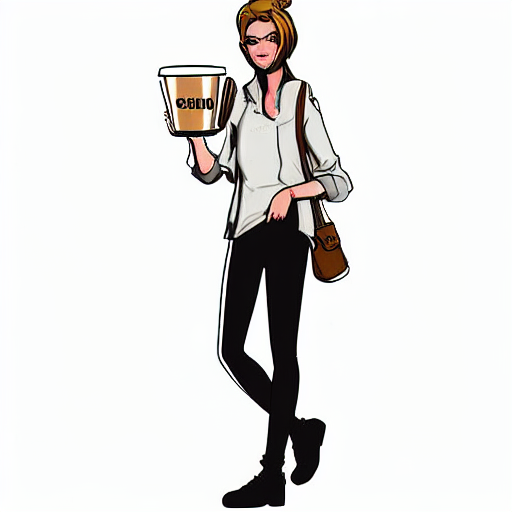}\vspace{3mm}
     \includegraphics[width=\linewidth]{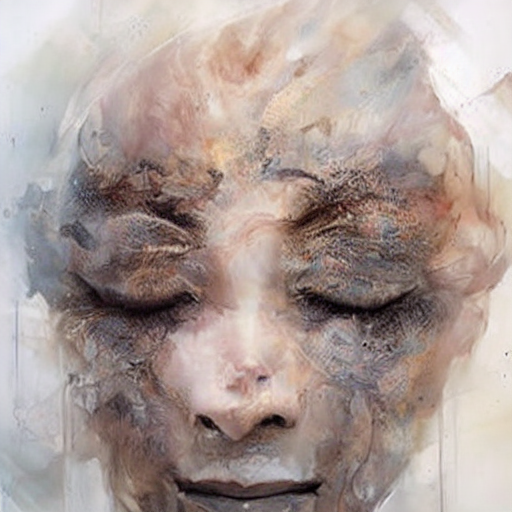}
     \end{minipage}
     }
    \hspace{-2.8mm}
    \subfloat[\emph{T=50, K=25}]{
     \begin{minipage}{0.1\linewidth}
     \includegraphics[width=\linewidth]{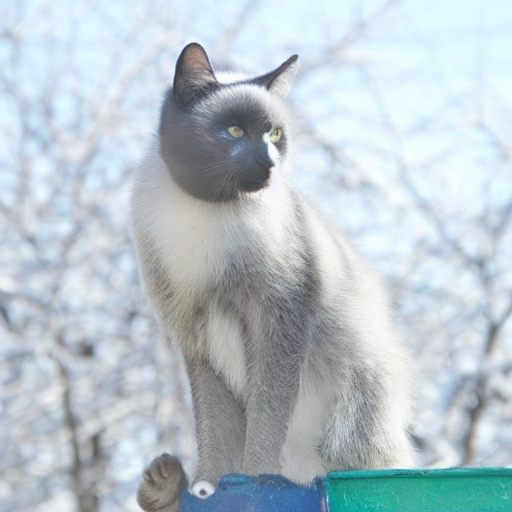}\vspace{3mm}
     \includegraphics[width=\linewidth]{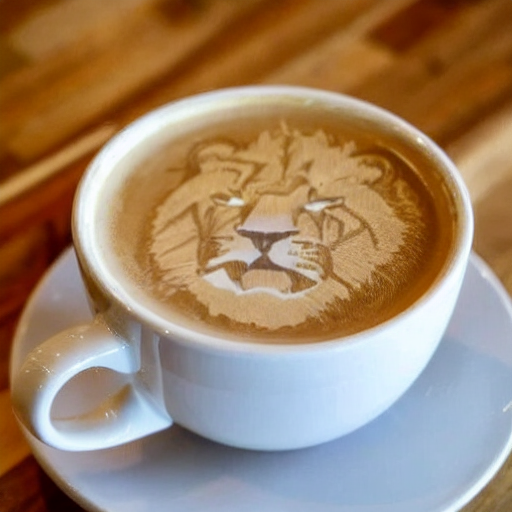}\vspace{3mm}
     \includegraphics[width=\linewidth]{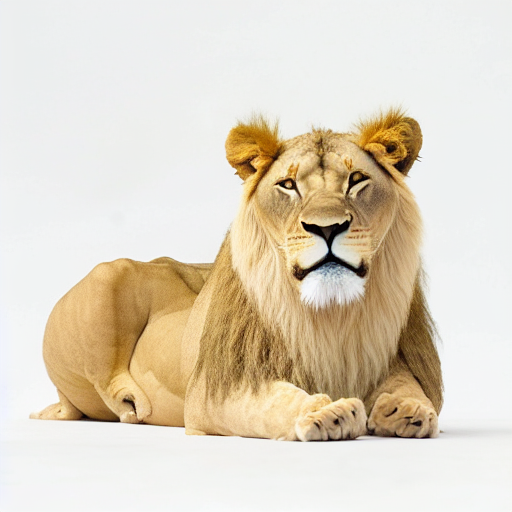}\vspace{3mm}
     \includegraphics[width=\linewidth]{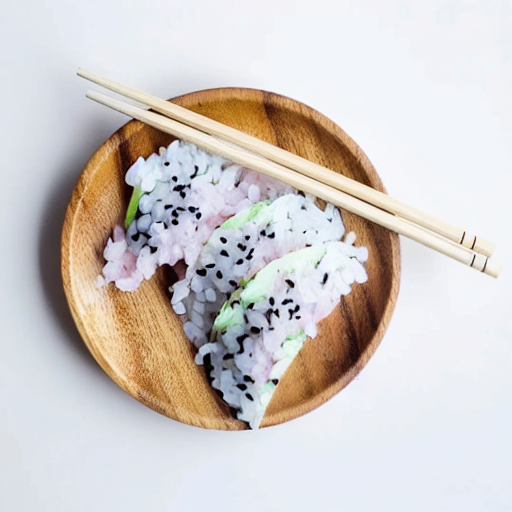}\vspace{3mm}
     \includegraphics[width=\linewidth]{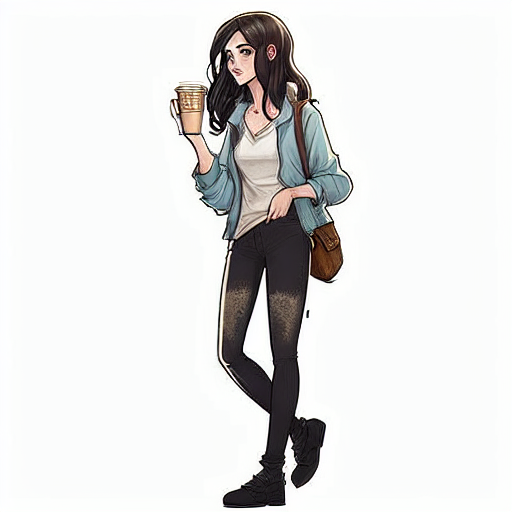}\vspace{3mm}
     \includegraphics[width=\linewidth]{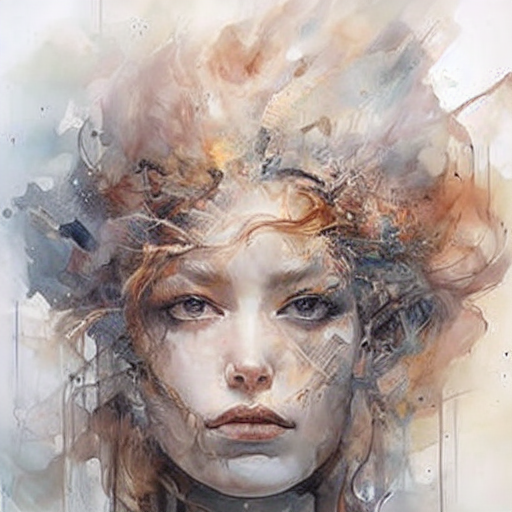}
     \end{minipage}
     }
    \hspace{-2.8mm}
    \subfloat[\emph{T=50, K=50}]{
     \begin{minipage}{0.1\linewidth}
     \includegraphics[width=\linewidth]{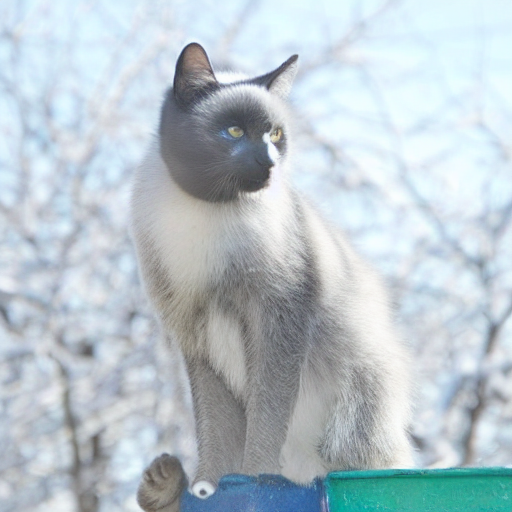}\vspace{3mm}
     \includegraphics[width=\linewidth]{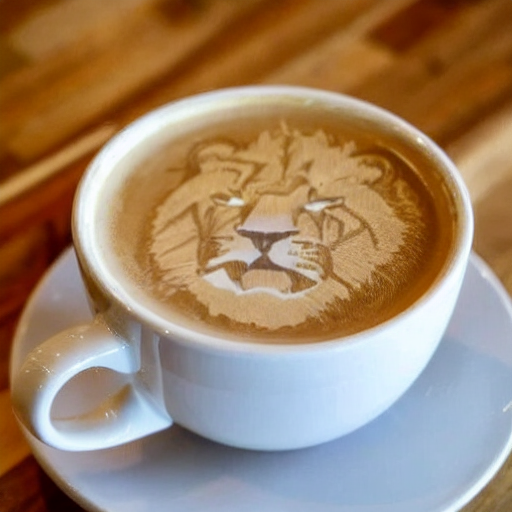}\vspace{3mm}
     \includegraphics[width=\linewidth]{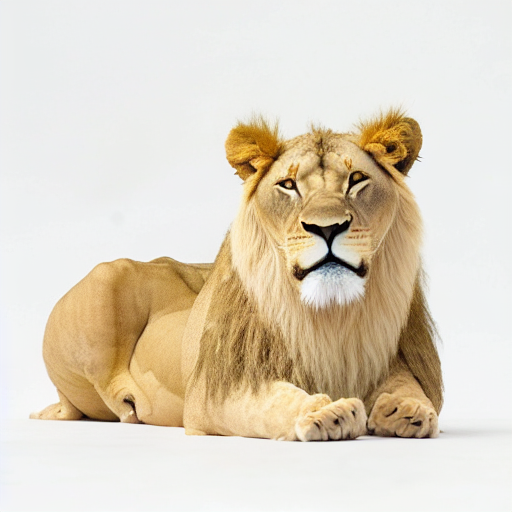}\vspace{3mm}
     \includegraphics[width=\linewidth]{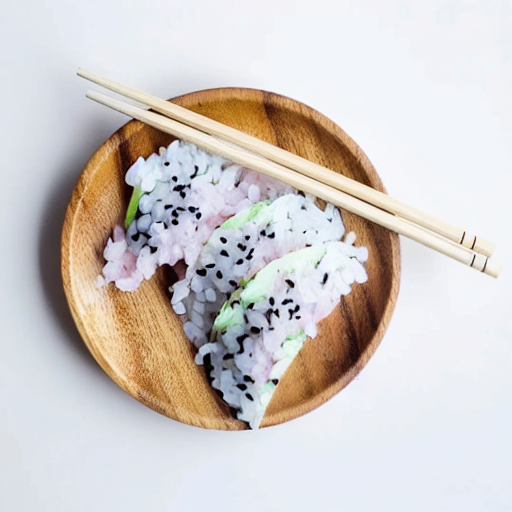}\vspace{3mm}
     \includegraphics[width=\linewidth]{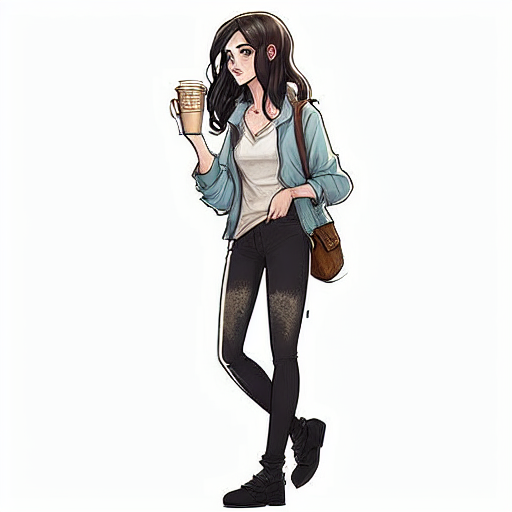}\vspace{3mm}
     \includegraphics[width=\linewidth]{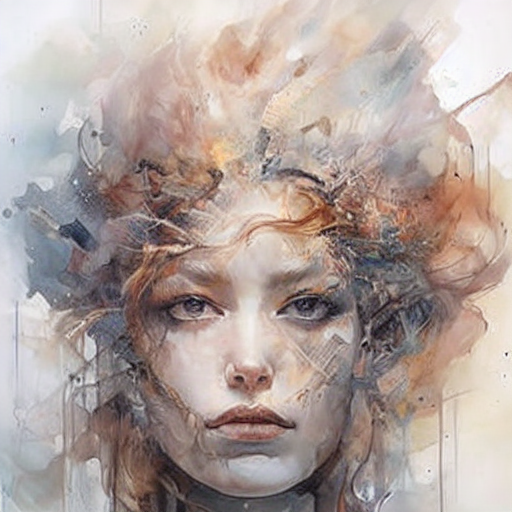}
     \end{minipage}
     }
    \hspace{-2.8mm}
    \subfloat[\emph{T=10, K=10}]{
     \begin{minipage}{0.1\linewidth}
     \includegraphics[width=\linewidth]{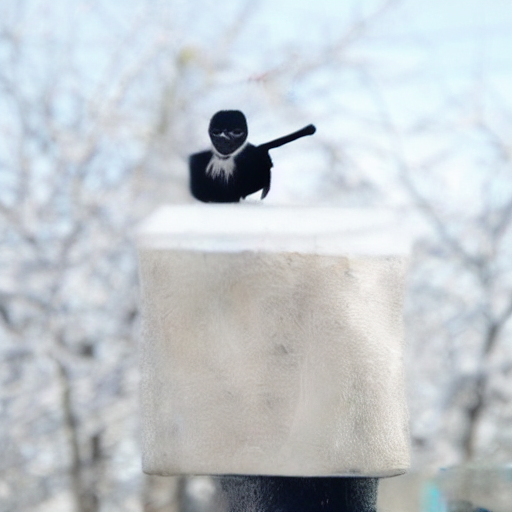}\vspace{3mm}
     \includegraphics[width=\linewidth]{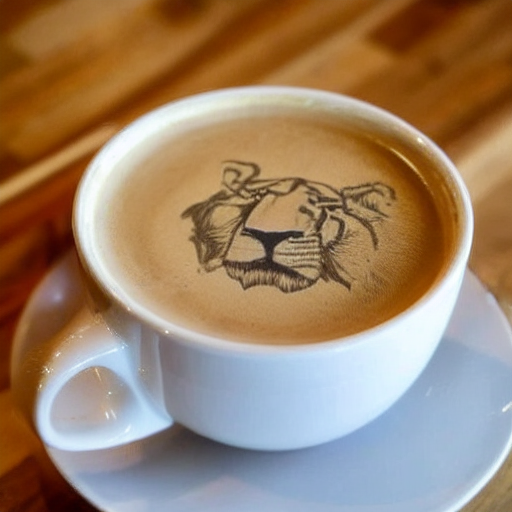}\vspace{3mm}
     \includegraphics[width=\linewidth]{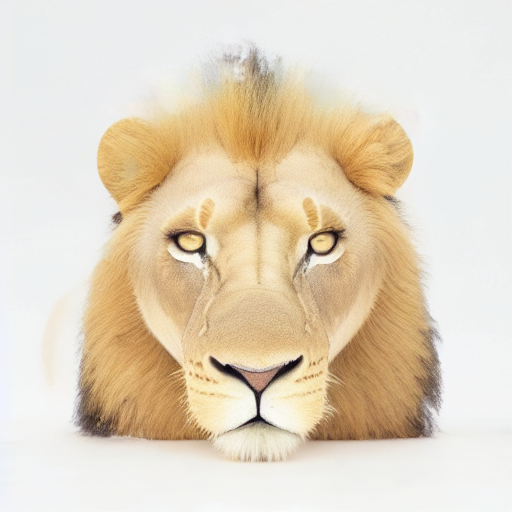}\vspace{3mm}
     \includegraphics[width=\linewidth]{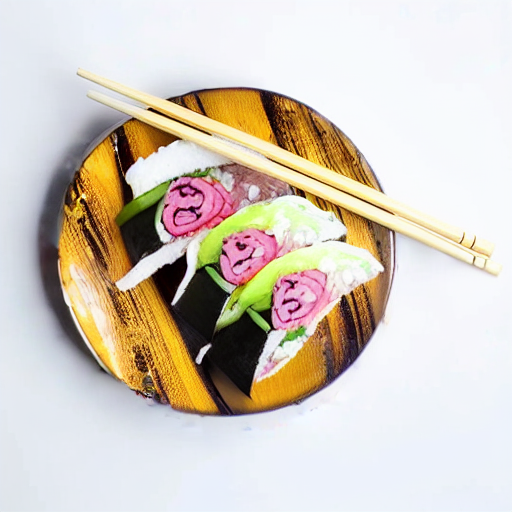}\vspace{3mm}
     \includegraphics[width=\linewidth]{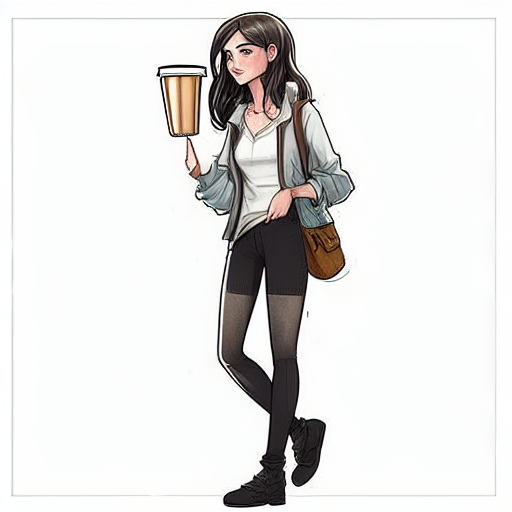}\vspace{3mm}
     \includegraphics[width=\linewidth]{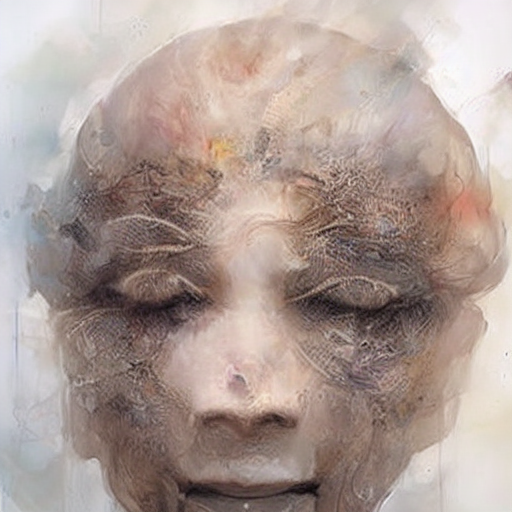}
     \end{minipage}
     }
    \hspace{-2.8mm}
    \subfloat[\emph{T=100, K=10}]{
     \begin{minipage}{0.1\linewidth}
     \includegraphics[width=\linewidth]{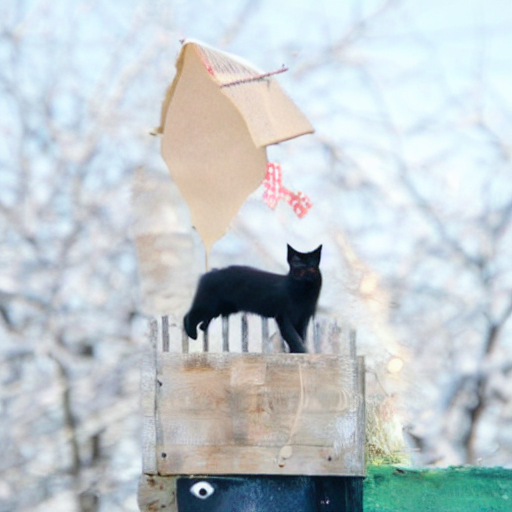}\vspace{3mm}
     \includegraphics[width=\linewidth]{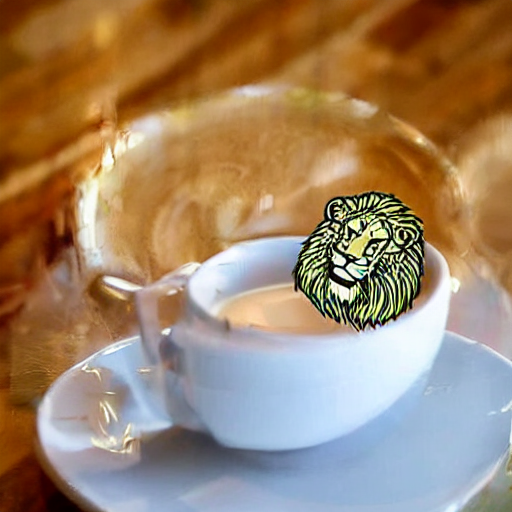}\vspace{3mm}
     \includegraphics[width=\linewidth]{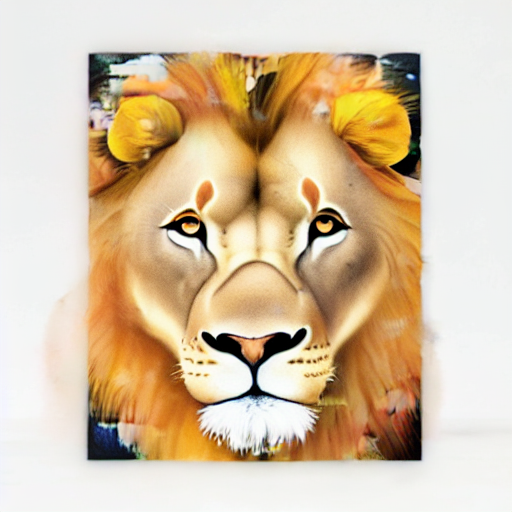}\vspace{3mm}
     \includegraphics[width=\linewidth]{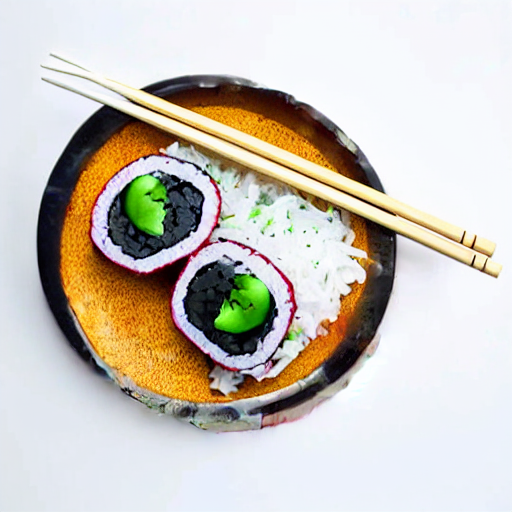}\vspace{3mm}
     \includegraphics[width=\linewidth]{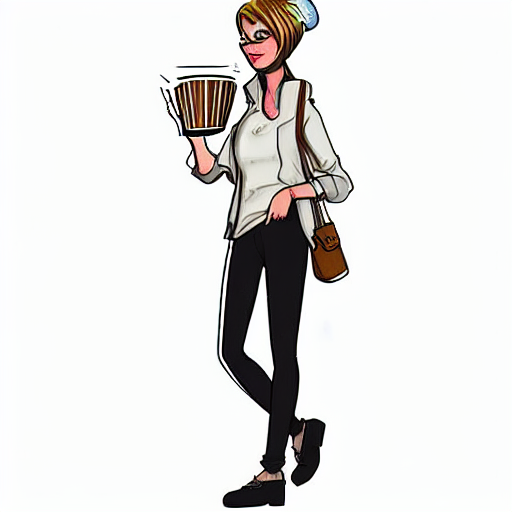}\vspace{3mm}
     \includegraphics[width=\linewidth]{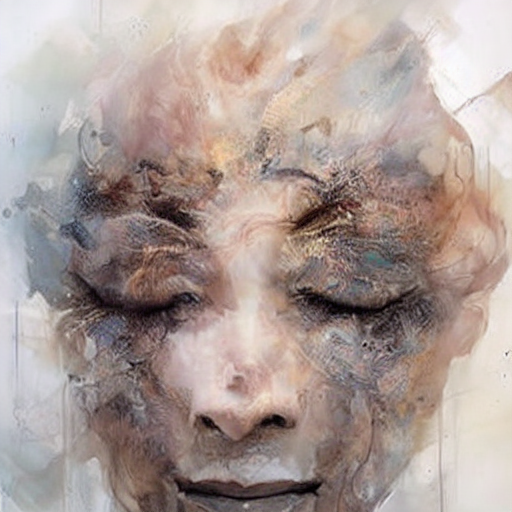}
     \end{minipage}
     }
    \hspace{-2.8mm}
    \subfloat[\emph{w/o} TOPO]{
     \begin{minipage}{0.1\linewidth}
     \includegraphics[width=\linewidth]{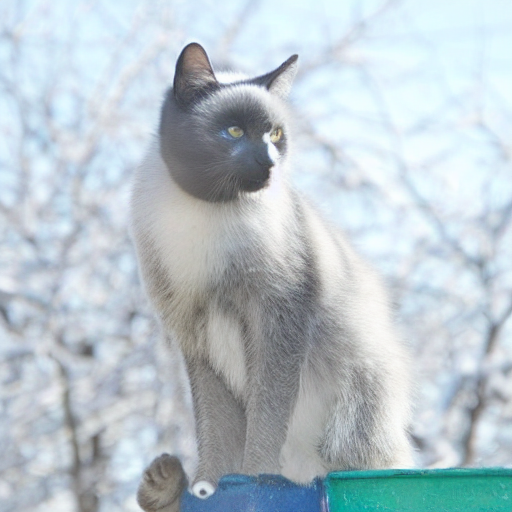}\vspace{3mm}
     \includegraphics[width=\linewidth]{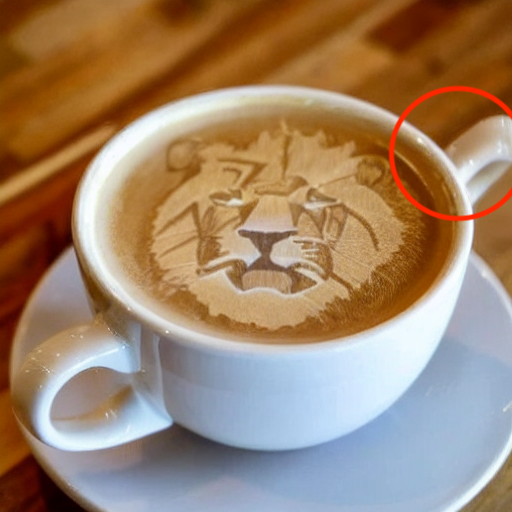}\vspace{3mm}
     \includegraphics[width=\linewidth]{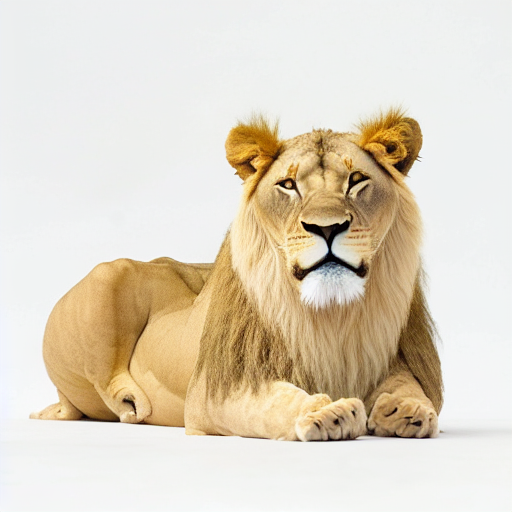}\vspace{3mm}
     \includegraphics[width=\linewidth]{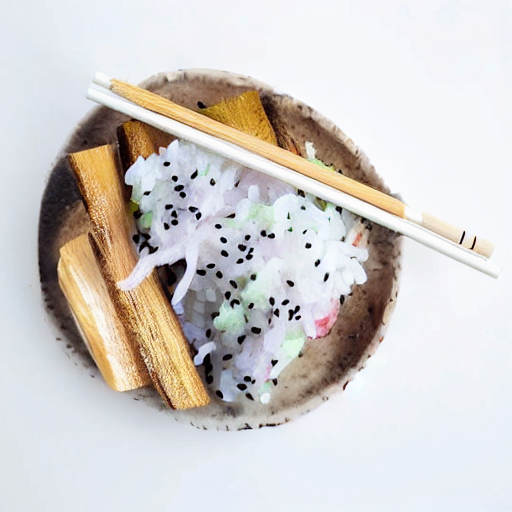}\vspace{3mm}
     \includegraphics[width=\linewidth]{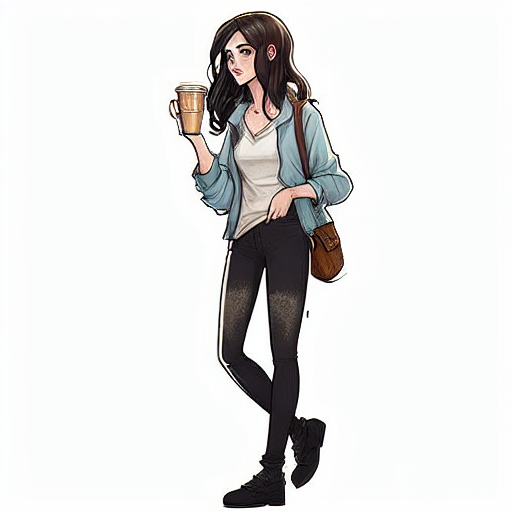}\vspace{3mm}
     \includegraphics[width=\linewidth]{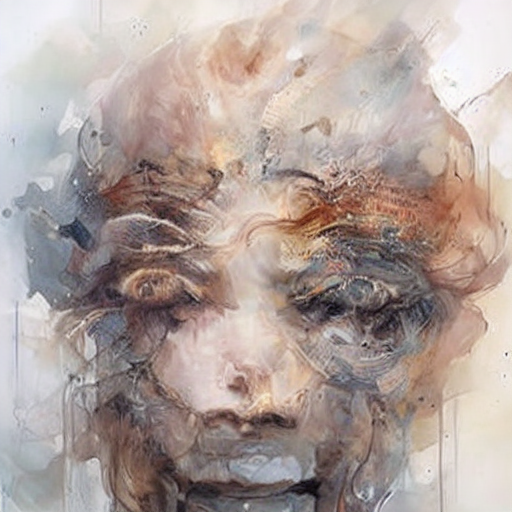}
     \end{minipage}
     }
    \hspace{-2.8mm}
    \subfloat[TODInv]{
     \begin{minipage}{0.1\linewidth}
     \includegraphics[width=\linewidth]{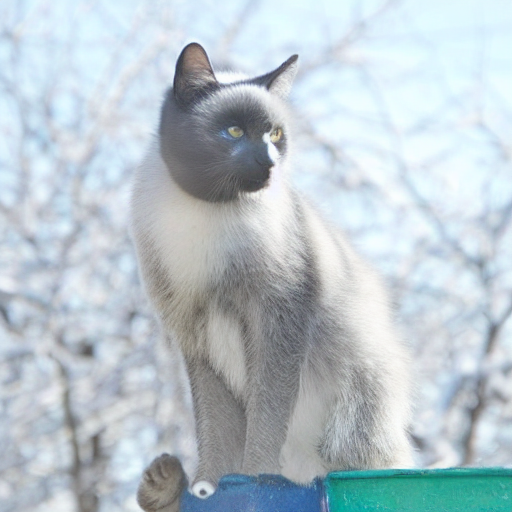}\vspace{3mm}
     \includegraphics[width=\linewidth]{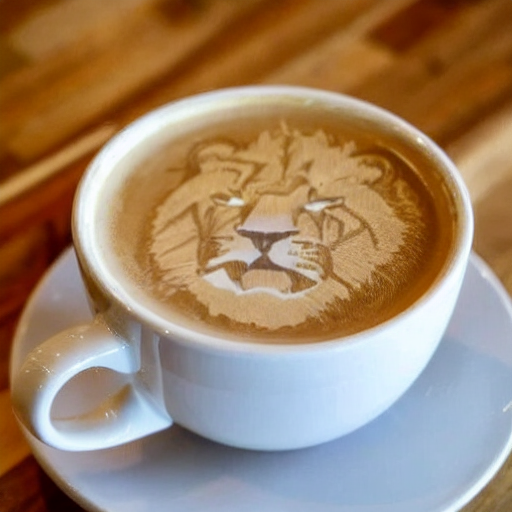}\vspace{3mm}
     \includegraphics[width=\linewidth]{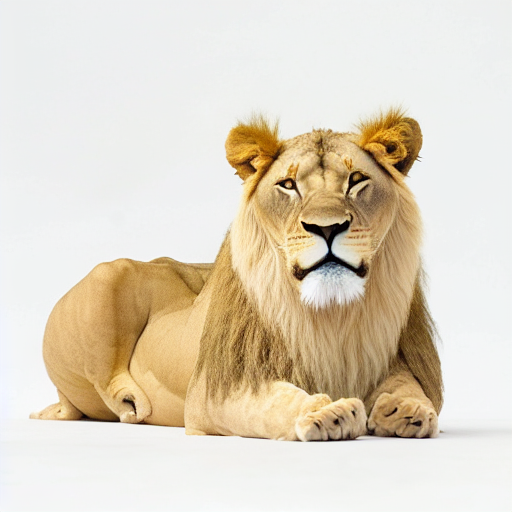}\vspace{3mm}
     \includegraphics[width=\linewidth]{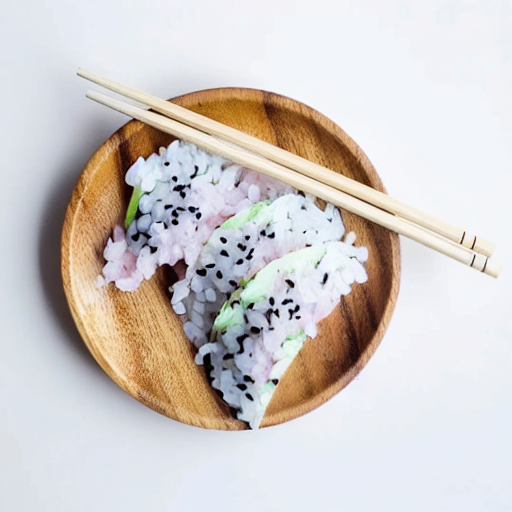}\vspace{3mm}
     \includegraphics[width=\linewidth]{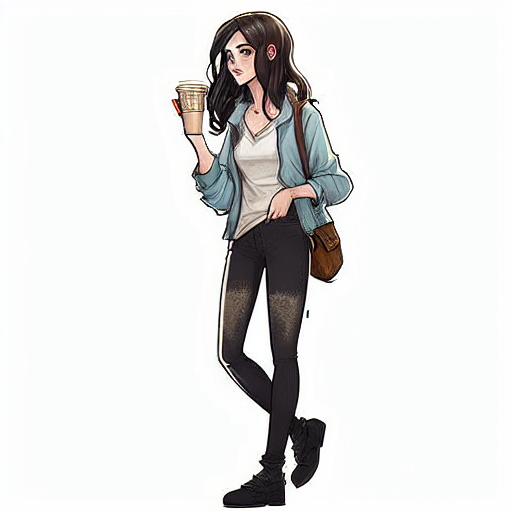}\vspace{3mm}
     \includegraphics[width=\linewidth]{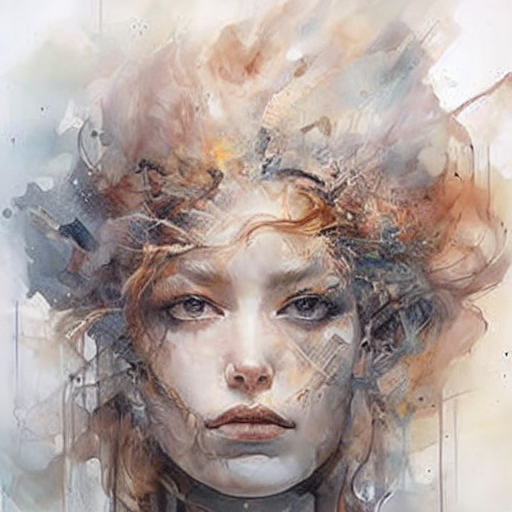}
     \end{minipage}
     }
\vspace{-2mm}
\caption{Qualitative comparison with various variants using P2P editing method.}\vspace{-3mm}
\label{fig:figure_ab}
\end{figure*}

We present a qualitative comparison of different variants in Fig.~\ref{fig:figure_ab}. The images edited by the variants \emph{Opti.} in $\mathcal{P}$, \emph{Opti.} in $\mathcal{P}_t$, and \emph{Opti.} in $\mathcal{P}$+ show inferior results. These variants fail to preserve necessary information from the source images. In contrast, TODInv not only edits the images according to the target prompt but also maintains the unchanged parts of the image. This demonstrates the effectiveness of optimization in the $\mathcal{P}^*$ space, which preserves source information and allows for effective editing. Variants \emph{T=50, K=25} and \emph{T=50, K=50} yield results similar to TODInv, indicating that additional optimization steps are unnecessary for TODInv.

In comparison, the variant \emph{w/o} TOPO shows structural deformation in the last sample of Fig.~\ref{fig:figure_ab} and background perturbation in the $2_{nd}$ and $4_{th}$ samples. With our task-oriented prompt optimization strategy, we only optimize prompt embeddings relevant to the current editing type. This approach not only reconstructs the unedited regions but also preserves editability.

\subsection{Analysis on Task-Oriented Prompt Optimization Strategy}

To demonstrate the effectiveness of our task-oriented prompt optimization strategy, we present a quantitative comparison across different editing types. We evaluate variants \emph{w/o} TOPO for appearance editing and \emph{w/o} TOPO for structure editing. Additionally, we present the results of reversing the editing type (TODInv-Reverse), wherein appearance editing is applied to samples originally intended for structure editing and vice versa. As discussed in Sec.~\ref{expset}, the \textbf{Structure Distance} metric is not suitable for evaluating whether the images are correctly edited; therefore, we exclude this metric from the evaluation of structure editing.

The quantitative comparison is shown in Tab.~\ref{tab:ablation_app_str}. All variants achieve similar performance in background preservation metrics for both appearance and structure editing, as they are all optimized in the expressive $\mathcal{P}^*$ space. Our strategy optimizes prompt embeddings that are independent of the editing type, which enhances editability. Consequently, TODInv-Reverse exhibits poorer performance in CLIP similarity metrics for both appearance and structure editing compared to TODInv, which achieves the best CLIP similarity performance.

\begin{table}[htbp]
\centering%
\caption{{Qualitative comparisons with various variants on different editing types.}}
\vspace{-4mm}
\small
\centering
\renewcommand\arraystretch{0.01}
\setlength{\tabcolsep}{0.05mm}{
\begin{threeparttable}
\begin{tabular}{c|c|c|c|c|c|c|c|c}
\toprule
\toprule

\multirow{2}{*}{{\textbf{Variant}}}   &  \multirow{2}{*}{{\textbf{Editing Type}}}      & \textbf{Structure}          & \multicolumn{4}{c|}{\textbf{Background Preservation}} & \multicolumn{2}{c}{\textbf{CLIP Similarity}} \\
\cmidrule{3-9} &   &\textbf{Distance}$_{^{\times 10^3}}$ $\downarrow$ & \textbf{PSNR} $\uparrow$     & \textbf{LPIPS}$_{^{\times 10^3}}$ $\downarrow$  & \textbf{MSE}$_{^{\times 10^4}}$ $\downarrow$     & \textbf{SSIM}$_{^{\times 10^2}}$ $\uparrow$    & \textbf{Whole}  $\uparrow$          & \textbf{Edited}  $\uparrow$       \\

\midrule

{\emph{w/o} TOPO}     &\textbf{Appearance}   &\textbf{8.87}&\textbf{29.16}  &\textbf{38.71}           &\textbf{26.75}          &86.85    &{25.44}          &{23.41}           \\

{{TODInv-Reverse}}  &\textbf{Appearance}     &9.18         &{29.00}         &38.95          &27.79          &{86.97}       &{25.29}      &{23.09}      \\

{\textbf{TODInv}}    &\textbf{Appearance}    &{9.17}       &{29.07}         &{38.83}      &{27.65}         &\textbf{86.94} &\textbf{26.23}      & \textbf{24.04}        \\

\midrule
{\emph{w/o} TOPO}          &\textbf{Structure}          & -   &{28.31}  &{44.22}   &25.84             & 84.70           &{24.23}      &{19.62}     \\

{{TODInv-Reverse}}  &\textbf{Structure}    &-                &\textbf{27.66}         &{44.85}          &{25.62}         &{84.74}        &{24.18}      & {19.53}      \\

{\textbf{TODInv}}          &\textbf{Structure}          &-       &{28.01}     &\textbf{42.49}          &\textbf{24.39}         &\textbf{85.07} &\textbf{25.24}      &\textbf{20.63}    \\

\bottomrule
\bottomrule
\end{tabular}
\end{threeparttable}}
\label{tab:ablation_app_str}
\vspace{-3mm}
\end{table}

\begin{figure*}[h]
    \centering
    \captionsetup[subfloat]{labelformat=empty,justification=centering}
    \subfloat[Source]{
     \begin{minipage}{0.14\linewidth}
     \includegraphics[width=\linewidth]{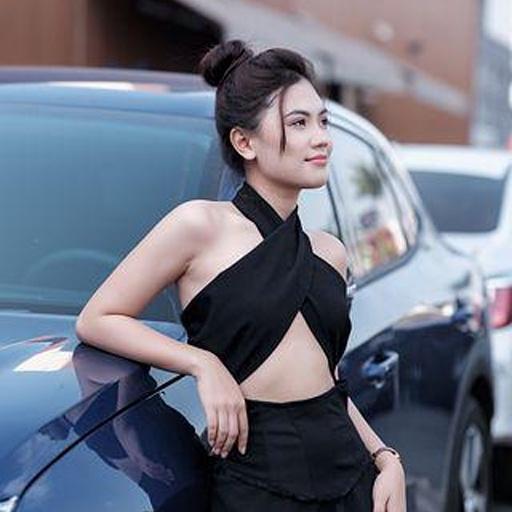}\vspace{3mm}
     \includegraphics[width=\linewidth]{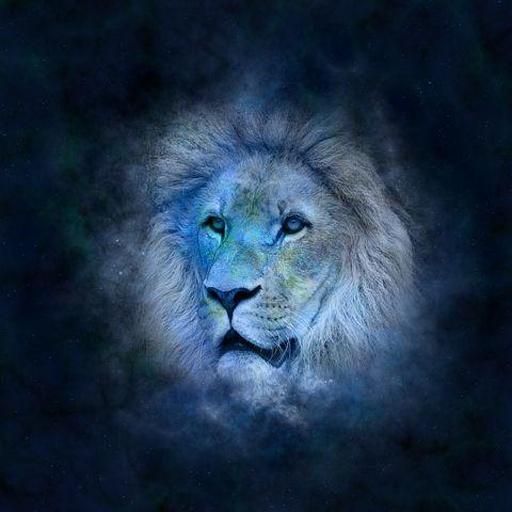}\vspace{3mm}
     \includegraphics[width=\linewidth]{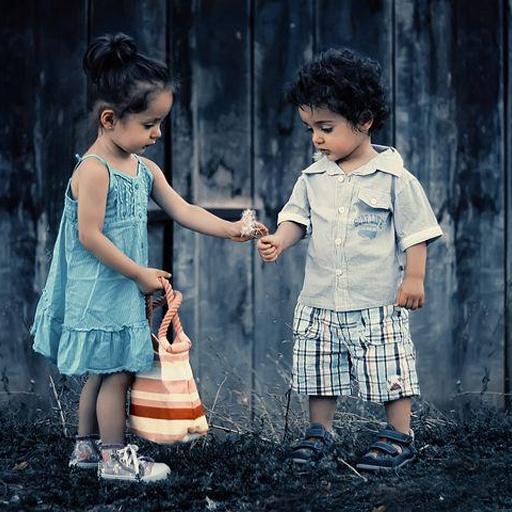}\vspace{3mm}
     \includegraphics[width=\linewidth]{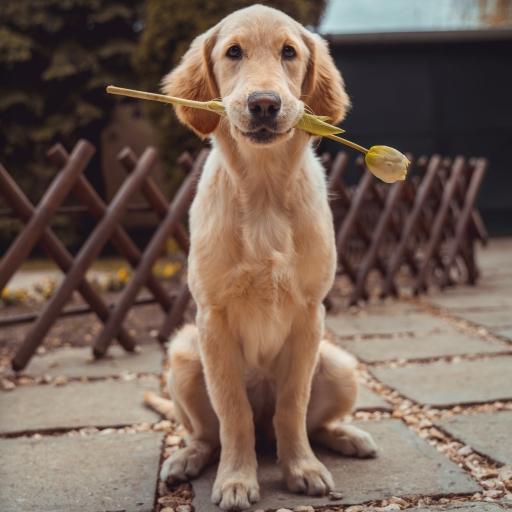}
     \end{minipage}
     }
     \put(-112.5,130){{\scriptsize{\textsc{a woman in black dress leaning against a \rev{car} $\rightarrow$ a woman in black dress leaning against a \rev{wall}}}}}
     \put(-95,64){{\scriptsize{\textsc{a lion in the sky with a \rev{blue background} $\rightarrow$ a lion in the sky with a \rev{red background}}}}}
     \put(-75,-1){{\scriptsize{\textsc{two young children ... $\rightarrow$ \rev{stained glass window of} two young children ...}}}}
     \put(-115,-66){{\scriptsize{\textsc{a golden retriever \rev{holding a flower} sitting ... $\rightarrow$ a golden retriever sitting ...}}}}
    \hspace{-2.8mm}
    \subfloat[\emph{w/o} TOPO]{
     \begin{minipage}{0.14\linewidth}
     \includegraphics[width=\linewidth]{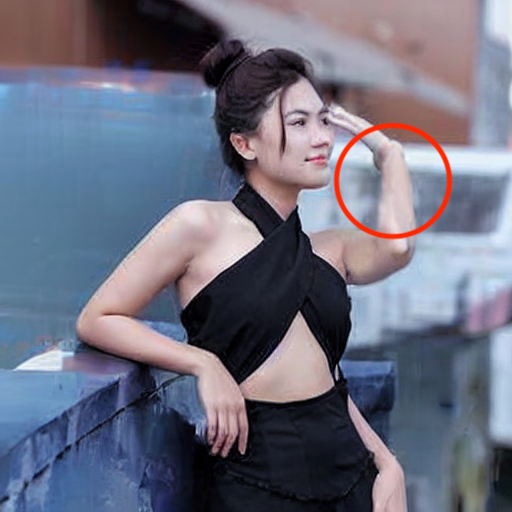}\vspace{3mm}
     \includegraphics[width=\linewidth]{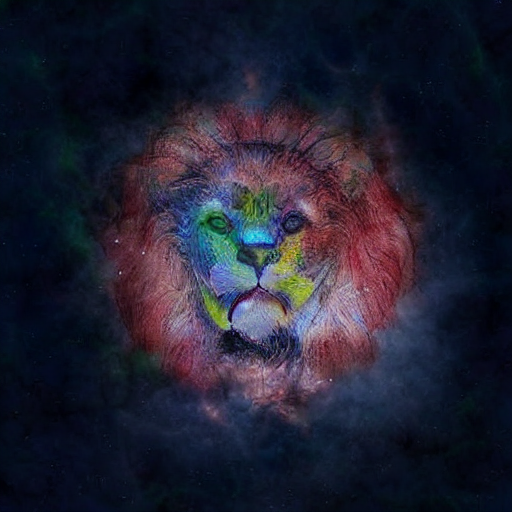}\vspace{3mm}
     \includegraphics[width=\linewidth]{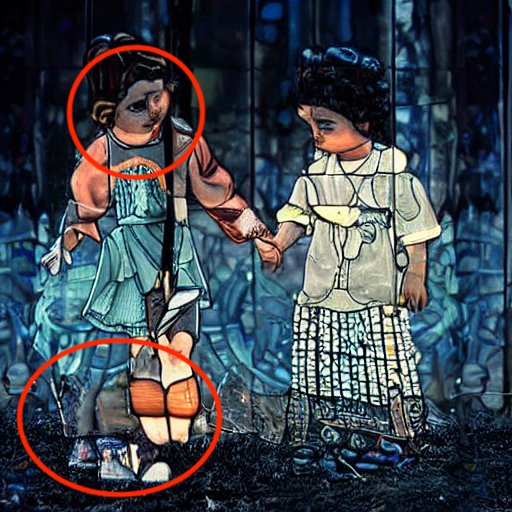}\vspace{3mm}
     \includegraphics[width=\linewidth]{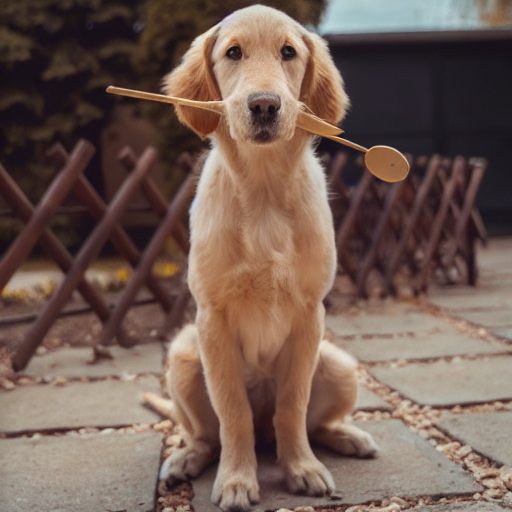}
     \end{minipage}
     }
    \hspace{-2.8mm}
    \subfloat[TODInv-Reverse]{
     \begin{minipage}{0.14\linewidth}
     \includegraphics[width=\linewidth]{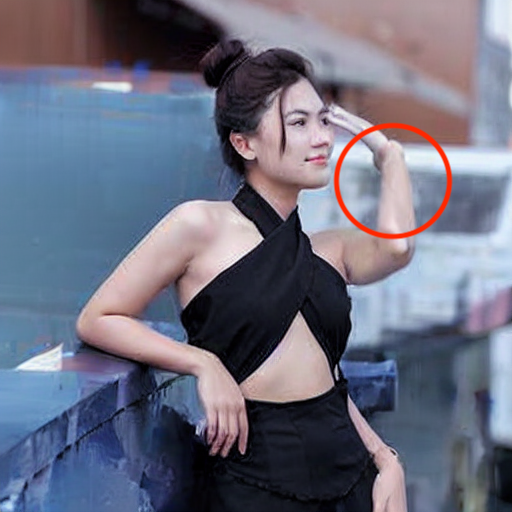}\vspace{3mm}
     \includegraphics[width=\linewidth]{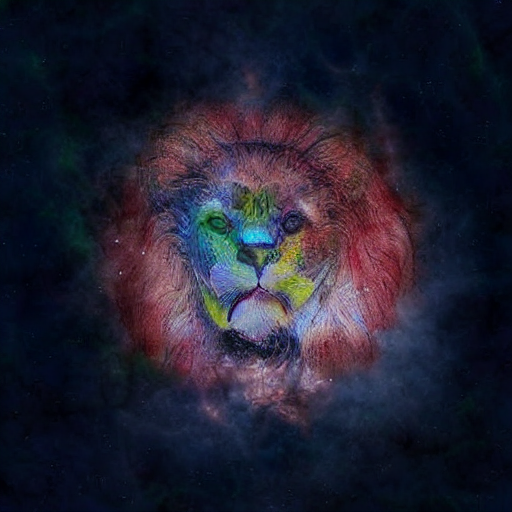}\vspace{3mm}
     \includegraphics[width=\linewidth]{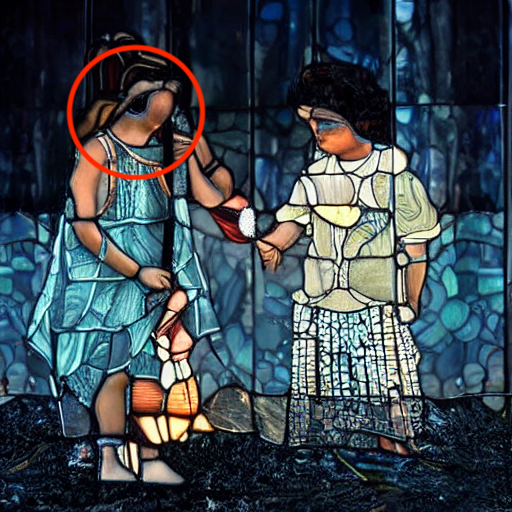}\vspace{3mm}
     \includegraphics[width=\linewidth]{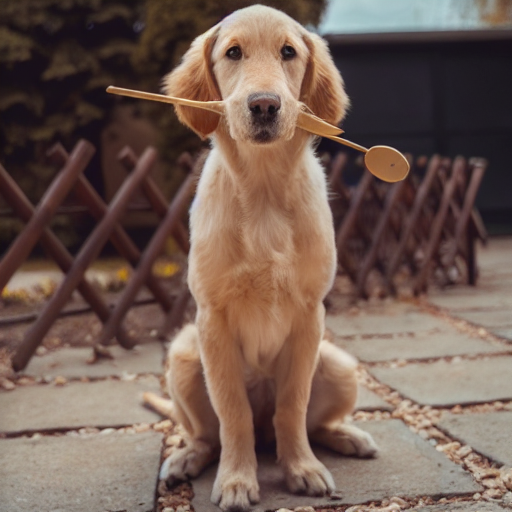}
     \end{minipage}
     }
    \hspace{-2.8mm}
    \subfloat[TODInv]{
     \begin{minipage}{0.14\linewidth}
     \includegraphics[width=\linewidth]{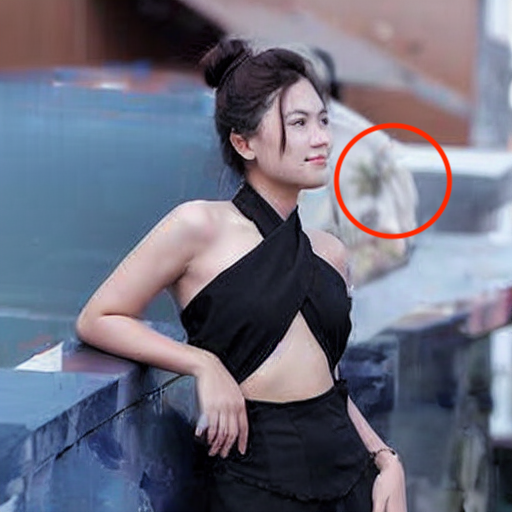}\vspace{3mm}
     \includegraphics[width=\linewidth]{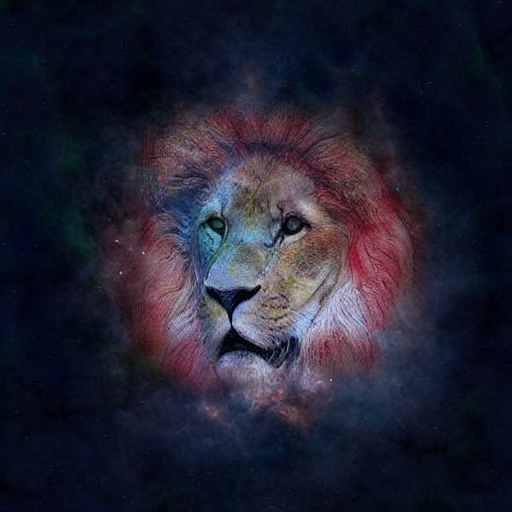}\vspace{3mm}
     \includegraphics[width=\linewidth]{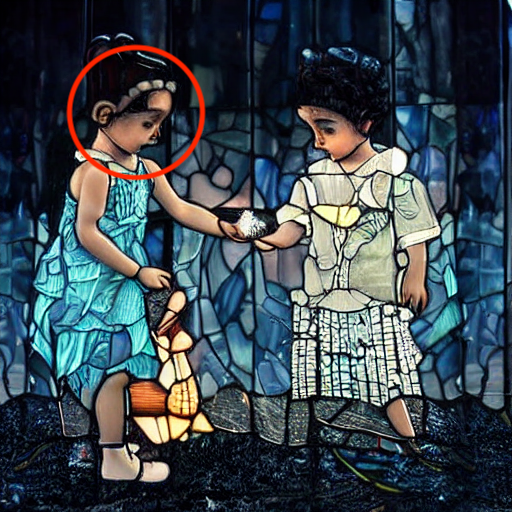}\vspace{3mm}
     \includegraphics[width=\linewidth]{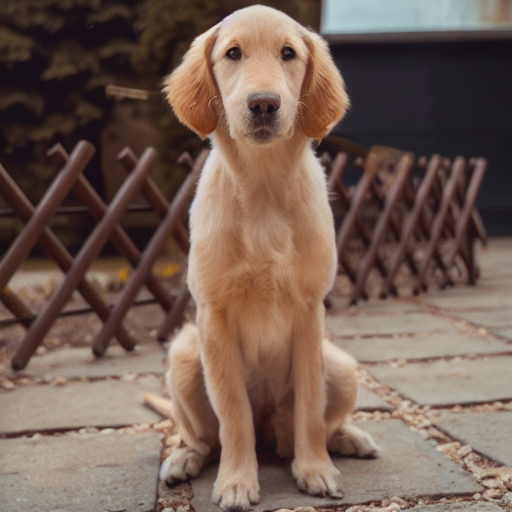}
     \end{minipage}
     }
\vspace{-2mm}
\caption{Qualitative comparison with \emph{w/o} TOPO and TODInv-Reverse variants using P2P editing method.}\vspace{-3mm}
\label{fig:figure_ab_str}
\end{figure*}

We also present the qualitative comparison in Fig.~\ref{fig:figure_ab_str}, showing that variant \emph{w/o} TOPO and TODInv-Reverse easily present the structure deformation. As shown in the red circle in $1_{st}$ sample of Fig.~\ref{fig:figure_ab_str}, variant \emph{w/o} TOPO and TODInv-Reverse present the undesired arms in the edited images and modify the view of lions in $2_{nd}$ sample. In $3_{rd}$ sample, both variant \emph{w/o} TOPO and TODInv-Reverse fail to preserve the facial features of source faces, and variant TODInv-Reverse also modifies the ``\textsc{legs}'' of children. In $4_{th}$ sample, neither variant \emph{w/o} TOPO and TODInv-Reverse failed to remove the ``\textsc{flower}'', which further demonstrates the effectiveness of our task-oriented prompt optimization strategy.

\subsection{Quantitative Comparison on different editing categories}

We present the quantitative comparison on different editing categories date in Tab.~\ref{tab:cate_app}, Tab.~\ref{tab:cate_str}, and Tab.~\ref{tab:cate_global}. Here we use the edited results of other methods provided by PNP's re-implementation~\cite{PnPInversion}. From Tab.~\ref{tab:cate_app} we can see that our TODInv outperforms other methods with P2P, MasaCtrl, and PNP editing methods on appearance editing categories on all metrics, especially on the structure preservation, our method outperforms other methods with a large step, that demonstrates the effectiveness of our TOPO strategy, by only optimizing the irrelevant layers with appearance editing, our TODInv preserves the structures information of original images effectively.

The quantitative comparison of the images with structure editing category can be seen in Tab.~\ref{tab:cate_str}. Our TODInv outperforms other methods on all metrics with most editing methods, except with the P2P-Zero editing on background preservation, that is because P2P-Zero is proposed for image translation but not prompt-driven image editing. Compared with P2P, PNP, and MasaCtrl, DDIM and PNPInv inversion methods also receive worse performance on background preservation.

At last, the quantitative comparison of the images with global editing category can be seen in Tab.~\ref{tab:cate_global}. Our TODInv also goes beyond other methods on most metrics.

\begin{table*}[htbp]
\caption{{Qualitative comparisons on \textbf{appearance editing category} with related works using various text-guided editing methods.}}
\vspace{-0.3cm}
\small
\centering
  \renewcommand\arraystretch{0.01}
\setlength{\tabcolsep}{0.05mm}{
\begin{threeparttable}
\begin{tabular}{c|c|c|c|cccc|cc}
\toprule
\toprule
\multicolumn{3}{c|}{\textbf{Method}}           & \textbf{\textbf{Structure}}          & \multicolumn{4}{c|}{\textbf{Background Preservation}} & \multicolumn{2}{c}{\textbf{CLIP Similarity}} \\
\cmidrule{1-10}
\textbf{Inverse}          & \textbf{Editing}   &\textbf{Editing Type}           & \textbf{Distance}$_{^{\times 10^3}}$ $\downarrow$ & \textbf{PSNR} $\uparrow$     & \textbf{LPIPS}$_{^{\times 10^3}}$ $\downarrow$  & \textbf{MSE}$_{^{\times 10^4}}$ $\downarrow$     & \textbf{SSIM}$_{^{\times 10^2}}$ $\uparrow$    & \textbf{Whole}  $\uparrow$          & \textbf{Edited}  $\uparrow$       \\

\midrule
\textbf{DDIM} & \textbf{P2P}     &\textbf{Appearance}       &67.93           &17.97      &203.70   &214.33   &72.71     &{25.21}      &{23.75}    \\

\textbf{NTI}  & \textbf{P2P}     &\textbf{Appearance}           &14.45           &{28.10}    &{55.73}  &{32.46}  &{85.64}   &25.77        &24.06      \\

\textbf{NPI}  & \textbf{P2P}     &\textbf{Appearance}            &18.63           &26.78      &66.08    &39.24    &84.70     &25.43        &23.80    \\

\textbf{StyleD}  & \textbf{P2P}  &\textbf{Appearance}            &{12.11}         &26.76      &63.86    &36.88    &84.81     & 25.27        &23.40     \\

\textbf{PNPInv} & \textbf{P2P}   &\textbf{Appearance}             &{12.39}         &{28.53}    &{48.22}  &\textbf{27.65}  &{86.39}   &{25.69}      &{23.93}     \\

\midrule
\textbf{TODInv}             & \textbf{P2P}   &\textbf{Appearance}   &\textbf{9.17}       &\textbf{29.07}  &\textbf{38.83}  &\textbf{27.65}         &\textbf{86.94} &\textbf{26.23}      & \textbf{24.04}       \\

\midrule
\midrule

\textbf{DDIM} & \textbf{MasaCtrl}       &\textbf{Appearance}         & 29.09       &22.38  &101.20  &84.88  &81.03 &24.00         & 22.20        \\
\textbf{PNPInv}  & \textbf{MasaCtrl}    &\textbf{Appearance}             & {24.49}           & {22.95}  & {84.23}  & {79.83}  & {82.50} & {24.37}        & \textbf{22.55}      \\

\midrule
{\textbf{TODInv}}   & \textbf{MasaCtrl}    & \textbf{Appearance}           &\textbf{18.66}       &\textbf{24.66}  &\textbf{66.94}  &\textbf{60.81}  &\textbf{84.30} &\textbf{24.66}         &\textbf{22.55}       \\
\midrule
\midrule

\textbf{DDIM}   &\textbf{PNP}  &\textbf{Appearance}    &30.91 &22.61 &110.11 & 76.64 & 80.18 &  26.2 &24.49   \\
\textbf{PNPInv} & \textbf{PNP} &\textbf{Appearance}    &{26.40} & 22.89 & 104.77 &73.82 & 81.02 &26.21 &24.62    \\
\midrule
{\textbf{TODInv}}       & \textbf{PNP}   &\textbf{Appearance}   &\textbf{24.22}    &\textbf{25.31}         &\textbf{77.87}  &\textbf{54.32}  &\textbf{84.13}        &\textbf{27.50}            &\textbf{25.43}           \\
\midrule
\midrule

\textbf{DDIM} & \textbf{P2P-Zero}   &\textbf{Appearance}   &74.20  &20.21  & 169.57 & 147.82 & 76.12 & 22.95 & 21.76 \\
\textbf{PNPInv} & \textbf{P2P-Zero} &\textbf{Appearance}   & {65.51} & \textbf{21.30} & \textbf{137.77} & \textbf{134.84} & {78.45} & {23.53} & {22.18} \\
\midrule
\textbf{TODInv}             & \textbf{P2P-Zero}    & \textbf{Appearance}    &\textbf{62.70}       & {21.05}  & {138.70}  & {137.10}  & \textbf{78.60} & \textbf{24.39} & \textbf{22.62}   \\

\bottomrule
\bottomrule
\end{tabular}

\end{threeparttable}}

\label{tab:cate_app}
\end{table*}

\begin{table*}[htbp]
\caption{{Qualitative comparisons on \textbf{structure editing category} with related works using various text-guided editing methods.}}
\vspace{-0.3cm}
\small
\centering
  \renewcommand\arraystretch{0.01}
\setlength{\tabcolsep}{0.1mm}{
\begin{threeparttable}
\begin{tabular}{c|c|c|cccc|cc}
\toprule
\toprule
\multicolumn{3}{c|}{\textbf{Method}}                   & \multicolumn{4}{c|}{\textbf{Background Preservation}} & \multicolumn{2}{c}{\textbf{CLIP Similarity}}  \\
\cmidrule{1-9}
\textbf{Inverse}          & \textbf{Editing}   &\textbf{Editing Type}           & \textbf{PSNR} $\uparrow$     & \textbf{LPIPS}$_{^{\times 10^3}}$ $\downarrow$  & \textbf{MSE}$_{^{\times 10^4}}$ $\downarrow$     & \textbf{SSIM}$_{^{\times 10^2}}$ $\uparrow$    & \textbf{Whole}  $\uparrow$          & \textbf{Edited}  $\uparrow$       \\

\midrule
\textbf{DDIM} & \textbf{P2P}     &\textbf{Structure}              &17.27      &230.19   &237.38   &68.45     &{24.98}      &\textbf{21.33}     \\
\textbf{NTI}  & \textbf{P2P}     &\textbf{Structure}                   &{26.30}    &{69.64}  &{38.70}  &{82.43}   &24.23        &20.44      \\
\textbf{NPI}  & \textbf{P2P}     &\textbf{Structure}             &25.66      &76.98    &41.20    &81.82     &24.15        &20.54    \\
\textbf{StyleD}  & \textbf{P2P}  &\textbf{Structure}                &25.53      &72.62    &39.44    &81.99     & 24.57        &20.65     \\
\textbf{PNPInv} & \textbf{P2P}   &\textbf{Structure}                  &{26.41}    &{61.44}  &{35.57}  &{83.24}   &{24.72}      &{20.94}   \\

\midrule
\textbf{TODInv}             & \textbf{P2P}   &\textbf{Structure}         &\textbf{28.01}  &\textbf{42.49}  &\textbf{24.39}         &\textbf{85.07} &\textbf{25.24}      & {20.63}      \\
\midrule
\midrule

\textbf{DDIM} & \textbf{MasaCtrl}       &\textbf{Structure}              &21.51  &118.38  &95.02  &77.73 &24.29       & 20.49    \\
\textbf{PNPInv}  & \textbf{MasaCtrl}    &\textbf{Structure}                     & {21.99}  & {97.51}  & {88.16}  & {79.62} & {24.76}        & \textbf{20.65}       \\
{\textbf{TODInv}}   & \textbf{MasaCtrl}    & \textbf{Structure}               & \textbf{23.82}  &\textbf{77.82}  & {66.50}  &\textbf{81.36} &\textbf{25.15}         &{22.49}        \\

\midrule
\midrule

\textbf{DDIM}   &\textbf{PNP}  &\textbf{Structure}   & 21.73 & 125.06  &90.09 & 77.12 & 25.12 &21.25\\
\textbf{PNPInv} & \textbf{PNP} &\textbf{Structure}   & 21.86 & 116.15 &86.30    & 77.83 &25.15 &21.35  \\
\midrule
{\textbf{TODInv}}       & \textbf{PNP}   &\textbf{Structure}    &\textbf{25.04}         &\textbf{82.28}  &\textbf{47.75}  &\textbf{81.55}        &\textbf{25.48}            &20.76            \\

\midrule
\midrule

\textbf{DDIM} & \textbf{P2P-Zero}   &\textbf{Structure}   &19.88  & 193.89 & 156.54 & 71.94 & 22.54 & 19.49 \\
\textbf{PNPInv} & \textbf{P2P-Zero} &\textbf{Structure}   & \textbf{21.00} & \textbf{156.94} & \textbf{136.81} & \textbf{74.53} & {22.95} & {19.98}  \\

\midrule
\textbf{TODInv}             & \textbf{P2P-Zero}    &\textbf{Structure}         &{20.80}  & {158.12}  &{150.68}  &{74.27} & \textbf{23.90}  & \textbf{20.03}   \\

\bottomrule
\bottomrule
\end{tabular}

\end{threeparttable}}


\label{tab:cate_str}

\end{table*}

\begin{table*}[htbp]
\caption{{Qualitative comparisons on \textbf{global editing category} with related works using various text-guided editing methods.}}
\vspace{-0.3cm}
\small
\centering
  \renewcommand\arraystretch{0.01}
\setlength{\tabcolsep}{0.05mm}{
\begin{threeparttable}
\begin{tabular}{c|c|c|c|cccc|cc}
\toprule
\toprule
\multicolumn{3}{c|}{\textbf{Method}}           & \textbf{\textbf{Structure}}          & \multicolumn{4}{c|}{\textbf{Background Preservation}} & \multicolumn{2}{c}{\textbf{CLIP Similarity}}  \\
\cmidrule{1-10}
\textbf{Inverse}          & \textbf{Editing}   &\textbf{Editing Type}           & \textbf{Distance}$_{^{\times 10^3}}$ $\downarrow$ & \textbf{PSNR} $\uparrow$     & \textbf{LPIPS}$_{^{\times 10^3}}$ $\downarrow$  & \textbf{MSE}$_{^{\times 10^4}}$ $\downarrow$     & \textbf{SSIM}$_{^{\times 10^2}}$ $\uparrow$    & \textbf{Whole}  $\uparrow$          & \textbf{Edited}  $\uparrow$       \\

\midrule
\textbf{DDIM} & \textbf{P2P}     &\textbf{Global}           &66.97           &19.12      &165.37   &185.68   &75.70     &{24.78}      &\textbf{23.02}    \\
\textbf{NTI}  & \textbf{P2P}     &\textbf{Global}               &16.56           &{27.50}    &{48.43}  &{34.37}  &{86.10}   &24.40        &21.69      \\
\textbf{NPI}  & \textbf{P2P}     &\textbf{Global}                &17.80           &26.93      &53.82    &36.87    &85.73     &24.42        &21.98      \\
\textbf{StyleD}  & \textbf{P2P}  &\textbf{Global}                &{14.44}         &26.54      &53.52    &38.47    &85.33     &24.49        &21.64     \\
\textbf{PNPInv} & \textbf{P2P}   &\textbf{Global}                 &{12.58}         &{27.80}    &{45.03}  &{31.73}  &{86.62}   &{24.68}      &{22.00}   \\
\midrule
\textbf{TODInv}             & \textbf{P2P}   &\textbf{Global}     &\textbf{9.48}       &\textbf{28.59}  &\textbf{34.90}  &{26.83}         &\textbf{87.40} &\textbf{25.89}      & {21.62}        \\
\midrule
\midrule

\textbf{DDIM} & \textbf{MasaCtrl}       &\textbf{Global}         & 25.61           &23.45  &85.26   &70.79  &82.75 &23.15        &21.10       \\
\textbf{PNPInv}  & \textbf{MasaCtrl}    &\textbf{Global}              & {22.52}           & {23.79}  & {69.85}  & {66.34}  & {84.07} & {23.54}        & {21.12}     \\
\midrule
{\textbf{TODInv}}   & \textbf{MasaCtrl}    & \textbf{Global}           &{19.39}         & {25.29}  &\textbf{55.96}  & {54.13}  &\textbf{85.23} &\textbf{23.90}         &\textbf{22.86}      \\
\midrule
\midrule

\textbf{DDIM}   &\textbf{PNP}  &\textbf{Global}     &29.69 &23.20 & 90.48 & 75.78   &82.32   & 24.90  &\textbf{22.57}  \\
\textbf{PNPInv} & \textbf{PNP} &\textbf{Global}     &{ 27.09} & 23.38 & 84.53 &73.56    & 82.56 &24.81 &22.51   \\
\midrule
{\textbf{TODInv}}       & \textbf{PNP}   &\textbf{Global}   &\textbf{26.74}        &\textbf{25.17}         &\textbf{70.53}  &\textbf{51.64}  &\textbf{84.47}        &\textbf{25.45}            &22.06           \\

\midrule
\midrule

\textbf{DDIM} & \textbf{P2P-Zero}   &\textbf{Global}       &57.89 &21.92  & 125.83 & 112.53 & 79.43 & 23.16 & 21.10 \\
\textbf{PNPInv} & \textbf{P2P-Zero} &\textbf{Global}     & {42.69} & \textbf{22.93} & {99.60} & {98.70} & \textbf{81.40} & {23.80} & \textbf{21.81} \\

\midrule

\textbf{TODInv}             & \textbf{P2P-Zero}    &\textbf{Global}     &\textbf{43.25}       & {22.84}  & \textbf{98.12}  & \textbf{96.21}  &{81.27} & \textbf{24.54}   & {21.50}      \\

\bottomrule
\bottomrule
\end{tabular}

\end{threeparttable}}

\label{tab:cate_global}
\end{table*}

\subsection{More Qualitative Comparison with MasaCtrl, PNP, and P2P-Zero editing methods}

\begin{figure}[h]
    \centering
    \vspace{-2mm}
    \captionsetup[subfloat]{labelformat=empty,justification=centering}
    \subfloat[Source]{
     \begin{minipage}{0.14\linewidth}
     \includegraphics[width=\linewidth]{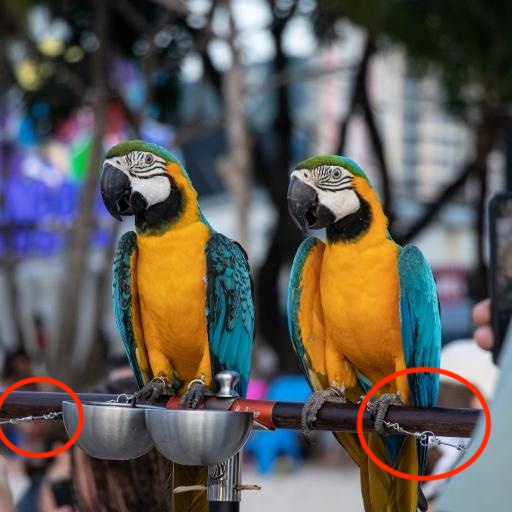}\vspace{3mm}
     \includegraphics[width=\linewidth]{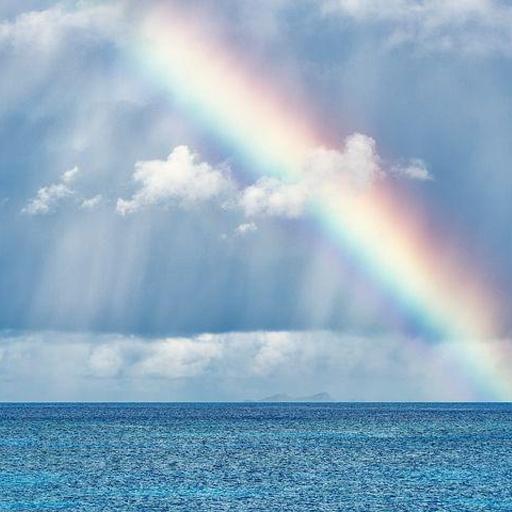}\vspace{3mm}
     \includegraphics[width=\linewidth]{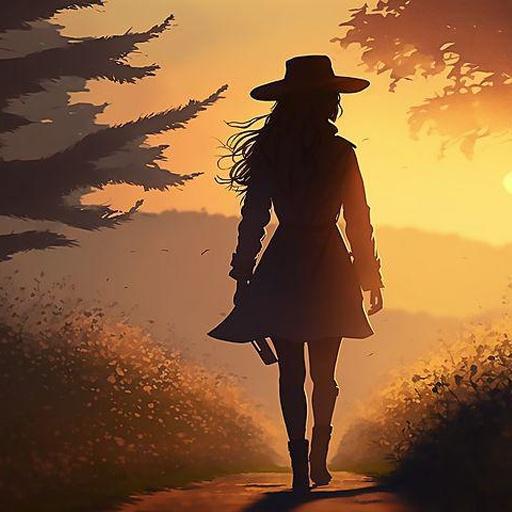}
     \end{minipage}
     }
    \put(-7.5,97){{\scriptsize{\textsc{Two parrots sitting on a stick... $\rightarrow$ Two \rev{kissing} parrots sitting on a stick...}}}}
    \hspace{-2.8mm}
    \subfloat[DDIM]{
     \begin{minipage}{0.14\linewidth}
     \includegraphics[width=\linewidth]{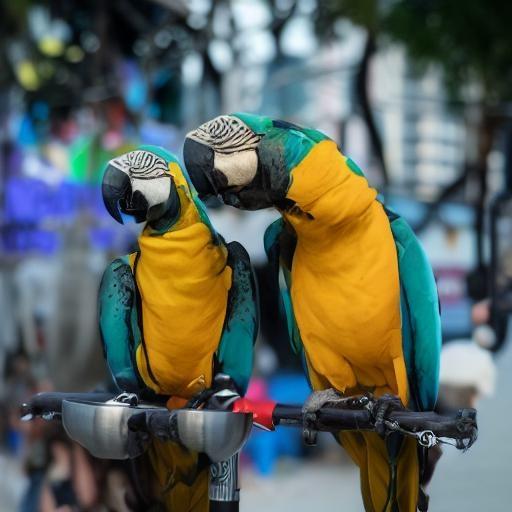}\vspace{3mm}
     \includegraphics[width=\linewidth]{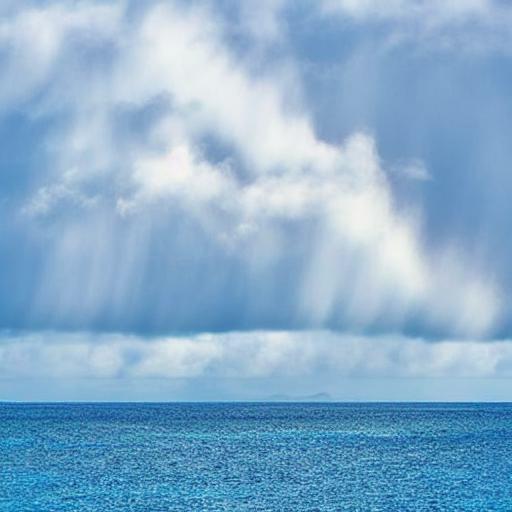}\vspace{3mm}
     \includegraphics[width=\linewidth]{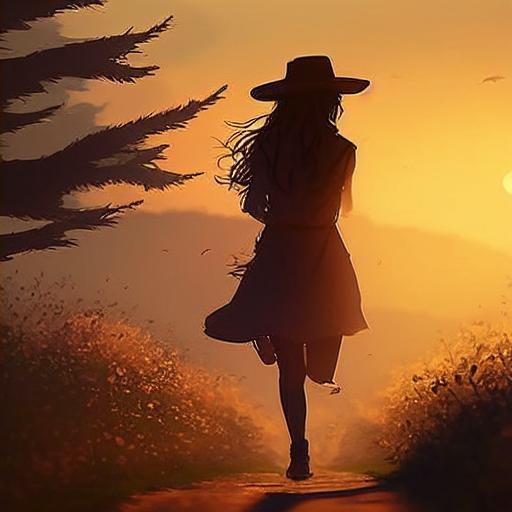}
     \end{minipage}
     }
    \put(-37.5,31.5){{\scriptsize{\textsc{\rev{Rainbow over} the ocean $\rightarrow$ The ocean}}}}
    \hspace{-2.8mm}
    \subfloat[AIDI]{
     \begin{minipage}{0.14\linewidth}
     \includegraphics[width=\linewidth]{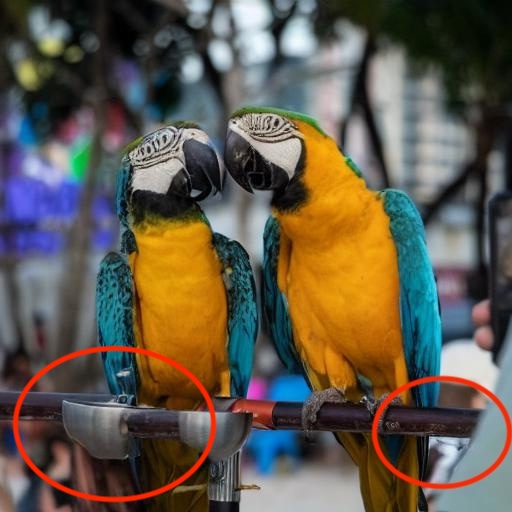}\vspace{3mm}
     \includegraphics[width=\linewidth]{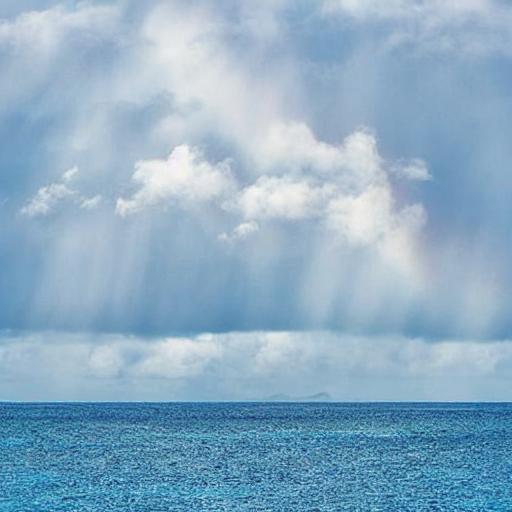}\vspace{3mm}
     \includegraphics[width=\linewidth]{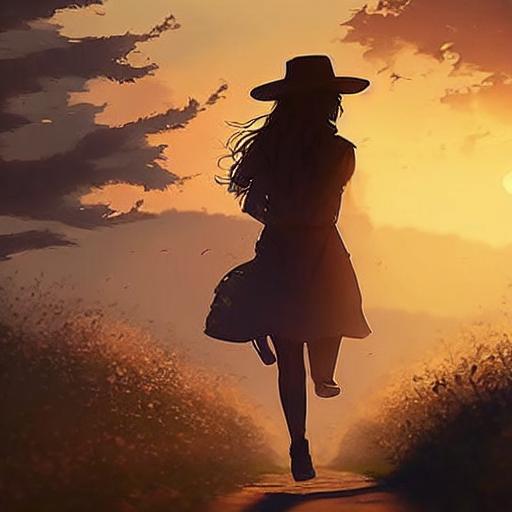}
     \end{minipage}
     }
    \put(-140,-34.5){{\scriptsize{\textsc{A woman in a hat and dress \rev{walking}... $\rightarrow$ A woman in a hat and dress \rev{running} ...}}}}
    \hspace{-2.8mm}
    \subfloat[PNPInv]{
     \begin{minipage}{0.14\linewidth}
     \includegraphics[width=\linewidth]{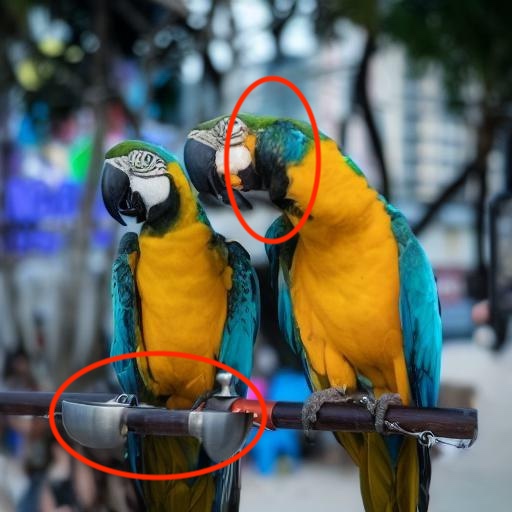}\vspace{3mm}
     \includegraphics[width=\linewidth]{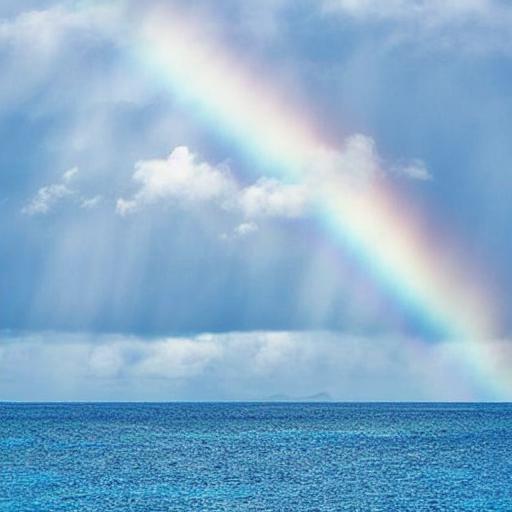}\vspace{3mm}
     \includegraphics[width=\linewidth]{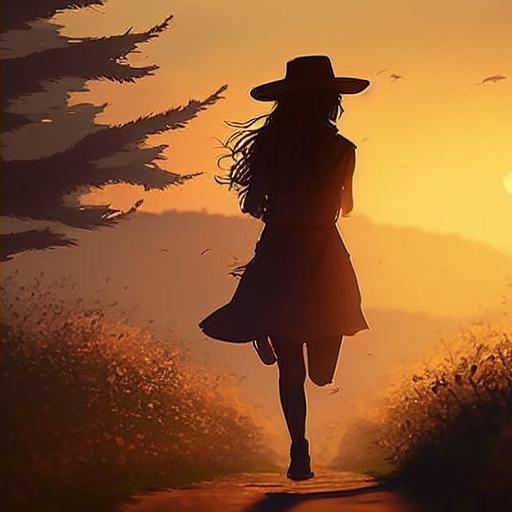}
     \end{minipage}
     }
    \hspace{-2.8mm}
    \subfloat[SPDInv]{
     \begin{minipage}{0.14\linewidth}
     \includegraphics[width=\linewidth]{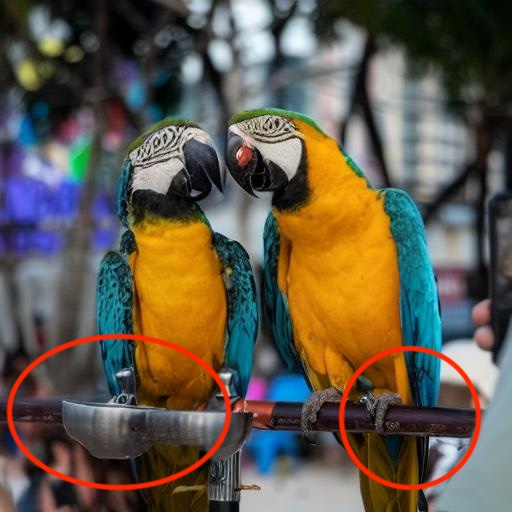}\vspace{3mm}
     \includegraphics[width=\linewidth]{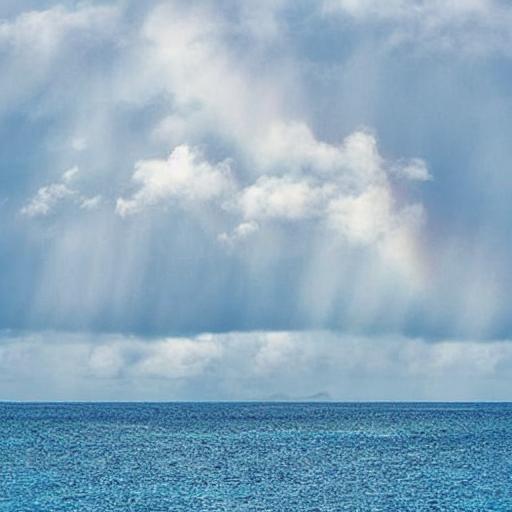}\vspace{3mm}
     \includegraphics[width=\linewidth]{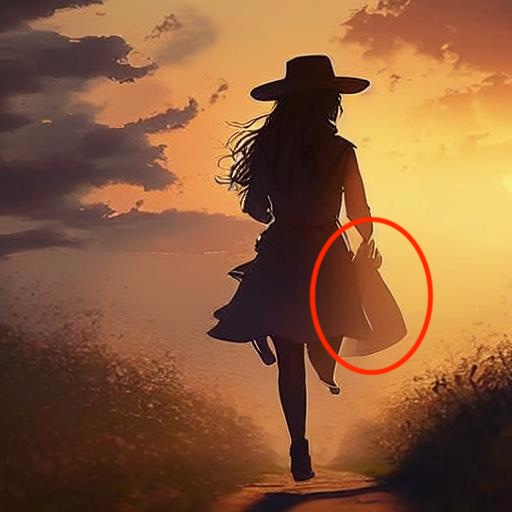}
     \end{minipage}
     }
    \hspace{-2.8mm}
    \subfloat[TODInv]{
     \begin{minipage}{0.14\linewidth}
     \includegraphics[width=\linewidth]{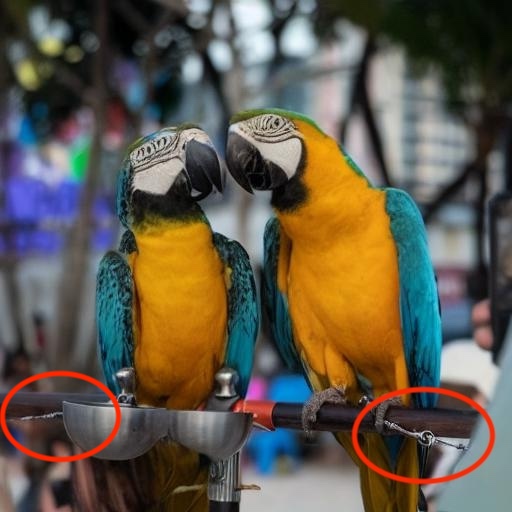}\vspace{3mm}
     \includegraphics[width=\linewidth]{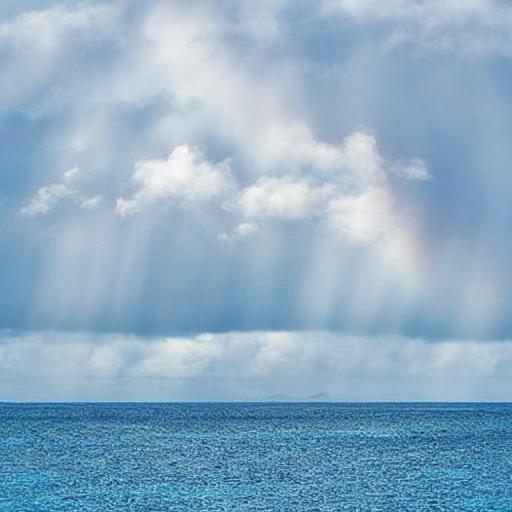}\vspace{3mm}
     \includegraphics[width=\linewidth]{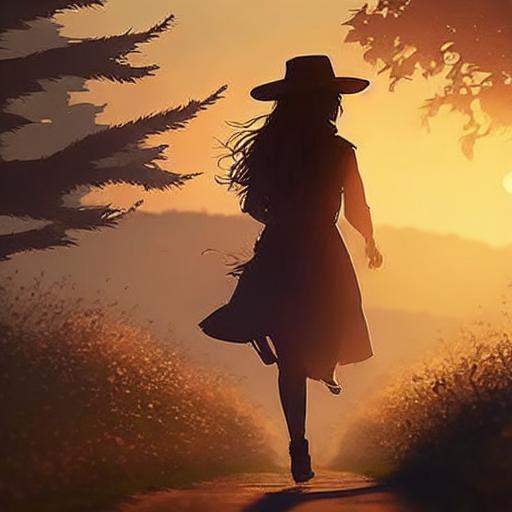}
     \end{minipage}
     }
\vspace{-2mm}
\caption{Qualitative comparison with various inversion methods using MasaCtrl editing method.}
\label{fig:figure_MASA}
\end{figure}

\begin{figure*}[h]
    \centering
    \captionsetup[subfloat]{labelformat=empty,justification=centering}
    \subfloat[Source]{
     \begin{minipage}{0.15\linewidth}
     \includegraphics[width=\linewidth]{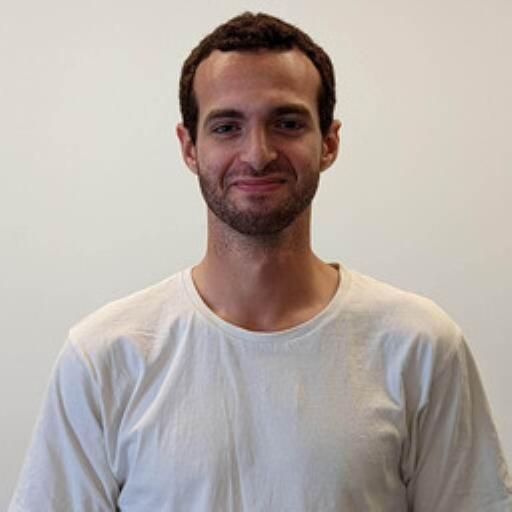}\vspace{3mm}
     \includegraphics[width=\linewidth]{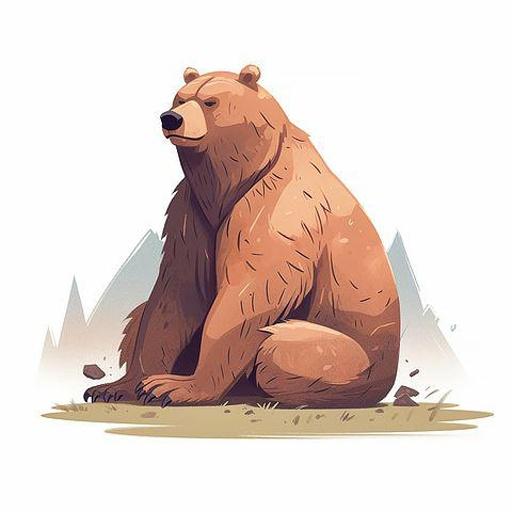}\vspace{3mm}
     \includegraphics[width=\linewidth]{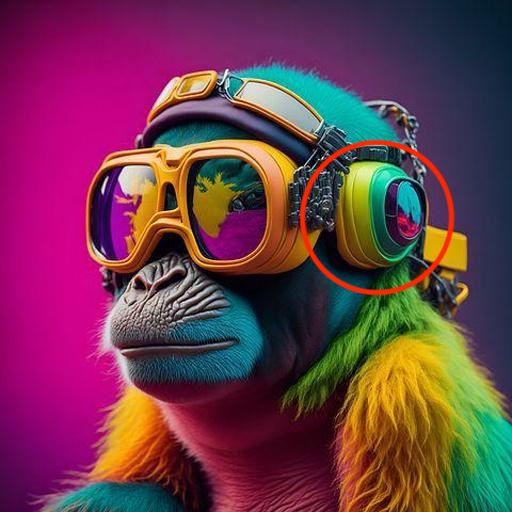}\vspace{3mm}
     \includegraphics[width=\linewidth]{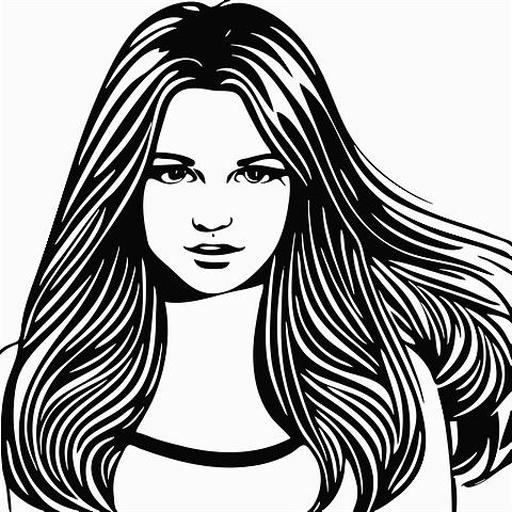}\vspace{3mm}
     \includegraphics[width=\linewidth]{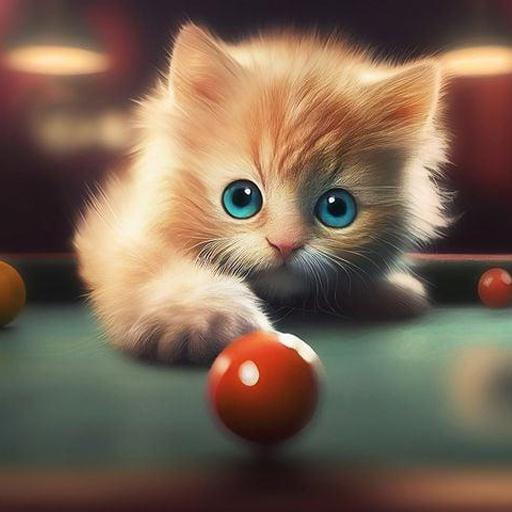}\vspace{3mm}
     \includegraphics[width=\linewidth]{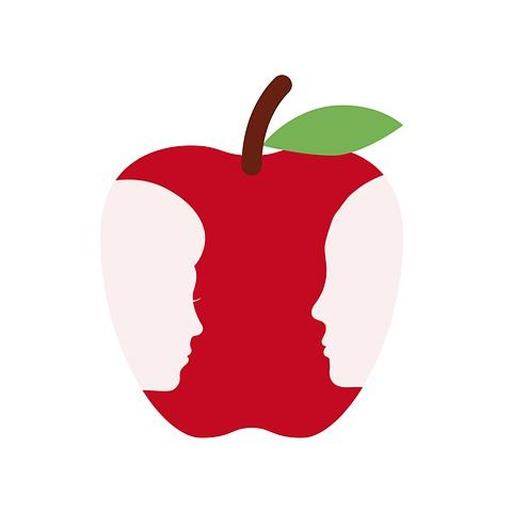}
     \end{minipage}
     }
     \put(-30,207){{\scriptsize{\textsc{A man wearing a \rev{shirt} $\rightarrow$ A man wearing a \rev{sweater}}}}}
     \put(-92.5,137){{\scriptsize{\textsc{A light brown bear \rev{sitting} on the ground $\rightarrow$ A light brown bear \rev{stand} on the ground}}}}
     \put(-85,68){{\scriptsize{\textsc{A \rev{monkey} wearing colorful goggles ... $\rightarrow$ A \rev{man} wearing colorful goggles ...}}}}
     \put(-100,-1){{\scriptsize{\textsc{A \rev{black and white} drawing of a woman ... A $\rightarrow$ \rev{colorful and detailed} drawing of a woman ...}}}}
     \put(-45,-70){{\scriptsize{\textsc{A kitten \rev{playing with balls} $\rightarrow$ A kitten}}}}
     \put(-125,-140){{\scriptsize{\textsc{An apple with two \rev{faces} on it in a white background $\rightarrow$ An apple with two \rev{hands} on it in a white background}}}}
    \hspace{-2.8mm}
    \subfloat[DDIM]{
     \begin{minipage}{0.15\linewidth}
     \includegraphics[width=\linewidth]{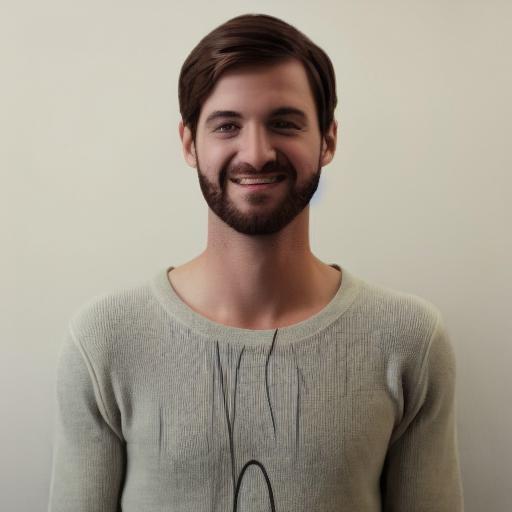}\vspace{3mm}
     \includegraphics[width=\linewidth]{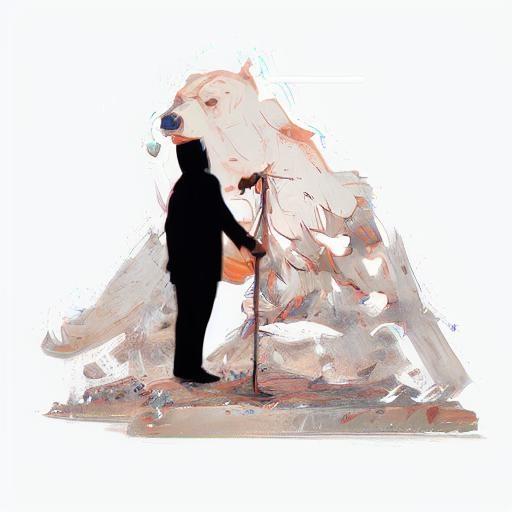}\vspace{3mm}
     \includegraphics[width=\linewidth]{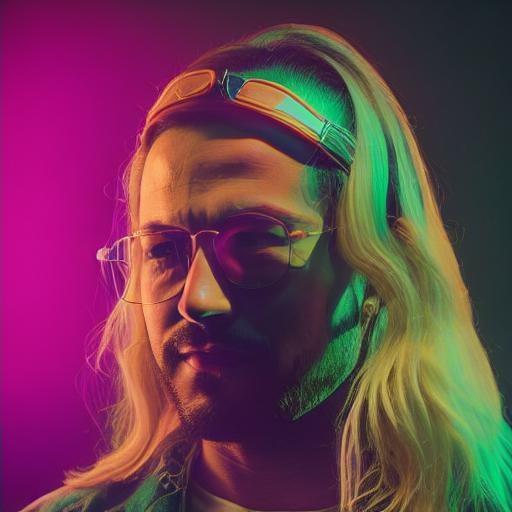}\vspace{3mm}
     \includegraphics[width=\linewidth]{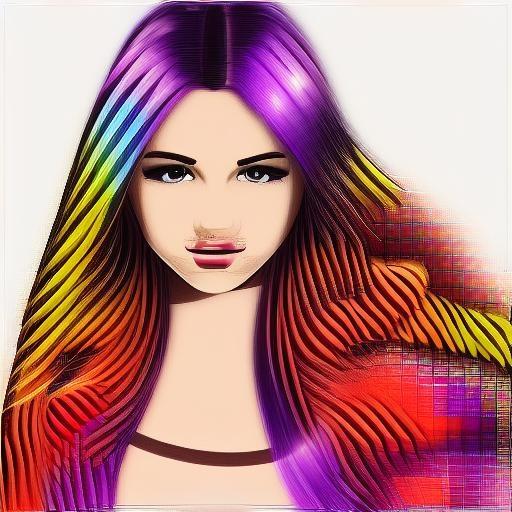}\vspace{3mm}
     \includegraphics[width=\linewidth]{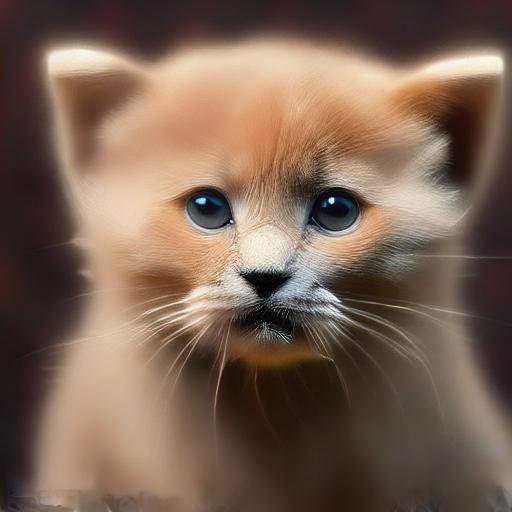}\vspace{3mm}
     \includegraphics[width=\linewidth]{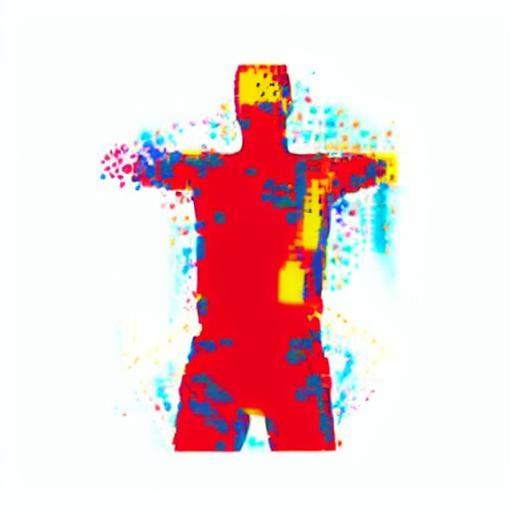}
     \end{minipage}
     }
     \hspace{-2.8mm}
    \subfloat[PNPInv]{
     \begin{minipage}{0.15\linewidth}
     \includegraphics[width=\linewidth]{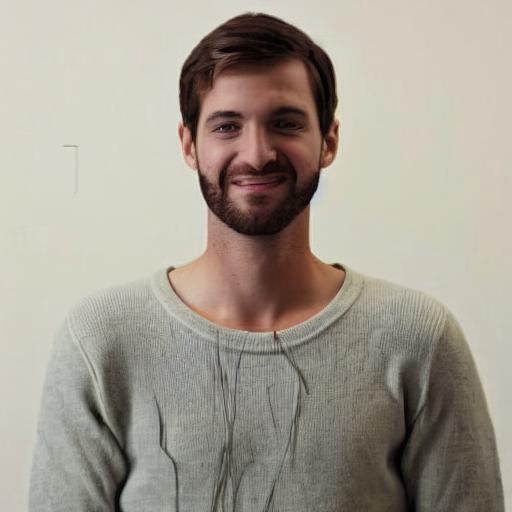}\vspace{3mm}
     \includegraphics[width=\linewidth]{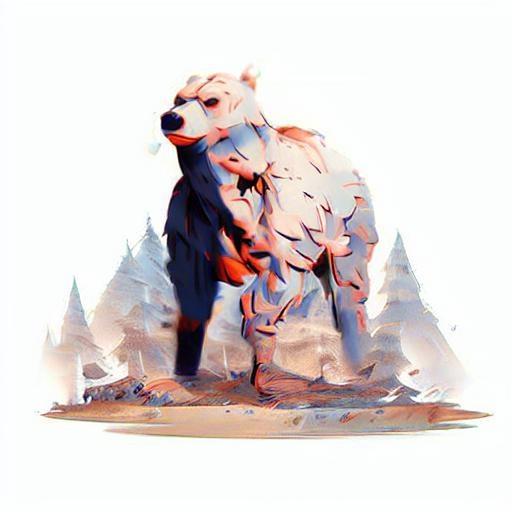}\vspace{3mm}
     \includegraphics[width=\linewidth]{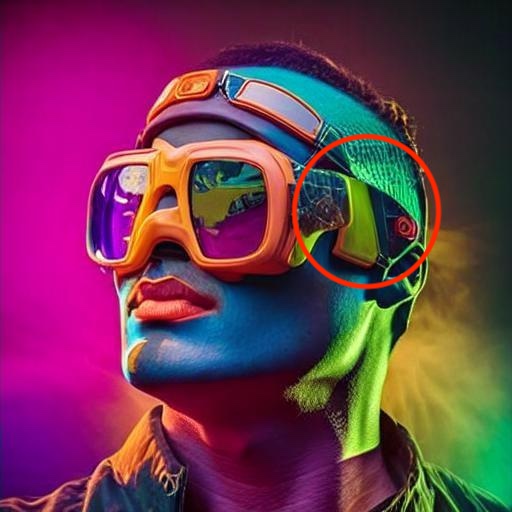}\vspace{3mm}
     \includegraphics[width=\linewidth]{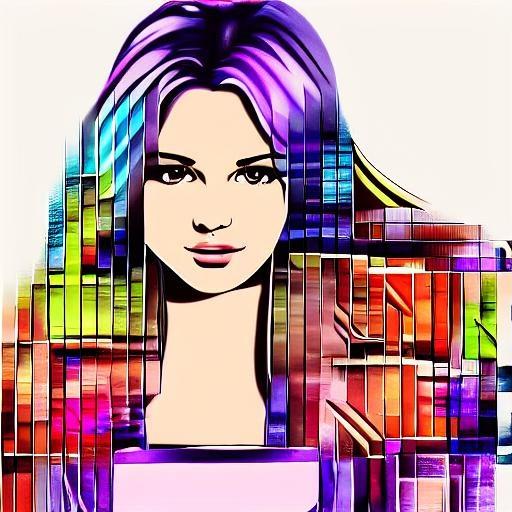}\vspace{3mm}
     \includegraphics[width=\linewidth]{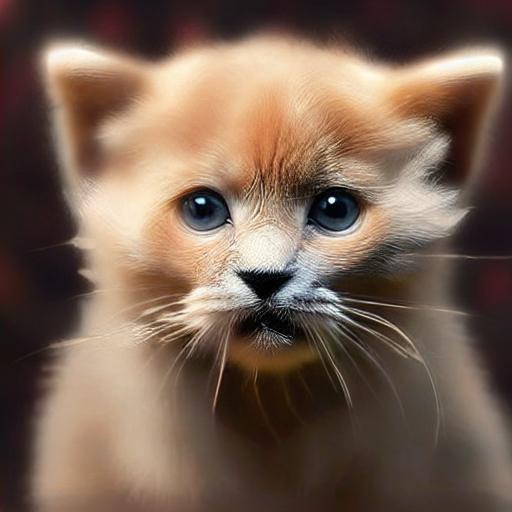}\vspace{3mm}
     \includegraphics[width=\linewidth]{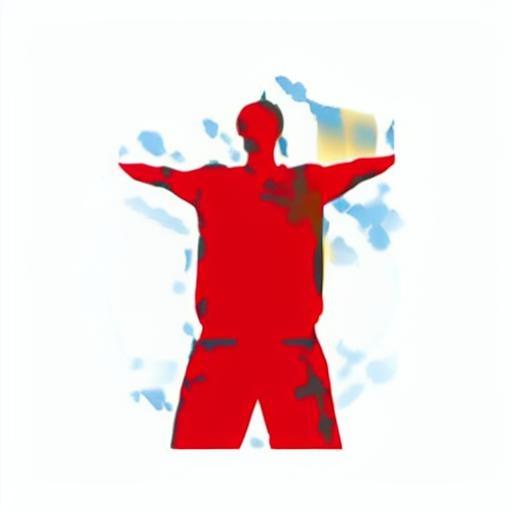}

     \end{minipage}
     }
     \hspace{-2.8mm}
    \subfloat[TODInv]{
     \begin{minipage}{0.15\linewidth}
     \includegraphics[width=\linewidth]{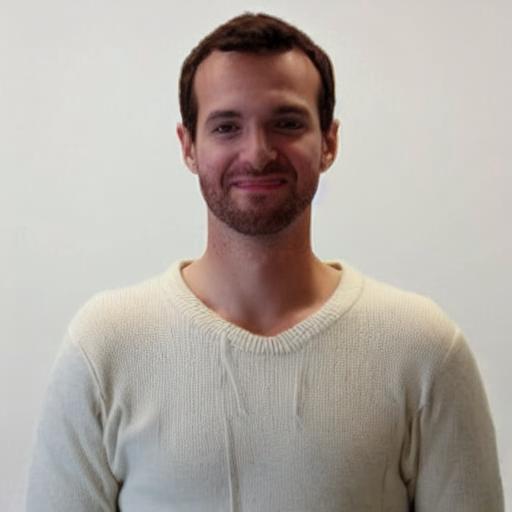}\vspace{3mm}
     \includegraphics[width=\linewidth]{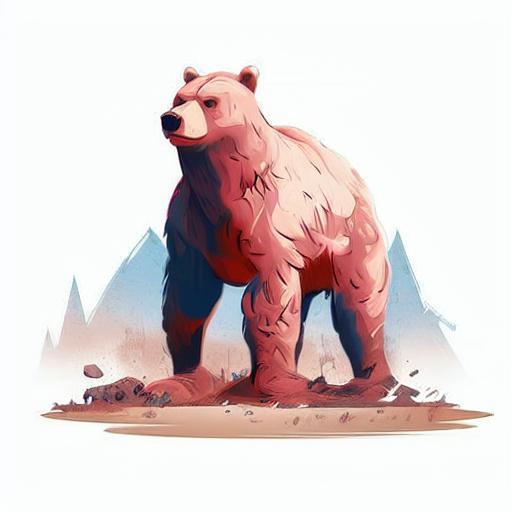}\vspace{3mm}
     \includegraphics[width=\linewidth]{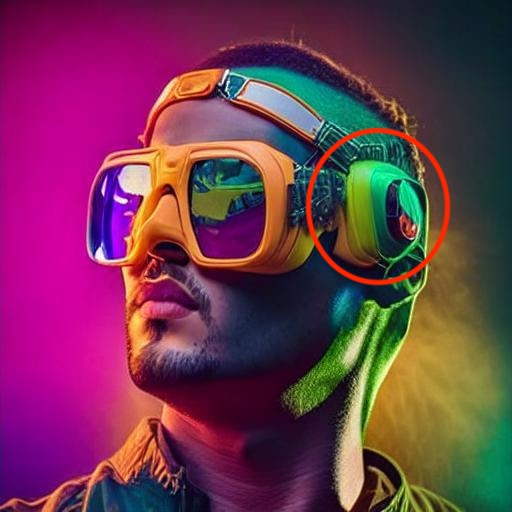}\vspace{3mm}
     \includegraphics[width=\linewidth]{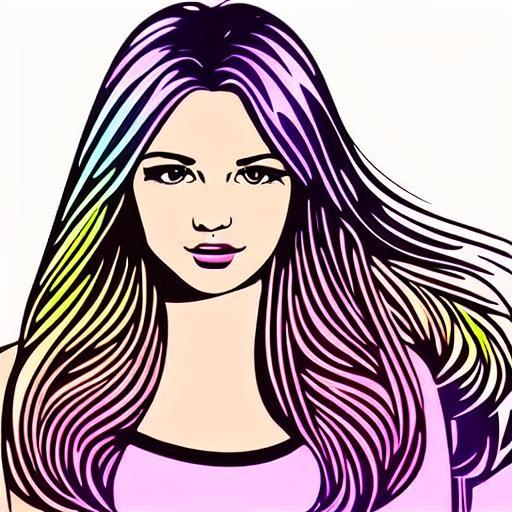}\vspace{3mm}
     \includegraphics[width=\linewidth]{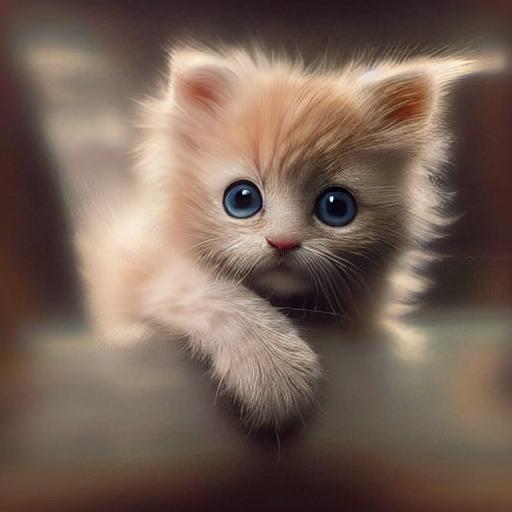}\vspace{3mm}
     \includegraphics[width=\linewidth]{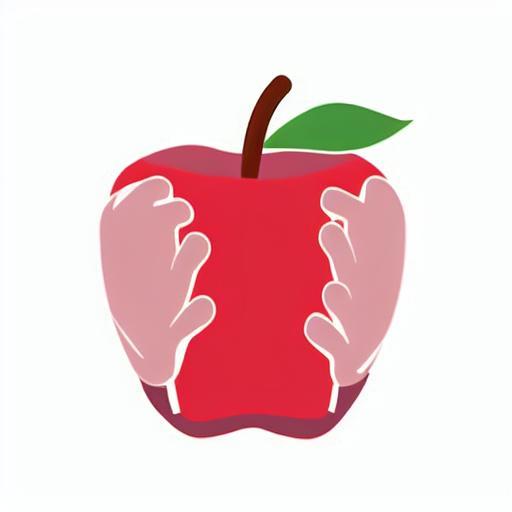}
     \end{minipage}
     }
\vspace{-4mm}
\caption{Qualitative comparison with various inversion methods using P2P-Zero editing method.}
\label{fig:figure_pzero}
\end{figure*}

\begin{figure}[h]
    \centering
    \captionsetup[subfloat]{labelformat=empty,justification=centering}
    \subfloat[Source]{
     \begin{minipage}{0.14\linewidth}
     \includegraphics[width=\linewidth]{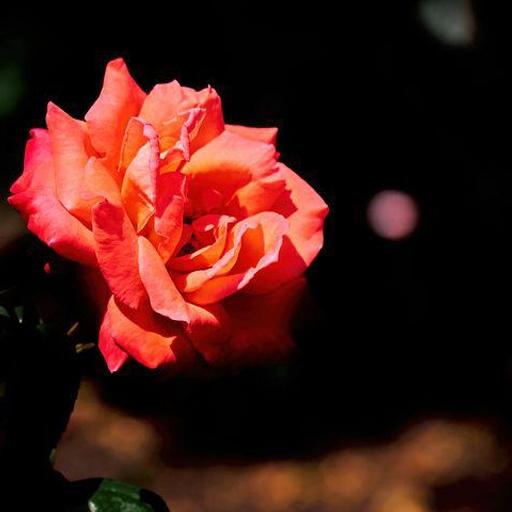}\vspace{3mm}
     \includegraphics[width=\linewidth]{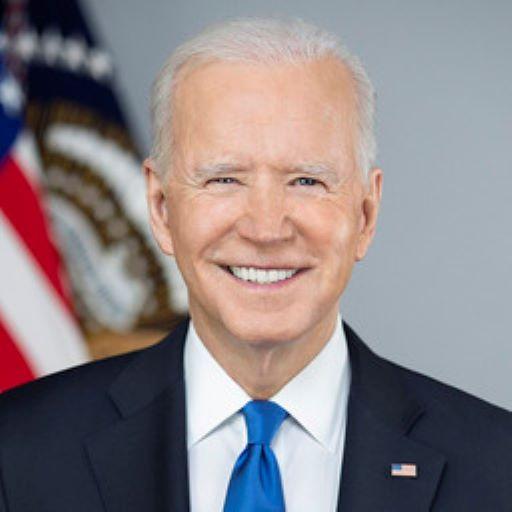}\vspace{3mm}
     \includegraphics[width=\linewidth]{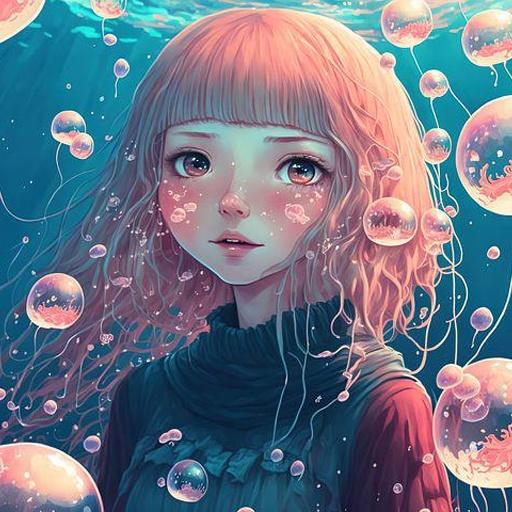}\vspace{3mm}
     \includegraphics[width=\linewidth]{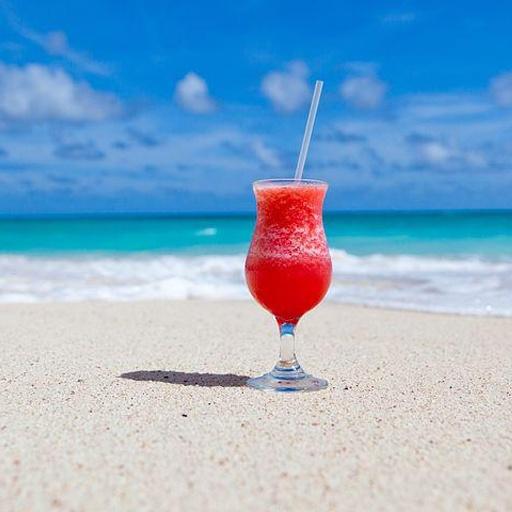}
     \end{minipage}
     }
    \put(25,130){{\scriptsize{\textsc{A \rev{red} rose in the dark $\rightarrow$ A \rev{blue} rose in the dark}}}}
    \hspace{-2.8mm}
    \subfloat[DDIM]{
     \begin{minipage}{0.14\linewidth}
     \includegraphics[width=\linewidth]{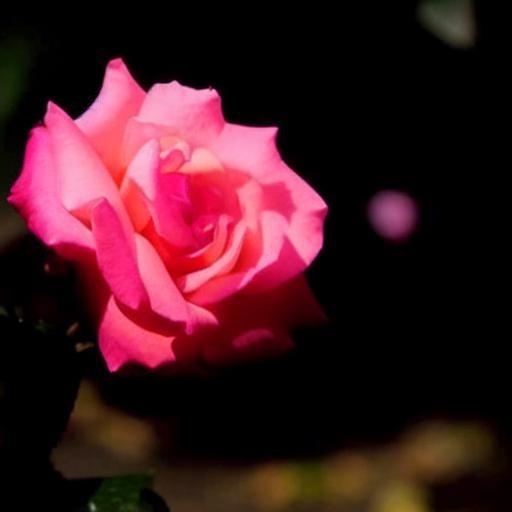}\vspace{3mm}
     \includegraphics[width=\linewidth]{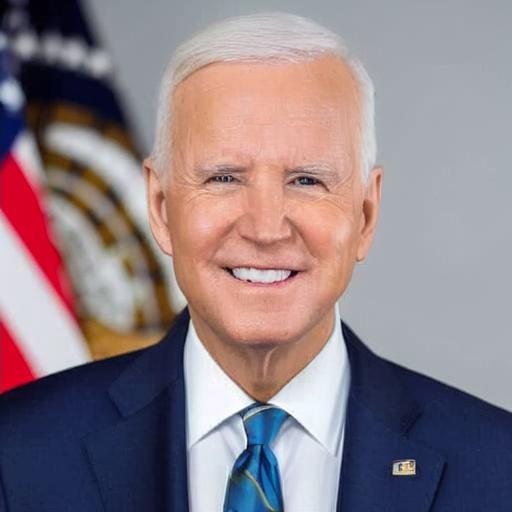}\vspace{3mm}
     \includegraphics[width=\linewidth]{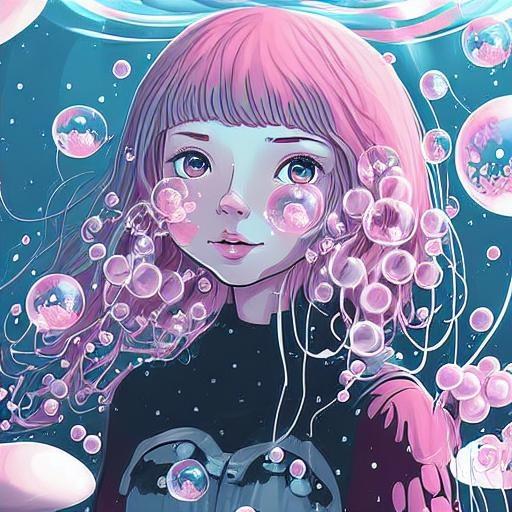}\vspace{3mm}
     \includegraphics[width=\linewidth]{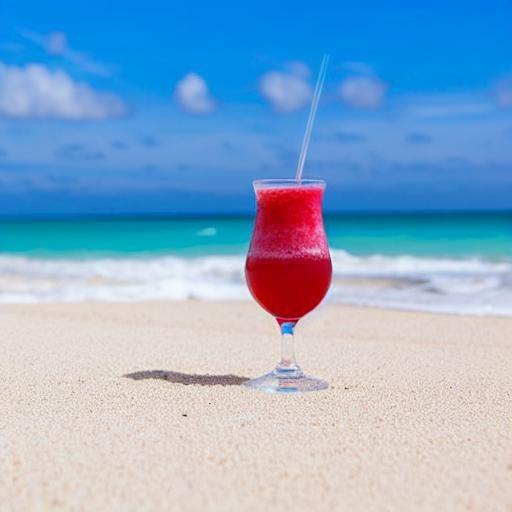}
     \end{minipage}
     }
    \put(-75,64.5){{\scriptsize{\textsc{A man wearing a tie $\rightarrow$A man wearing a~\rev{black and yellow stripes} tie}}}}
    \hspace{-2.8mm}
    \subfloat[AIDI]{
     \begin{minipage}{0.14\linewidth}
     \includegraphics[width=\linewidth]{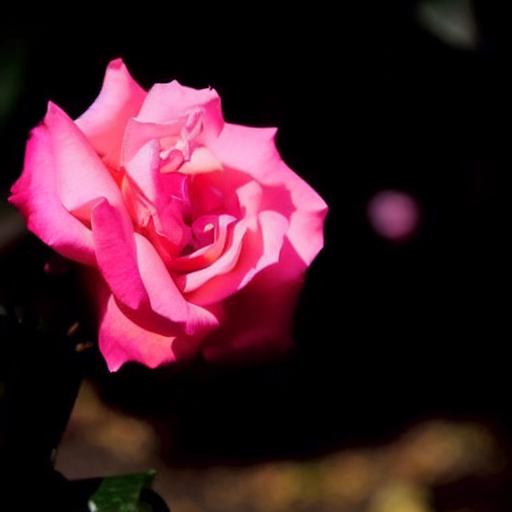}\vspace{3mm}
     \includegraphics[width=\linewidth]{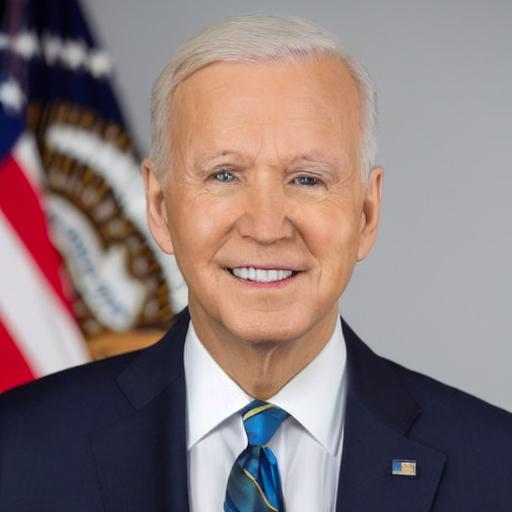}\vspace{3mm}
     \includegraphics[width=\linewidth]{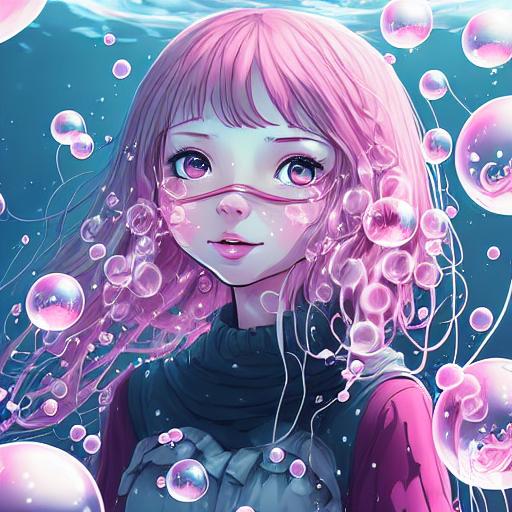}\vspace{3mm}
     \includegraphics[width=\linewidth]{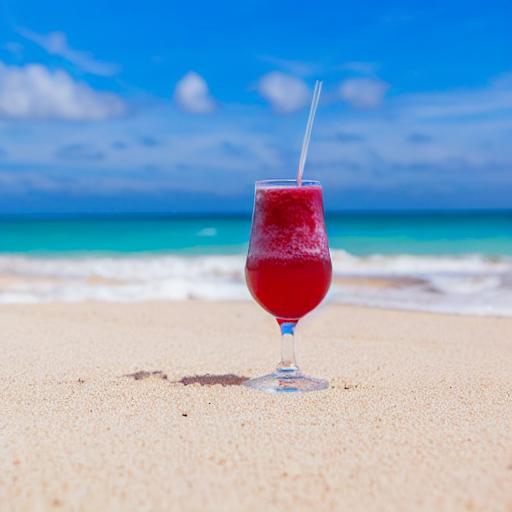}
     \end{minipage}
     }
    \hspace{-2.8mm}
    \subfloat[PNPInv]{
     \begin{minipage}{0.14\linewidth}
     \includegraphics[width=\linewidth]{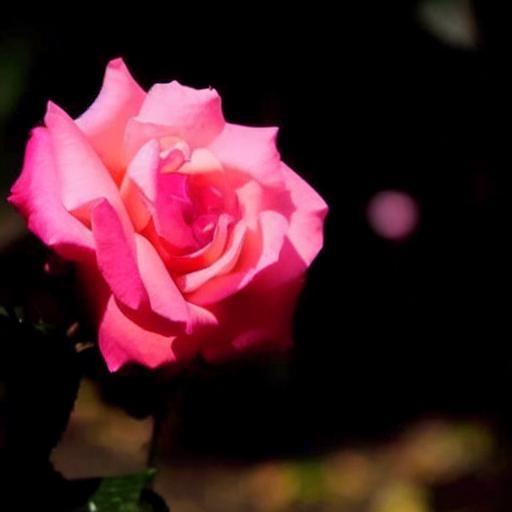}\vspace{3mm}
     \includegraphics[width=\linewidth]{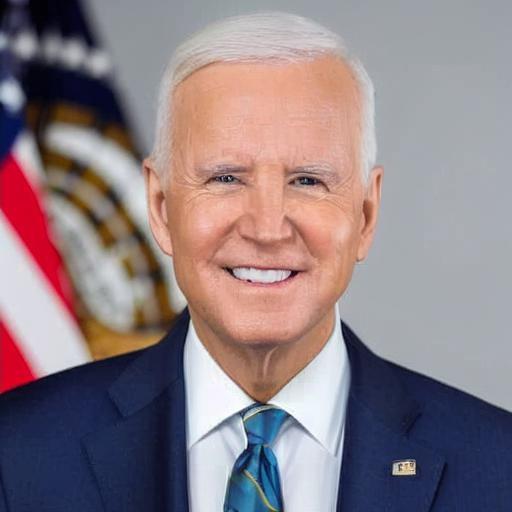}\vspace{3mm}
     \includegraphics[width=\linewidth]{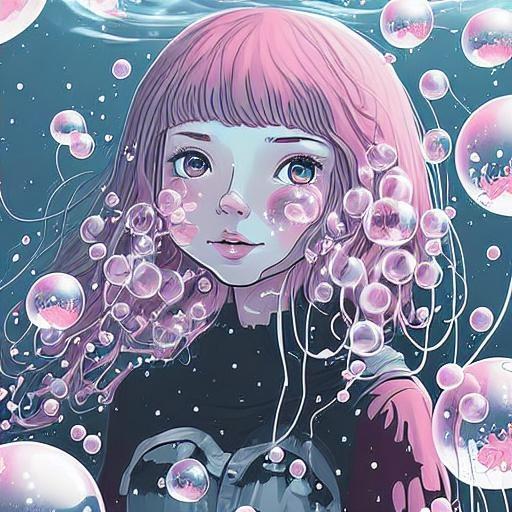}\vspace{3mm}
     \includegraphics[width=\linewidth]{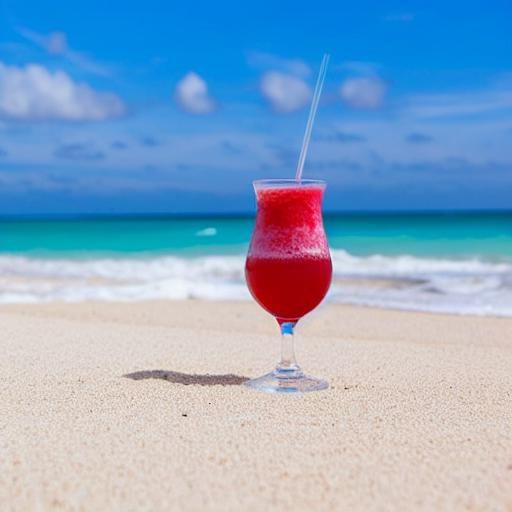}
     \end{minipage}
     }
    \hspace{-2.8mm}
    \subfloat[SPDInv]{
     \begin{minipage}{0.14\linewidth}
     \includegraphics[width=\linewidth]{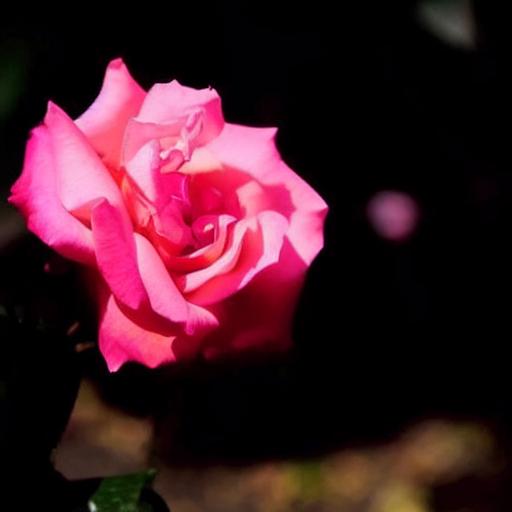}\vspace{3mm}
     \includegraphics[width=\linewidth]{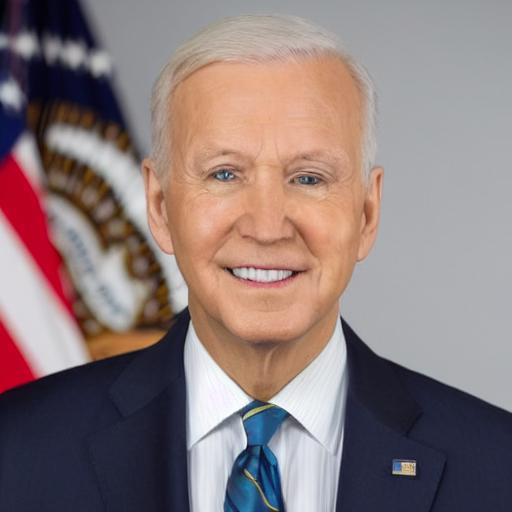}\vspace{3mm}
     \includegraphics[width=\linewidth]{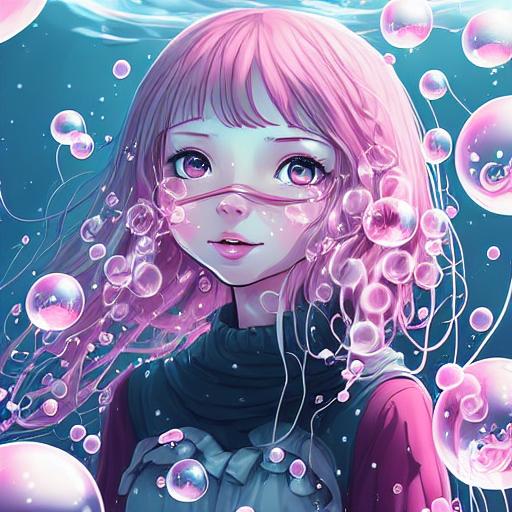}\vspace{3mm}
     \includegraphics[width=\linewidth]{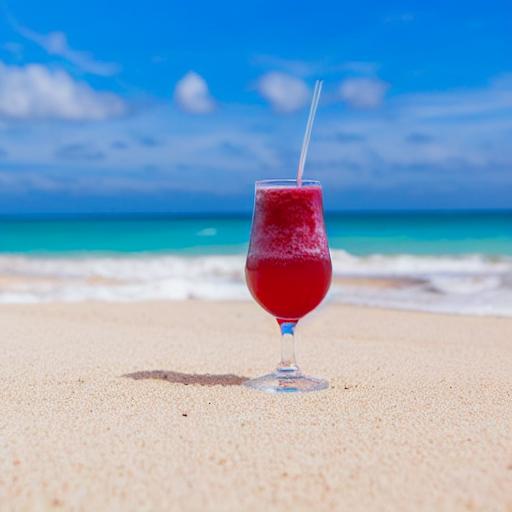}
     \end{minipage}
     }
    \put(-250,-1){{\scriptsize{\textsc{A girl with pink hair ... $\rightarrow$ \rev{A black and white sketch of} a girl with pink hair ...}}}}
    \hspace{-2.8mm}
    \subfloat[TODInv]{
     \begin{minipage}{0.14\linewidth}
     \includegraphics[width=\linewidth]{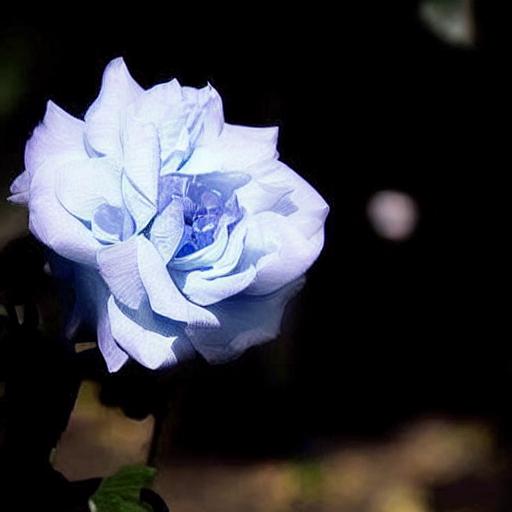}\vspace{3mm}
     \includegraphics[width=\linewidth]{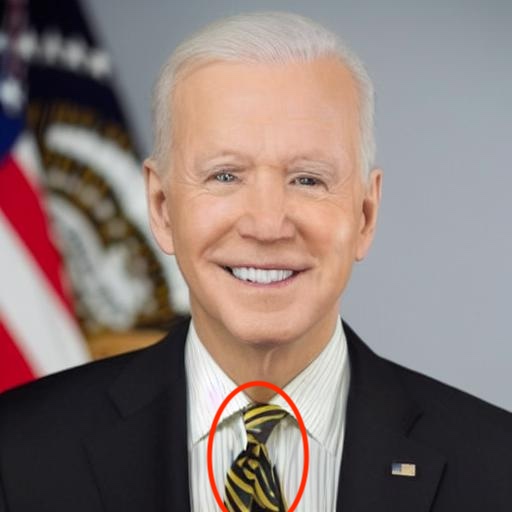}\vspace{3mm}
     \includegraphics[width=\linewidth]{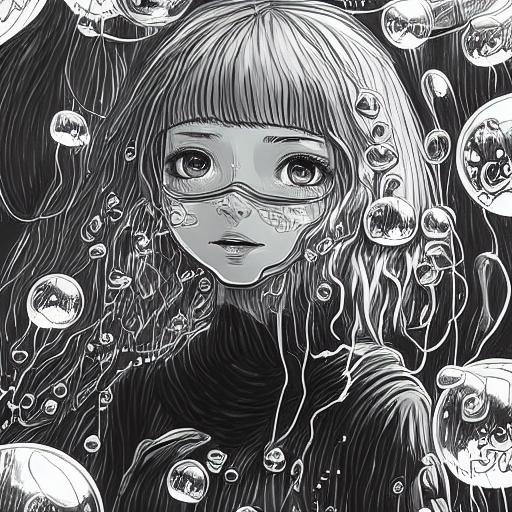}\vspace{3mm}
     \includegraphics[width=\linewidth]{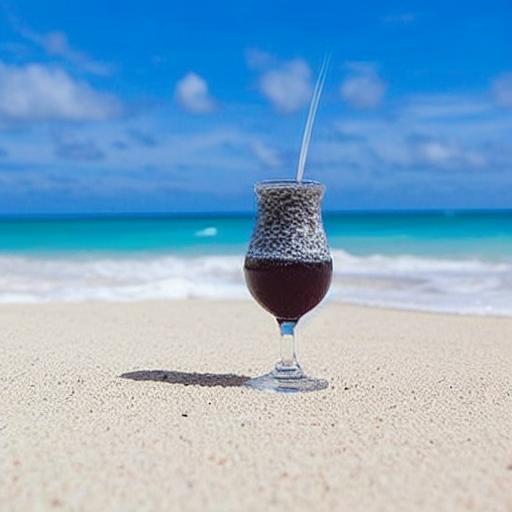}
     \end{minipage}
     }
    \put(-290,-66.5){{\scriptsize{\textsc{A glass of~\rev{red drink} on the beach $\rightarrow$ A glass of \rev{red wine} on the beach}}}}
\vspace{-2mm}
\caption{Qualitative comparison with various inversion methods using PNP editing method.}
\label{fig:figure_PNP}
\end{figure}

The qualitative comparison based on MasaCtrl, PNP, and P2P-Zero editing methods are shown in Fig.~\ref{fig:figure_MASA}, Fig.~\ref{fig:figure_PNP}, and Fig.~\ref{fig:figure_pzero} respectively. 

As shown in the $2_{nd}$ sample of Fig.~\ref{fig:figure_PNP}, all competitors fail on local appearance editing. In the $3_{rd}$ sample, none of the competitors capture the editing instruction of ``\textsc{A black and white sketch}'', and pay more attention to ``\textsc{Pink}'' incorrectly. The same problem also emerged on modifying the ``\textsc{Red drink}'' to ``\textsc{Red wine}''. That evidences the effectiveness of our method in capturing semantic instruction. 

As shown in the red cycles in $1_{st}$ sample of Fig.~\ref{fig:figure_MASA}, most of competitors can not preserve the chains in the original image. Our TODInv is also skilled at object removal rainbow.

In Fig.~\ref{fig:figure_pzero}, DDIM and PNPInv fail to preserve face details when editing the  ``\textsc{Shirt}'' to ``\textsc{Sweater}'', and they also failed to preserve the color of the bear. Our TODInv preserves more source details during editing. That should contribute to our task-oriented strategy, as we optimize the prompt embeddings that are irrelevant to the current editing, which preserves the source details effectively.

\subsection{The Algorithm of TODInv}

The algorithm of our TODInv inversion and editing can be seen in Alg.~\ref{alg1}.

\begin{algorithm}
    \SetKwInOut{KwResult}{Part \uppercase\expandafter{\romannumeral1}}
    \SetKwInOut{KwData}{Part \uppercase\expandafter{\romannumeral2}}

    \renewcommand{\algorithmicrequire}{\textbf{Input:}}
    \renewcommand{\algorithmicensure}{\textbf{Output:}}
    \caption{Algorithm of TODInv.}
    \label{alg:inversion}
    \KwResult{\textbf{Inversion Pipeline}}
    \begin{algorithmic}[1]
        \REQUIRE Source image latent $z_0$, DDIM steps $T$, source prompt embedding $P$, maximal optimization step $K$, threshold $\delta$, Editing type \texttt{Type}.
        \ENSURE  Latent noise $z_T$, Optimized prompt embedding in each timestep ${P}^*_{t}$.
        \FOR{$t\leftarrow1$ to T}
            \STATE Get $z_{t}$ from $z_{t-1}$ using DDIM inversion (Eq.~\ref{eq.ddim_inversion2});
            \FOR{$i\leftarrow0$ to K}
                \STATE Initialize the current prompt embedding ${P}_{t}$ as $P$;
                \STATE Update $z_{t}^\prime$ using $z_{t}$ and ${P}_{t}$ (Eq.~\ref{eq.ddim_inversion7});
                \STATE Optimize specific layers of ${P}^*_{t}$ (determined by \texttt{Type}) by minimizing $\|z_{t} - z_{t}^\prime\|^2_2$ (Eq.~\ref{eq.ddim_inversion8});
                \STATE \textbf{if} $\|z_{t} - z_{t}^\prime\|^2_2<\delta$ \textbf{then} \textit{Break} \textbf{end if}
            \ENDFOR
        \ENDFOR
    \end{algorithmic} 
    \vspace{1mm} \hrule  \vspace{1mm}

    \KwData{\textbf{Reconstruction and Edit Pipeline}}
    \begin{algorithmic}[1]
        \REQUIRE Target prompt embedding ${P}{^{target}}$, latent noise $z_T$, optimized prompt embedding in each timestep ${P}^*_{t}$, text-guided image editing method $E$.
        \ENSURE Reconstructed latent $z^r_0$, Edited latent $z^e_0$.
        \FOR{$t\leftarrow T $ to 0}
            \STATE Update the reconstructed latent $z^r_t$ based $z_T$ and ${P}^*_{t}$ using DDIM sampler;
            \STATE Renew the target prompt embedding $\tilde{P}{^{target}_t}$ using ${P}^*_{t}$ and ${P}{^{target}_t}$ (Eq.~\ref{eq.ddim_inversion9});
            \STATE Update the edited latent $z^e_t$ based $z_T$ and $\tilde{P}{^{target}_t}$ using $E$,  
        \ENDFOR

    \end{algorithmic} 
\label{alg1}
\end{algorithm}


\subsection{Editing Type Determination}

\begin{figure*}
\centering
\hspace{-1.5mm}\includegraphics[width=0.99\textwidth]{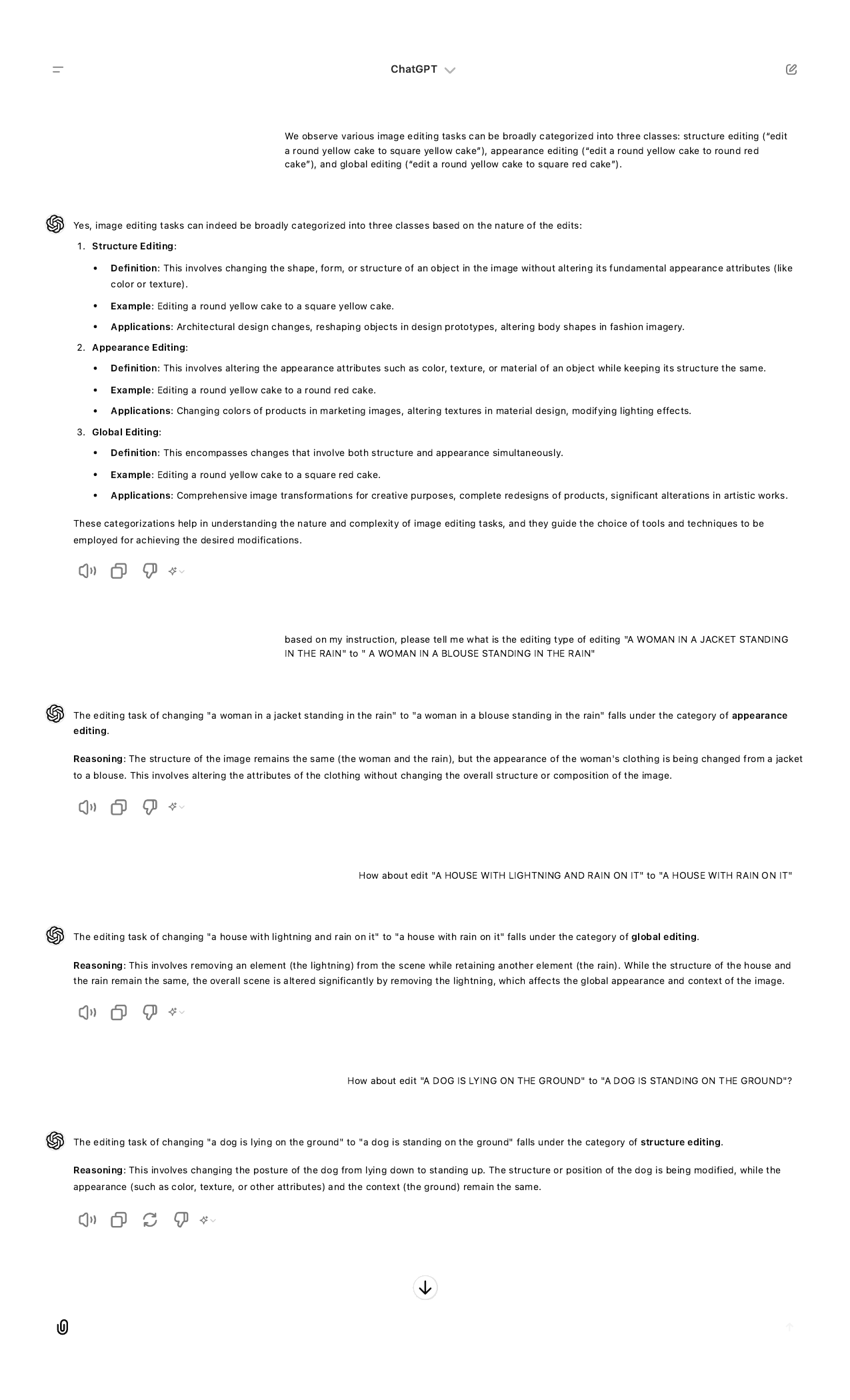}
\vspace{-2.5mm}
\caption{Illustration of determining editing types with ChatGPT.}
\label{fig:chatgpt}
\vspace{-3mm}
\end{figure*}

As discussed in Sec.~\ref{sec:conclusion}, the main limitation of ToDInv is we need to determine the editing type before inversion. It may be not easy for unprofessional users. However, it is easy to determine the edit type based on the source and target prompts using ChatGPT. We present the illustration of determining editing types with ChatGPT in Fig.~\ref{fig:chatgpt}, we can see that it is easy to determine the editing type with our given prompts.

\end{document}